\documentclass{article}

\usepackage[final]{corl_2022} 

\usepackage{booktabs}
\usepackage{wrapfig}
\usepackage{tikz}
\usepackage{multirow}
\usepackage{svg}
\usepackage{amsmath}
\usepackage{amssymb}
\usepackage{enumitem}
\usepackage{caption}
\usepackage{subcaption}

\usepackage[normalem]{ulem}

\usepackage[skip=2.5pt]{parskip}

\usepackage{etoolbox}
\usepackage{pgf}
\usepackage{colortbl}
\definecolor{high}{HTML}{ef3b2c}  
\definecolor{low}{HTML}{fff7bc}  
\newcommand*{\opacity}{90}
\newcommand*{\minval}{0.0}
\newcommand*{\maxval}{1.0}
\newcommand{\gradient}[1]{
	\ifdimcomp{#1pt}{>}{\maxval pt}{#1}{
		\ifdimcomp{#1pt}{<}{\minval pt}{#1}{
			\pgfmathparse{int(round(100*(#1/(\maxval-\minval))-(\minval*(100/(\maxval-\minval)))))}
			\xdef\tempa{\pgfmathresult}
			\cellcolor{high!\tempa!low!\opacity} #1
	}}
}

\title{Human-Robot Commensality: Bite Timing Prediction for Robot-Assisted Feeding in Groups}

\author{
    Jan Ondras$^*$\\
    Cornell University\\
    \texttt{janko@cs.cornell.edu}
    \And
    Abrar Anwar$^*$\\
    University of Southern California\\
    \texttt{abrar.anwar@usc.edu}
    \And
    Tong Wu$^*$\\
    Rutgers University\\
    \texttt{tong.wu96@rutgers.edu}
    \And
    Fanjun Bu\\
    Cornell Tech\\
    \texttt{fb266@cornell.edu}
    \And
    Malte Jung\\
    Cornell University\\
    \texttt{mfj28@cornell.edu}
    \And
    Jorge Jose Ortiz\\
    Rutgers University\\
    \texttt{jorge.ortiz@rutgers.edu}
    \And
    Tapomayukh Bhattacharjee\\
    Cornell University\\
    \texttt{tapomayukh@cornell.edu}
}

\begin{document}
\maketitle
\def\thefootnote{*}\footnotetext{These authors contributed equally to this work. }
\def\thefootnote{\arabic{footnote}}


\begin{abstract}
    We develop data-driven models to predict when a robot should feed during social dining scenarios. Being able to eat independently with friends and family is considered one of the most memorable and important activities for people with mobility limitations. While existing robotic systems for feeding people with mobility limitations focus on solitary dining, commensality, the act of eating together, is often the practice of choice. Sharing meals with others introduces the problem of socially appropriate bite timing for a robot, i.e. the appropriate timing for the robot to feed without disrupting the social dynamics of a shared meal.
    Our key insight is that bite timing strategies that take into account the delicate balance of social cues can lead to seamless interactions during robot-assisted feeding in a social dining scenario. We approach this problem by collecting a Human-Human Commensality Dataset (HHCD) containing 30 groups of three people eating together. We use this dataset to analyze human-human commensality behaviors and develop bite timing prediction models in social dining scenarios. 
    We also transfer these models to human-robot commensality scenarios. Our user studies show that prediction improves when our algorithm uses multimodal social signaling cues between diners to model bite timing.
    The HHCD dataset, videos of user studies, and code are available at \href{https://emprise.cs.cornell.edu/hrcom/}{https://emprise.cs.cornell.edu/hrcom/}
\end{abstract}


\section{Introduction}\label{sec:introduction}

Nearly $27\%$ of people living in the United States have a disability, and close to $24$ million people aged $18$ years or older need assistance with activities of daily living (ADL)~\cite{united2014americans}. Key among these activities is \textit{feeding}, which is both time-consuming for the caregiver, and challenging for the care recipient (patient) to accept socially~\cite{perry2008assisted}. Indeed, needing help with one or more ADLs is the most cited reason for moving to assisted or institutionalized living~\cite{Mlinac2016, agisLiving}. 
Although there are several automated feeding systems on the market~\cite{obi, myspoon, mealmate, mealbuddy, winsfordfeeder, bestic, mealtimepartner, neater, beeson}, they have lacked widespread acceptance. One of the key reasons is that all of them require manual triggering of bite timing by the user, which is challenging for users with cognitive disabilities and inconvenient in social settings. A key challenge for the realization of autonomous robotic feeding systems is therefore to infer proper bite timing~\cite{bhattacharjee2018food}. 

While existing systems focus on solitary dining (e.g.~\cite{candeias2018vision, gordon2021leveraging, higa2014vision, bhattacharjee2019community, jardon2012functional, bhattacharjee2019towards, alqasemi2005wheelchair, bien2004integration, park2016towards, feng2019robot, gallenberger2019transfer, belkhale2021balancing, canal2016personalization, rhodes2018robot, naotunna2015meal, gordon2019learning, park2018multimodal, park2017multimodal}), \textbf{commensality}, the act of eating together, is often the practice of choice. People like to share meals with others. The social experience of a shared meal is an important part of the overall eating experience and current robot feeding systems are not designed with that experience in mind. Transferring the challenge of inferring appropriate bite timing to a social dining setting requires not only attuning to the user's eating behavior but also to the complex social dynamics of the group. For example, a robot should not attempt to feed a user who is actively engaged in conversation. Motivated by a growing body of research that seeks to develop models for robots to function in group settings \cite{sebo2020robots, jung2020robot} we ask the seemingly simple question: \textit{How should an assistive feeding robot decide the right timing for feeding a user in ever-changing and dynamic social dining scenarios?}

\begin{figure}[t]
	\centering
		\includegraphics[width=.95\textwidth,clip,trim=0 1.5cm 0 0]{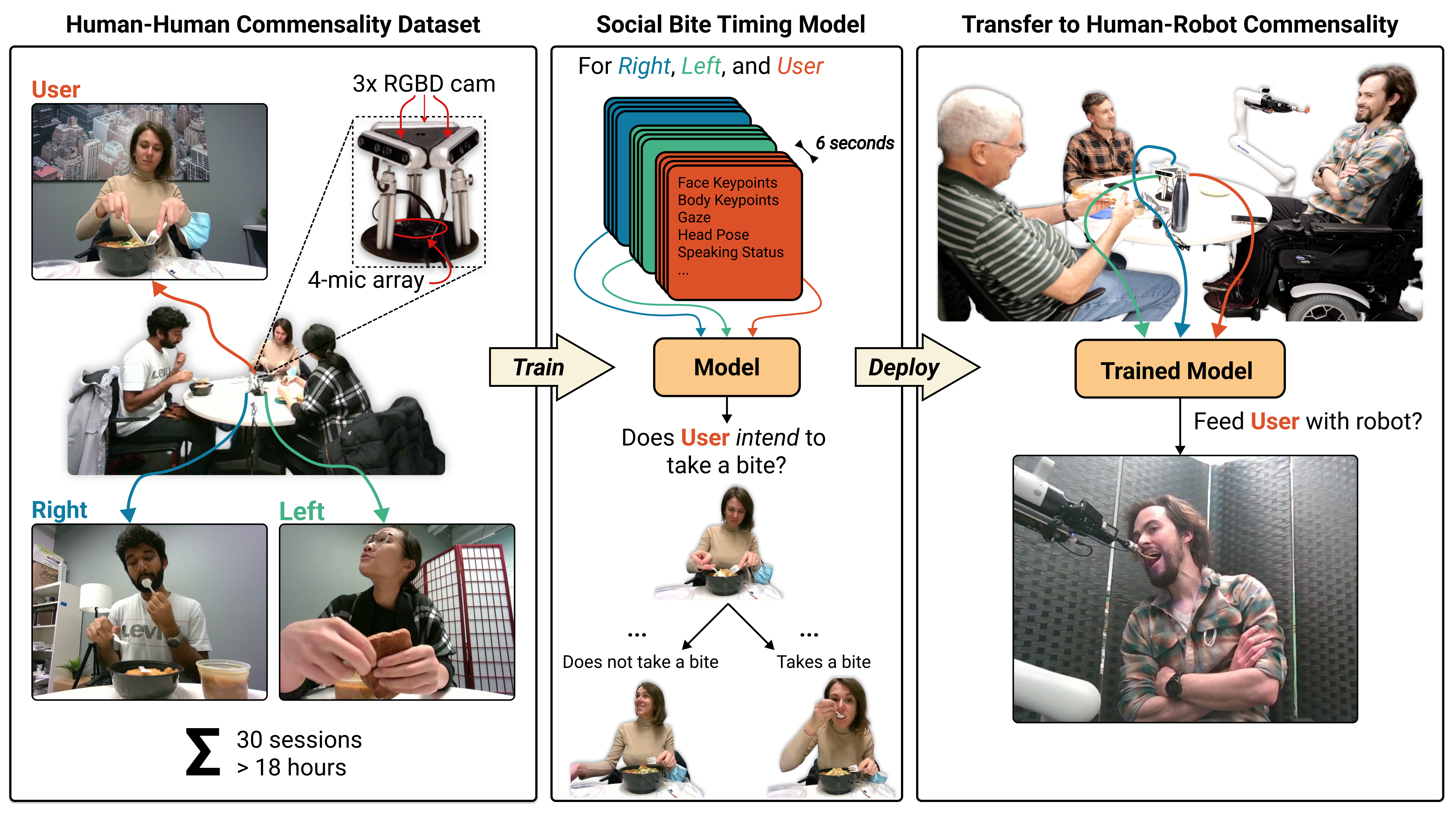}\vspace{-2mm}
	\caption{Our bite timing prediction workflow: 
	\textbf{(Left)} Human-Human Commensality Dataset collection:
	We record audio and video of participants eating food in triads.
	\textbf{(Middle)} Our Social Nibbling NETwork (SoNNET) learns to predict whether a user intends to take a bite based on a 6-second window of social signals. 
	\textbf{(Right)} We conduct a  social robot-assisted feeding user study by deploying a variation of SoNNET on a robot.
	We refer to the \textcolor[rgb]{0.867,0.318,0.161}{\textbf{User}} also as a \textcolor[rgb]{0.867,0.318,0.161}{\textbf{Target user}}.
	}
	\label{fig:introduction:main_diagram}
    \vspace{-.2in}

\end{figure}

We developed an intelligent autonomous robot-assisted feeding system that uses multimodal sensing to feed people in dynamic social dining scenarios.
We collected a novel audio-visual Human-Human Commensality Dataset (HHCD) capturing human social eating behaviors. 
Using this data, we then trained multimodal machine learning models to predict bite timing in human-human commensality. 
We explored how our models trained on human-human commensality scenarios performed in a human-robot commensality setting and evaluated them in a user study. The overall workflow is shown in Fig.~\ref{fig:introduction:main_diagram}. 
We made algorithmic and experimental design decisions by consulting with care recipients, caregivers, and occupational therapists.
We find that bite timing prediction improves when our model accounts for social signaling among diners, and such a model is preferred over a manual trigger and a fixed-interval trigger. Our main contributions include:
\begin{itemize}[nosep,leftmargin=*] 
	\item A SOcial Nibbling NETwork (SoNNET) which captures the subtle inter-personal social dynamics in human-human and human-robot groups for predicting bite timing in social-dining scenarios.
	\item Methods that can successfully transfer bite timing strategies learned from human-human commensality cues to human-robot commensality situations, which we evaluate in a user study with a robot in 10 triadic human groups.
	\item A socially-aware robot-assisted feeding system that extends our capacity to feed people in solitary settings to groups of people sharing a meal.
	\item An analysis of various social and functional factors that affect human feeding behaviors during human-human commensality. 
	\item A novel \emph{Human-Human Commensality Dataset (HHCD)} containing multi-view RGBD video and directional audio recordings capturing $30$ groups of three people sharing a meal.

\end{itemize}

\section{Human-Robot Commensality}
Eating is a complex activity that requires the sensitive coordination of several motor and sensory functions. Anyone who has fed another knows that feeding, particularly \emph{social feeding} where a person is being fed in a social setting, is a delicate \emph{dance} of multimodal signaling (via gaze, facial expressions, gestures, and speech, to name a few). Research on \textit{commensality}, the practice of eating together, has highlighted the importance of the social nature of eating for social communion, order, health, and well-being~\cite{jonsson2021commensality}. As a consequence, digital commensality has focused on understanding the role of technology in facilitating or inhibiting the more pleasurable social aspects of dining~\cite{spence2019digital}.

When a person relies on assisted feeding, meals require that patient and caregiver coordinate their behavior~\cite{athlin1990caregivers}. To achieve this subtle cooperation, the people involved must be able to initiate, perceive, and interpret each other's verbal and non-verbal behavior. The main responsibility for this cooperation lies with caregivers, whose experiences, educational background, and personal beliefs may influence the course of the mealtime \cite{athlin1987interaction}. Our goal in this work is to understand the rhythm and timing of this \emph{dance} to enable an automated feeding assistant to be thoughtful of when it should feed the user in social dining settings. We introduce the concept of \textbf{Human-Robot Commensality} at the intersection of commensality and robot-assisted feeding in social group settings. 




Our research is motivated by the key insight that bite timing strategies that take into account ever-changing social signals and group dynamics can lead to a seamless human-robot collaboration in social dining scenarios.
Fueled by this insight, we believe a feeding device that takes the initiative and offers bites proactively during the meal at times when a bite is likely to be desired will create a more seamless dining experience than a device that requires the user to initiate bites. 
Herlant \cite{herlant2018algorithms} designed an HMM to predict bite timing in dyadic robot-assisted feeding. However, her model only considered the social cues of the user.
Bhattacharjee et al.~\cite{bhattacharjee2020more} found users preferred less intrusive interfaces in a social dining scenario, specifically a web interface over a voice interface.
Our work aims to build non-intrusive bite timing strategies by focusing on learning when to feed a user in triadic scenarios while using implicit social features from all diners.

Particularly, bite timing is important because the consequences of presenting a bite to the diner earlier than expected is poorly tolerated. 
This can include an interruption to conversation or to finishing chewing the prior bite. The consequences of presenting a bite later than desired can include frustration towards the robot and disruption of the natural flow of conversation during the meal.
Parallels can be drawn to interruptibility research on finding the most appropriate timing to probe a user.
Researchers have found that people performed best on a task if interruptions were mediated rather than timed immediately or on scheduled intervals \cite{mcfarlane2002comparison,czerwinski2000instant}, often mediated based on modeling contextual and social factors \cite{pielot2017beyond, takemae2007estimating, banerjee2018effects, wu2021learning}.

A socially-aware robot-assisted feeding system should be designed such that if needed, the user should be able to communicate these intentions via multiple different modalities such as body language, gaze, or speech.
These various modalities have been found to be effective in modeling social interactions \cite{wang2021multi, muller2018robust, park20123d, joo2019towards}.
Capturing these natural social interactions in computational models are likely crucial to provide accurate bite timing without distracting users from the social ambiance.

\section{Problem Formulation}\label{sec:problem_formulation}
The objective of the bite timing prediction problem in robot-assisted feeding with a single diner is to predict the timing of \textit{when} this user will take a bite of food by capturing their signals $\mathbf{U}$ such as voice, body gestures, head movements or speaking status. 
We define the proper timing for when a robot should feed as when the user \textit{intends} to take a bite of food.
It takes input signals $\mathbf{U}(t_0 : t)$ from time $t_0$ to time $t$ and learns a function $\mathcal{F}(\mathbf{U})$ to predict a Boolean $y(t+h) = \mathcal{F}\left(\mathbf{U}\left(t_{0}: t\right)\right)$, 
which indicates whether the user intends to take a bite in the time horizon $h$ and trigger a bite transfer at time $t+1$. 
When a person lifts their fork off the plate to eat, they intend to take a bite of food, where this time horizon $h$ is the time it takes to transfer the food to their mouth from their plate.



In this paper, we consider a social variant of the bite timing prediction problem where a user is interacting with two co-diners. 
Our goal is to predict the timing of a user to take a bite of food based on the social cues within the interaction.
From an initial time $t_0$ to time $t$, the user receives social signals $\mathbf{L}(t_0:t)$ and $\mathbf{R}(t_0:t)$ from their left and right conversational co-diners, respectively. 
Given these external social signals and the target user's own history of signals $\mathbf{U}(t_0:t)$, we aim to predict $y$. 
We note that it may not always be possible to track the same set of features for a user and their co-diners. 
Therefore, for some time range $k=t-t_0$ and feature dimensions $n,m$ for the user and co-diners respectively, $\mathbf{U} \in \mathbb{R}^{k \times n}$ while $\mathbf{L},\mathbf{R} \in \mathbb{R}^{k \times m}$, where $n$ does not necessarily equal $m$.
The function to learn is:
$$
y(t+1) =  \mathcal{F}\left(\mathbf{U}\left(t_{0}: t\right), \mathbf{L}\left(t_{0}: t\right), \mathbf{R}\left(t_{0}: t\right)\right)
$$

\section{Model: SOcial Nibbling NETwork (SoNNET)}\label{sec:methodology}

We present the SOcial Nibbling NETwork (SoNNET) that predicts when a user has the intention to eat based on various social signals. We selected features to represent both human eating and social behavior: bite features, which include the number of bites taken so far and the time since the last bite of food $b \in \mathbb{R}^2$, a diner's gaze and head pose direction $d \in \mathbb{R}^4$, binary speaking status $s \in \{0,1\}$, and face and body keypoints $o \in \mathbb{R}^{168}$ from OpenPose~\cite{cao2018openpose}. 
We note that, in our case, the bite features $b$ are computed only for the user and not the co-diners, since we do not estimate in real-time whether a co-diner is taking a bite of food.
Thus, for a time interval $k=t-t_0$, these features are temporally stacked to construct the input signals $\mathbf{U} \in  \mathbb{R}^{k \times 175}, \mathbf{L} \in  \mathbb{R}^{k \times 173}, \mathbf{R} \in  \mathbb{R}^{k \times 173}$ for the user, left co-diner, and right co-diner, respectively.

\begin{figure}[t]
    \centering
    \includegraphics[width=.95\textwidth]{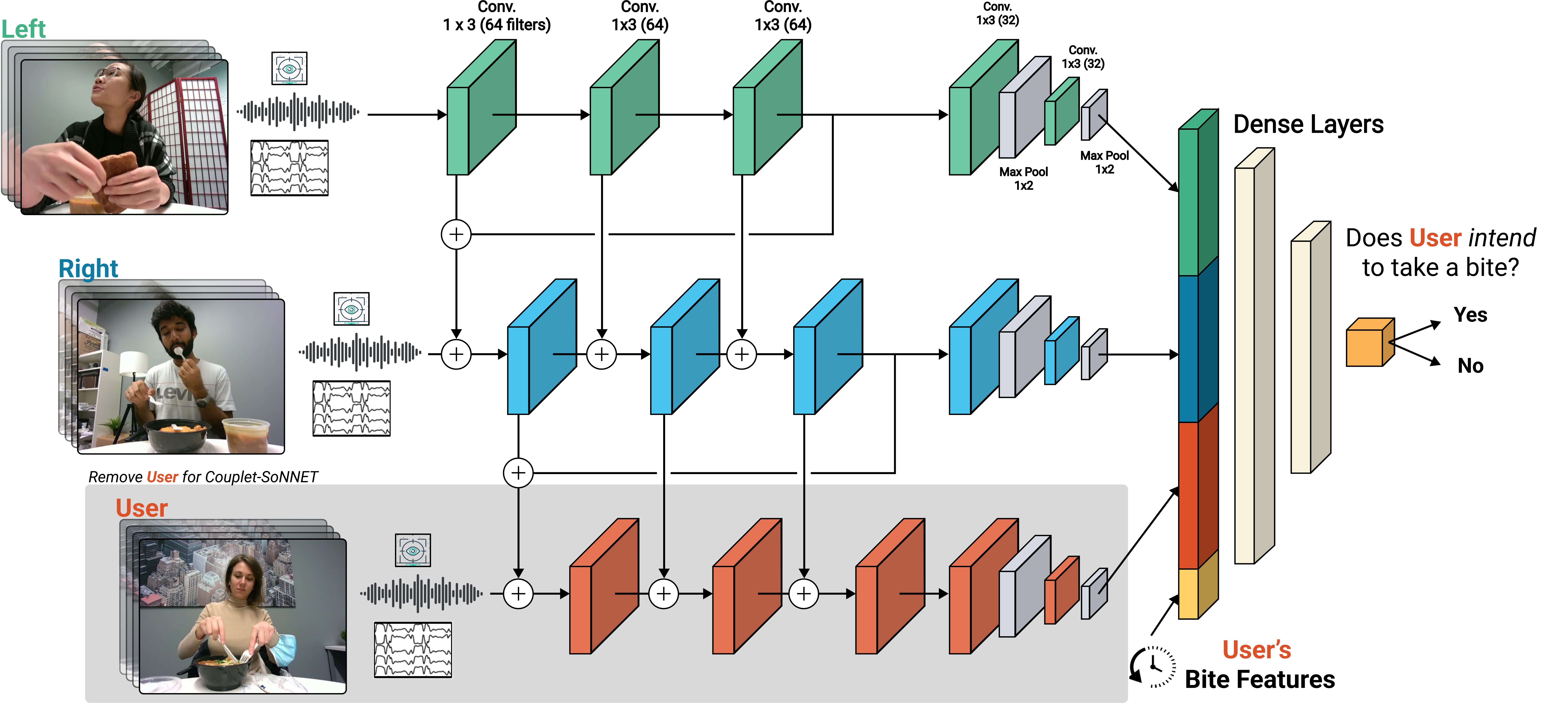}\vspace{-1mm}
    \caption{\textbf{Triplet-SoNNET} contains three interacting channels for features of a \textcolor[rgb]{0.867,0.318,0.161}{\textbf{target user}} and two co-diners. Each channel concatenates the input of time, gaze, speaking and skeleton features from each single diner. 
    \textbf{Couplet-SoNNET} eliminates all features from the \textcolor[rgb]{0.867,0.318,0.161}{\textbf{target user}} by dropping the last channel; however, it continues to use the user's bite features. Batch normalization layers are not shown in the figure.}
    \label{fig:methods:sonnet}
    \vspace{-7mm}
\end{figure}

Recently, convolutional neural networks (CNNs) have demonstrated significant success for multi-channel time series classification from various kinds of signals \cite{zhao2017convolutional, yang2015deep, liu2018time}. 
Wu et al.~\cite{wu2021learning} proposed PazNet: a multi-channel deep convolutional neural network which is able to handle inputs of different dimensions. 
PazNet is designed to predict the interruptibility of individual drivers.
However, the information of different channels is not shared, and it lacks ability to capture social interactions among multiple people.

We design the Social Nibbling NETwork (SoNNET), a new model architecture which follows a multi-channel pattern allowing multiple interconnected branches to interleave and fuse at different stages. 
We create input processing channels for each diner, then add interleaving tunnels between each convolutional module and adjacent branches.
The information capturing visually-observable behaviors between the diners is allowed to flow between the frames and channels. 
We conjecture that our model will learn a socially-coherent structure, allowing the model to implicitly represent the diners in an embedding space. 
Therefore, each channel has the same structure but does not share the same weight parameters. 
To help capture informative features, we performed dimension-reduction after the interleaving components using max pooling layers and $1 \times 1$ convolutional layers. 
These per-diner channels are concatenated and then followed by two dense layers for classification, which decides whether the user intends to feed or not.
For SoNNET, the range between $t$ and $t_0$ is six seconds. The social signals in this range are used to predict a user's bite intentions.

\textbf{Triplet-SoNNET.} 
For modeling the bite-timing prediction of three users with no mobility limitations, we propose Triplet-SoNNET which uses social signals from the left and right co-diners $\mathbf{L}, \mathbf{R}$ and signals from the user $\mathbf{U}$. 
Depicted in Fig.~\ref{fig:methods:sonnet}, Triplet-SoNNET ensures that the features from other co-diners $\mathbf{L},\mathbf{R}$ interleave into the target user's features $\mathbf{U}$. 

\textbf{Couplet-SoNNET.} 
To run Triplet-SoNNET in a robot-assisted feeding setting, there would be a distribution shift in the kinds of signals a target user outputs.
Our goal is to feed people with mobility limitations while they are engaged in social conversations.
The features from someone self-feeding are inherently different from someone using a robot-assisted feeding system.  
In the case of body pose, a target user with C3-C5 SCI would be largely still, which is different from the training data.
Our Human-Human Commensality Dataset consists of adult diners with no mobility limitations, thus applying a trained Triplet-SoNNET model to robot-assisted feeding of a user with mobility limitations would be out-of-distribution.
Although our target users with C3-C5 SCI cannot move their arms to feed themselves, there is still a large spectrum of severity in mobility limitations depending on the users' conditions. From our discussions with key stakeholders, caregivers, and occupational therapists we design,
Couplet-SoNNET, where we ignore most social signals from the target user by removing the last channel in Triplet-SoNNET. 
Therefore, the intention to feed $y(t+1) =  \mathcal{F}\left(\mathbf{U}_b\left(t_{0}: t\right), \mathbf{L}\left(t_{0}: t\right), \mathbf{R}\left(t_{0}: t\right)\right)$, where $\mathbf{U}_b \in \mathbb{R}^{k \times 2}$ are the user's bite features for $k=t-t_0$.
The user's bite features, such as the time since the last bite and the number of bites since the onset of the feeding activity, are the only social signals from the target user. Additional discussion on this design choice can be found in App.~\ref{couplet_sonnet_design}.

\section{Human-Human Commensality Dataset (HHCD)}\label{sec:dataset}

We introduce a novel Human-Human Commensality Dataset (HHCD) of three participants with no mobility limitations eating in a social scenario. 
We used this dataset to develop models that predict a diner's intention to take a bite of food while taking into account subtle social cues. 
We deployed the trained models in a social robot-assisted feeding setting where one diner is fed by a robot.
Beyond predicting bite timing, we are excited for the robot learning community to find other interesting challenges within our dataset that leverage understanding social dynamics.



\textbf{Data Collection Setup.}\label{sec:dataset:collection}
We recruited 90 people among our Institution-affiliated fully-vaccinated students, faculty, and staff to eat a meal in a triadic dining scenario. 
Each participant was 18+ years old and took part in the study only once. 
The study setup is illustrated in Fig.~\ref{fig:introduction:main_diagram} (left). 
There are three cameras (mutually at 120$^\circ$) in the middle of the table, each capturing one participant, 
and a fourth camera capturing the whole scene. 
All four cameras are Intel RealSense Depth Cameras D455~\cite{realsense}. 
The scene audio was captured by a ReSpeaker Mic Array v2.0~\cite{micarray} placed in the middle of the table. The ReSpeaker microphone array has four microphones arranged at the corners of a square and estimates the direction of sound arrival. 
For the study setup measures, see App.~\ref{sec:appendix:dataset:measures}.

Participants were free to bring any kind of food and any utensil with them.
They could also bring a drink (some drank from a cup, others from a bottle or both, with or without a straw) and were provided with napkins. 
Before the study, each participant was asked to fill in a pre-study questionnaire about their demographic background, relationship to other participants, and social dining habits. 
The experimenter then asked them to eat their meals and have natural conversations. 
At this point, the experimenter started the recording and left the room. 
When \textit{all} three participants finished eating or after 60 minutes have passed, whichever was earlier, the experimenter stopped the recording and asked participants to fill in a post-study questionnaire about their dining experience. 
The specific questions asked in both pre/post-study questionnaires can be found in App.~\ref{sec:appendix:dataset:collection}. 
The study was approved by Cornell's IRB.

\textbf{Data Annotation.}\label{sec:dataset:annotation}
We annotated each participant's video based on their interactions with food, drink, and napkins.  
In particular, we annotated \textit{food\_entered}, \textit{food\_lifted}, \textit{food\_to\_mouth}, \textit{drink\_entered}, \textit{drink\_lifted}, \textit{drink\_to\_mouth}, \textit{napkin\_entered}, \textit{napkin\_lifted}, \textit{napkin\_to\_mouth}, and \textit{mouth\_open} events. We chose these events as they are key transition points during feeding.
We spent 151 hours annotating and used the ELAN annotation tool~\cite{elan}. 
We assigned the annotation value $\in$ \{\textit{fork}, \textit{knife}, \textit{spoon}, \textit{chopsticks}, \textit{hand}\} based on the utensil performing the food-to-mouth handover. 
While annotating, we also noted down per-participant food types and observations of interesting behaviors.
All annotation types with detailed rules are provided in App.~\ref{sec:appendix:dataset:annotation}.



\textbf{Data Statistics.}\label{sec:dataset:stats}
There were 56 female and 34 male participants, and their ages ranged 18-38 ($\mu=22$, $\sigma=3$) years. 
Session durations ranged 21-55 ($\mu=37$, $\sigma=9$) minutes and 1 session was at breakfast, 10 at lunch, and 19 at dinner time. 
For additional dataset statistics, see App.~\ref{sec:appendix:dataset:stats}. 

For a summary of all available data in the dataset and its detailed analysis, see App.~\ref{sec:appendix:dataset}.
For the purposes of this work, we only consider bite features, speaking status, gaze and head pose, and body and face keypoints. 
\section{Model Evaluation on Human-Human Commensality Dataset}\label{sec:experiments}

\setlength{\tabcolsep}{1.5pt}
\begin{table}[!t]
    \centering
    \begin{minipage}{.45\textwidth}

    \centering
    \caption{Ablation study on different modalities from various data sources. 
    We use average over LOSO cross-validation.}
    \scalebox{0.9}{
    \begin{tabular}{@{}lccccc@{}}
    \toprule
    Method          & \multicolumn{1}{c}{Acc.} & \multicolumn{1}{c}{Prec.} & \multicolumn{1}{c}{Rec.} & \multicolumn{1}{c}{F1} & nMCC                       \\ \midrule
Triplet-SoNNET  & \textbf{0.820}              & 0.861                        & 0.871                     & 0.862                 & \textbf{0.772}            \\
\; - Speaking Status      & 0.816                        & \textbf{0.864}               & 0.863                     & 0.856                 & 0.771                     \\
    \; - Gaze \& Head Pose & 0.815                            & 0.863                            & 0.863                          & 0.856                      & 0.769                          \\
\; - Bite Features & 0.781                       & 0.832                        & 0.855                     & 0.834                 & \multicolumn{1}{r}{0.727} \\
\; - Body \& Face     & 0.820                       & 0.854                        & \textbf{0.886}            & \textbf{0.865}        & \multicolumn{1}{r}{0.771} \\ \bottomrule
    \end{tabular}
    \label{tab:ablation}

    }
    \end{minipage}\hspace{9mm}
    \begin{minipage}{.45\textwidth}\vspace{2.05mm}
        \centering
        \caption{Model comparison on LOSO cross-validation over 29 sessions.}

        \scalebox{0.9}{
        \begin{tabular}{@{}lccccc@{}}
        \toprule
        Method          & \multicolumn{1}{c}{Acc.} & \multicolumn{1}{c}{Prec.} & \multicolumn{1}{c}{Rec.} & \multicolumn{1}{c}{F1} & nMCC                       \\ \midrule
        Always Feed         & 0.72                       & 0.72                        & 1                          & 0.83                 & 0.5           \\ \midrule
        Linear SVM (SGD)            & 0.68                       & 0.82                        & 0.77                     & 0.74                 & 0.64           \\
        Triplet-TCN    & 0.82                        & 0.82                        & \textbf{0.96}            & \textbf{0.88}        & 0.72           \\
        Triplet-SoNNET & \textbf{0.82}              & \textbf{0.86}               & 0.87                     & 0.86                 & \textbf{0.77}  \\ \midrule
        Couplet-TCN    & 0.73                        & 0.73                        & \textbf{0.98}            & 0.83                 & 0.55          \\
        Couplet-SoNNET & \textbf{0.76}              & \textbf{0.78}               & 0.96                     & \textbf{0.85}        & \textbf{0.66} \\ \bottomrule
        \end{tabular}
        \label{tab:model_comp}

        }
    \end{minipage}
    \vspace{-4mm}
\end{table}

In this section, we evaluate Triplet- and Couplet-SoNNET against other models on the HHCD. 
In particular, we compare against a regularized linear SVM trained with SGD to evaluate performance of linear classifiers.
We also consider a Temporal Convolution Network (TCN) \cite{lea2016temporal, lea2017temporal}, which uses causal convolutions and dilations to represent temporal data. 
TCNs have been found to perform better than LSTMs and GRUs on temporal anomaly detection \cite{he2019temporal} and robot food manipulation tasks \cite{bhattacharjee2019towards}, therefore they would provide a strong baseline to compare our models to.
We also perform an ablation study to investigate the importance of various modalities. 
Implementation details about baseline models, SoNNET, and training procedure can be found in App.~\ref{sec:appendix:model_exps}.

For training, we use 6811 \textit{food\_lifted} annotations as positive training labels since they precede an actual bite of food and indicate an intention to eat.
We use a time interval of $k=t-t_0=6$~seconds because it takes roughly 6 seconds for the robot to move from its wait position to feeding the user.
Since bite actions are sparsely distributed over time, we select 2486 6-second clips as negative samples that are in the middle of two \textit{food\_lifted} annotations. 
All reported models are trained with leave-one-session-out (LOSO) cross-validation to evaluate generalizability to new groups of people. Due to an issue with recording, we train over 29 sessions if speaking status features are used.

The user's bite features $b \in \mathbb{R}^2$ (time since last bite and the number of bites eaten since the start) are indicators of eating rate.
To ensure this feature is not dominated by higher dimensional features, we scale the size of the input by $\gamma$. This hyperparameter $\gamma$ scales $b\in \mathbb{R}^2 \rightarrow b \in \mathbb{R}^{2\gamma}$. 
We selected $\gamma=100$ after a grid search over the training set on the TCN and SoNNET models.

\textbf{Evaluation Metrics.} A high recall indicates that our model can reliably feed when it should.
In contrast, a high precision indicates that a model tends to be stricter in deciding when to feed.
Due to our dataset imbalance, the average accuracy across 29 sessions for a model that predicts it should always feed is 71.56\%. 
This classifier achieves perfect recall, and relatively high precision, causing the model to have a high F1 score.
It is clear that given this class imbalance, a high F1 score poorly represents the capabilities of this model.
To evaluate our model effectively, we consider the normalized Matthews Correlation Coefficient (nMCC) in addition to F1 score, precision, recall, and accuracy. 
Unlike F1 score, nMCC considers the size of the majority and minority classes, and can only produce high scores if a classifier is able to make correct predictions for a majority of both the negative and positive classes \cite{chicco2020advantages}. 
A value of 0.5 indicates random prediction, while 0 is inverse prediction and 1 is perfect prediction.

\textbf{Effects of Modality.} We are interested in investigating features that are the most informative for designing a good bite timing predictor in social dining.
We perform a feature ablation study on the Triplet-SoNNET model, as shown in Table~\ref{tab:ablation}.
We selectively remove feature streams, such as body and face data from OpenPose, gaze and head features from RT-GENE, speaking status signals, and the user's bite features. 
We find that users' bite features such as the time since last bite and the number of bites are important, as accuracy drops drastically without them.
Intuitively, we believe this feature is important because a user's bite features are a proxy for their level of eating rate. 
We also see that without body and face features, F1 and recall slightly increase. 
This could be due to the fact these data streams are noisy; however, as indicated by the lower accuracy and nMCC when removing OpenPose features, these features are useful. 

\textbf{Effects of Model Type.} Table~\ref{tab:model_comp} shows the outcomes of various model comparisons when trained using LOSO. 
We compare performance of Triplet-SoNNET against a linear SVM and TCN trained on all three diners. We call this TCN a Triplet-TCN.
Triplet-TCN has all the diners' features concatenated per-timestep, and we compare this result to Triplet-SoNNET.
We find that Triplet-SoNNet achieves higher accuracy and nMCC compared to all other models; however, it performs worse on recall and F1 score compared to Triplet-TCN. 
In our scenario, we want to ensure that the robot feeds when it should and does not feed when it should not. A bite prediction model that overfeeds or underfeeds is not ideal. 
A high nMCC balances the roles of recall and precision and better represents whether a classifier should or should not feed. Therefore, for our scenarios, Triplet-SoNNET is a more effective predictor of bite timing than other models trained on all three diners.

\textbf{Effects of Social Scenario.} We are interested in comparing the ability of models to learn social behaviors using only two co-diners' features rather than having full observability.
We compare Couplet-SoNNET to a similarly-named Couplet-TCN trained on two co-diners' features and a user's bite features.
As expected, Couplet-TCN and Couplet-SoNNET perform worse than their Triplet- counterparts, with Couplet-TCN being close to random prediction with an nMCC of $0.5539$ 
while Couplet-SoNNET has an nMCC of $0.6648$. 
We find that Couplet-SoNNET performs better than Couplet-TCN. 
This result reveals Couplet-SoNNET is able to understand social signals better than a predictor that always feeds. This implies that it is possible to predict the behavior of a user using only their co-diner information, which indicates that there is social coordination in human-human commensality. These findings also suggest that social signals were captured by the interleaving structure of the SoNNET models.

\section{Transferring from Human-Human to Human-Robot Commensality}\label{sec:user_study}
Our objective is to develop a bite timing strategy for a robot that feeds a user in a social dining setting.
We design a study where users evaluate the effect of different bite timing strategies on their overall social dining experience.
To simulate robotic caregiving scenarios for people with upper-extremity mobility limitations, we instructed users to not move their upper body. 
This study was approved by Cornell's IRB.

\begin{figure}[t]
    \centering
    \begin{tabular}{l l}
        \parbox{.33\linewidth}{\includegraphics[width=1.7in]{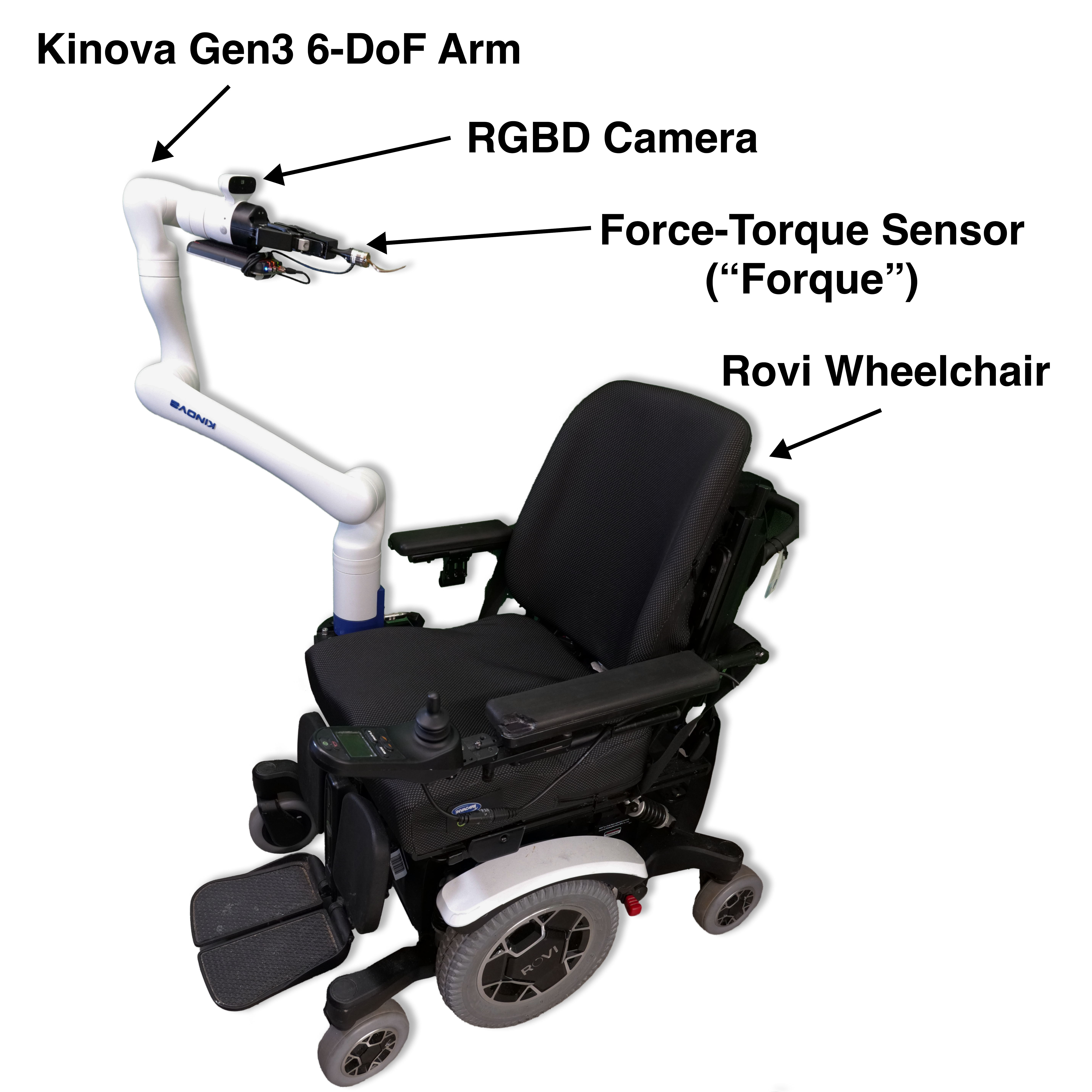}}\hspace{-4mm}
        & \parbox{.33\linewidth}{\includegraphics[width=1.7in]{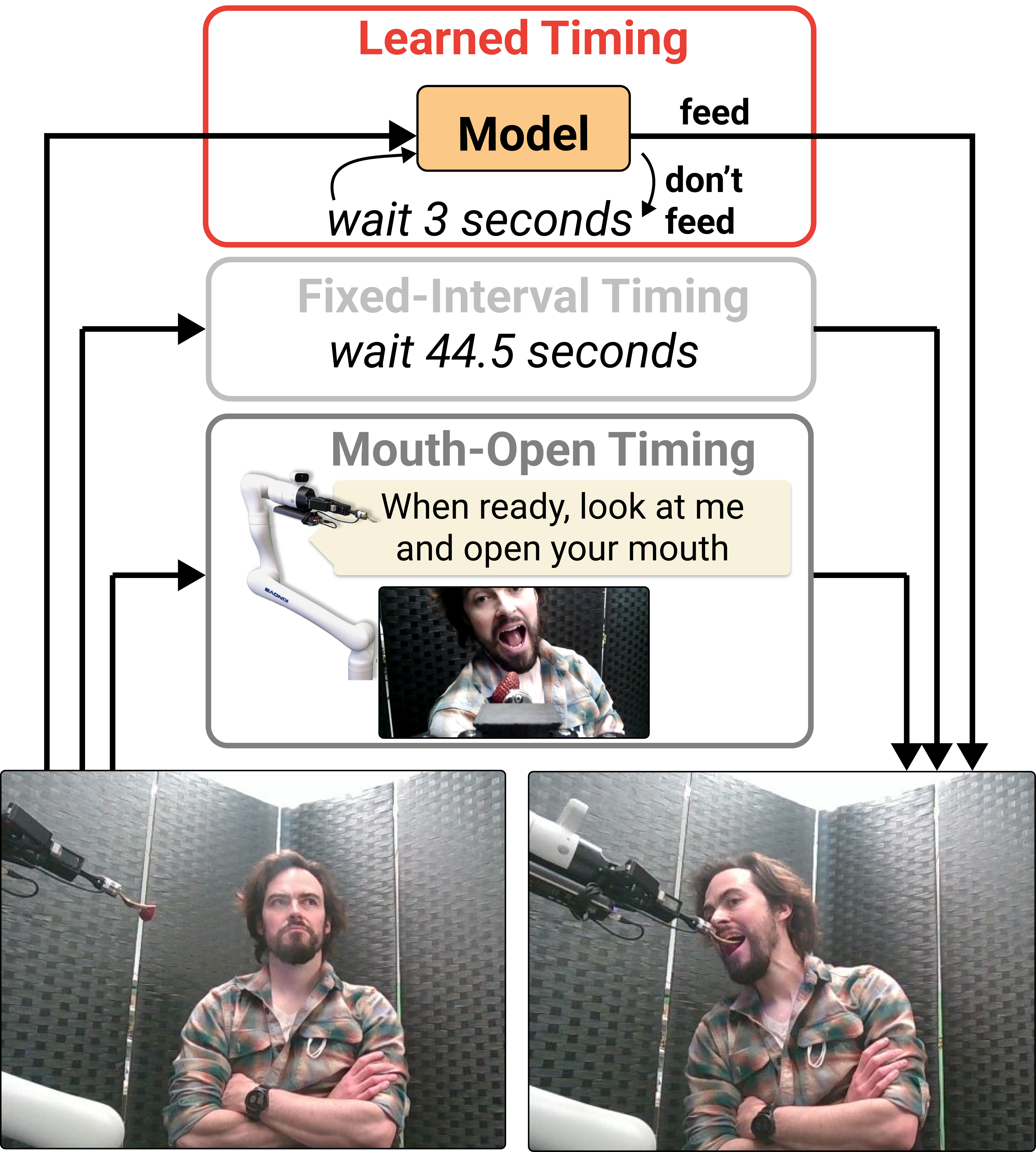}} 
        \parbox{.33\linewidth}{\hspace{-2mm}\includegraphics[height=1.0in]{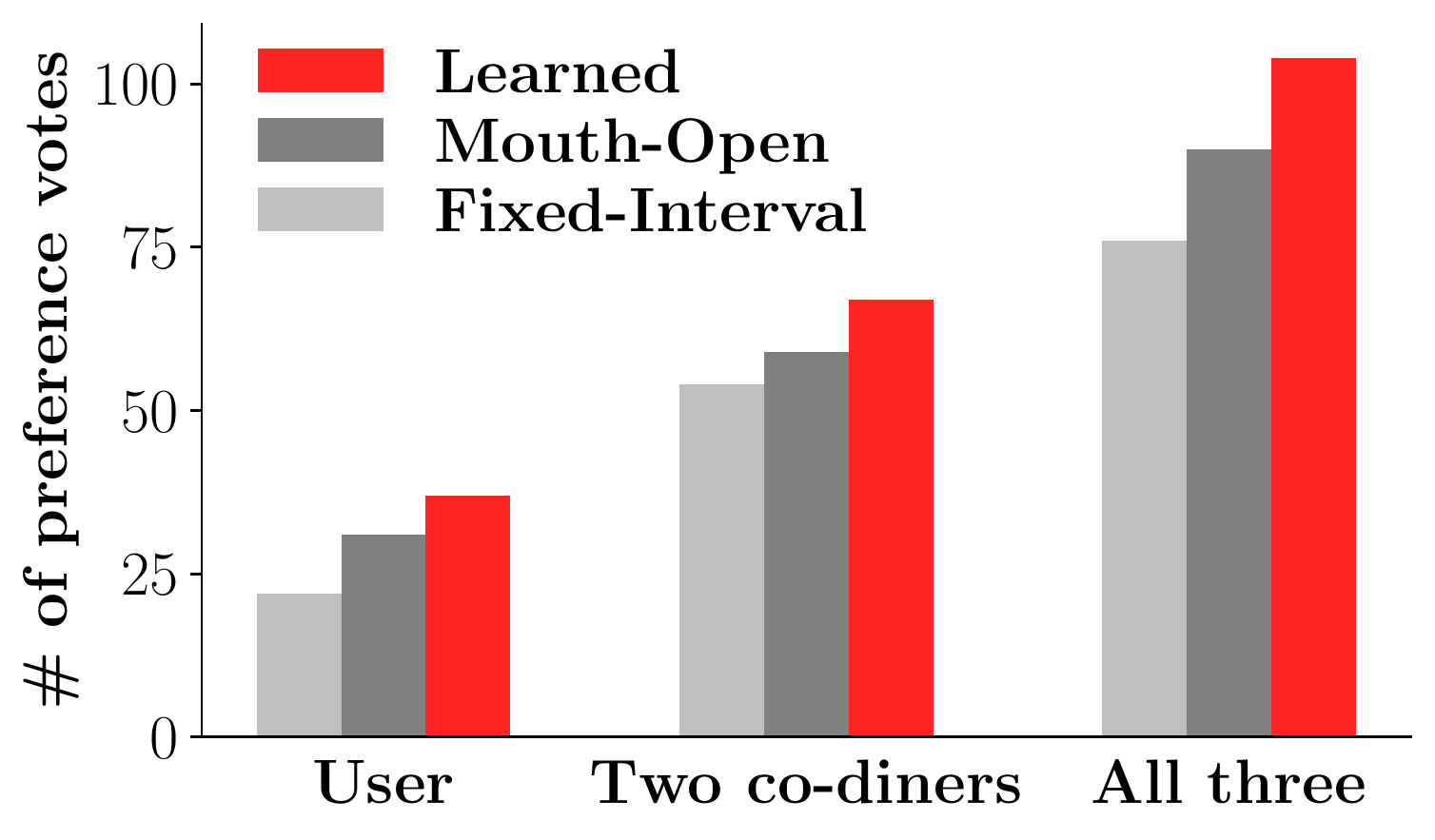}\\ \includegraphics[height=1.1in]{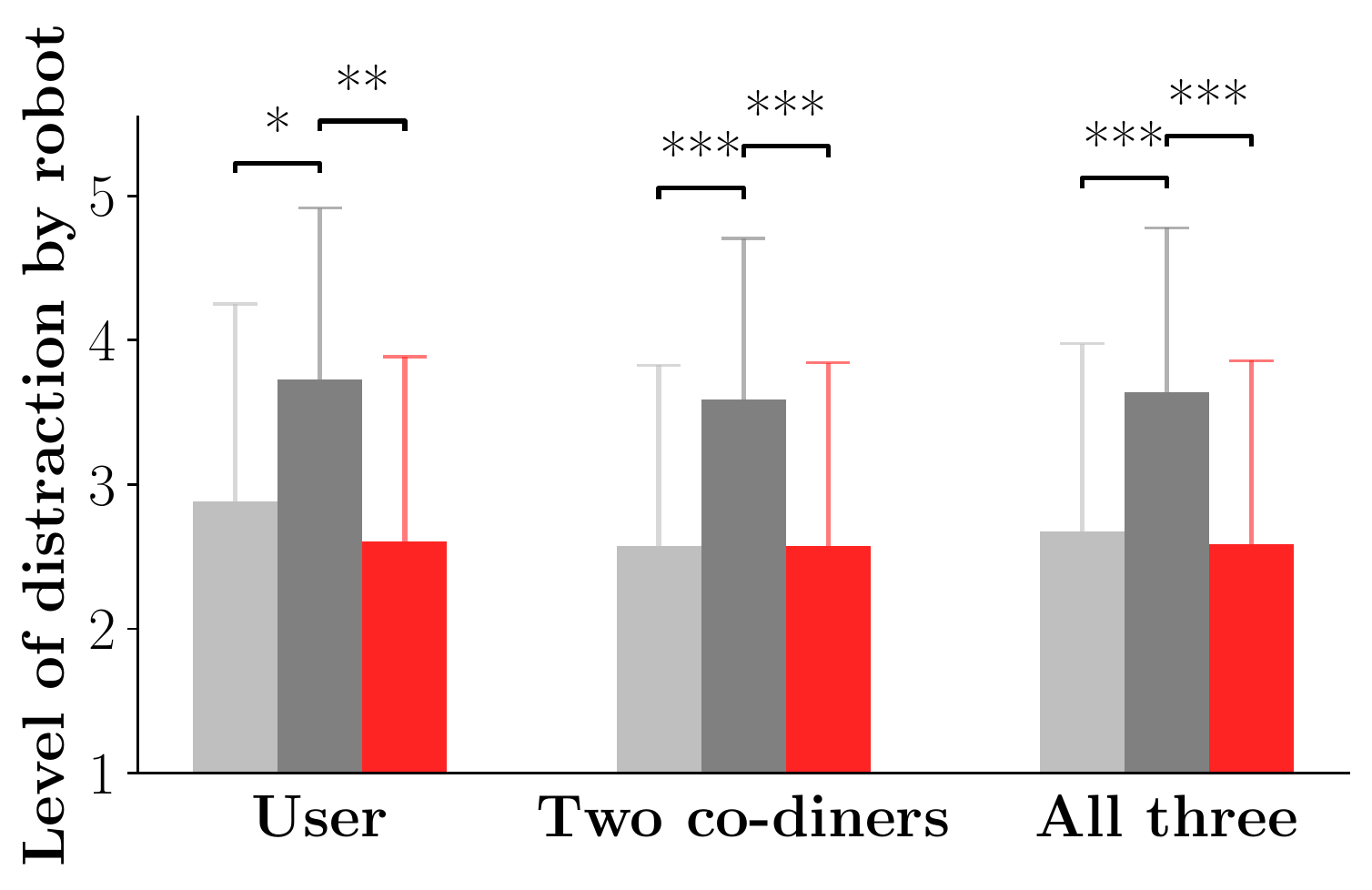}}\\
    \end{tabular}
    \vspace{-2mm}
    \caption{
    \textbf{Left:} We use a 6-DoF Kinova Gen3 robotic arm \cite{gen3} on Rovi wheelchair \cite{rovi}.
    \textbf{Middle:} User study conditions/bite timing strategies: Learned, Fixed-Interval, and Mouth-Open Timings. 
    \textbf{Top right:} Preferences for bite timing strategies 
    rated by users, two co-diners, and all three diners. 
    \textbf{Bottom right:} 
    Level of distraction by the robot perceived by users, two co-diners, and all three diners on a Likert scale 1-5 (agreement with ``I felt distracted by the robot''), 
    for each bite timing strategy.
    $*, **, ***$ denote statistically significant differences with $p_{0.05}, p_{0.005}, p_{0.0005}$ respectively.
    }
    \vspace{-.2in}
    \label{fig:exp}
\end{figure}

\textbf{Experimental Setup.}
We evaluate a learned bite timing strategy against two baseline bite timing strategies inspired by our conversations with care recipients, occupational therapists, and caregivers who told us how they know when to feed. The strategies are further depicted in Fig.~\ref{fig:exp}:


\begin{enumerate}[nosep,leftmargin=*]
  
    \item \textbf{Learned Timing.} This social, fully autonomous bite timing strategy feeds based on our Couplet-SoNNET model's output. 
    We sample this model every three seconds with the last six seconds of preprocessed features at a rate of 15 frames per second.
    This approach takes into account the social context. Since we want to evaluate the generalization performance, we train Couplet-SoNNET on 80\% of the HHCD data and use the remaining 20\% of HHCD data to select early-stopping criteria. 
    
    \item \textbf{Fixed-Interval Timing.} This fully autonomous bite timing strategy feeds every $44.5$ seconds, which is a scaled average time a robot should take to feed a user after it has picked up a food item. To derive this value, we first find the appropriate scaling factor between human motion from the HHCD and robot motion. We note the average time for a human from the \emph{food\_entered} transition to \emph{food\_lifted} transition is $9.9$ seconds. The robot end-effector motion is not designed to match the human speed but rather to be perceived as safe and comfortable to a user being fed. We find the equivalent key transitions for the robot to be $5x$ slower than a human. Since we define the intention to take a bite as when the food is lifted, the robot should take $49.5$ seconds to feed a user after picking up a food item. Given the robot takes roughly $5$ seconds to move to its wait position after picking the food, the robot waits $44.5$ seconds. Further details about this wait-time can be found in  App.~\ref{sec:appendix:user_study:conditions}.
    
    \item \textbf{Mouth-Open Timing.} This partially autonomous bite timing strategy feeds only when the user prompts the robot by opening their mouth. 
    The target user is prompted each time by the robot saying "When ready, look at me and open your mouth".
    This approach gives the user explicit control of when the robot should feed \cite{bhattacharjee2020more}.

\end{enumerate}



The robot user is seated on a wheelchair mounted with a Kinova Gen3 6-DoF arm \cite{gen3}, which is used to feed the participant (Fig.~\ref{fig:exp}, left). For discussion of our bite timing strategies, the use of voice prompting, and implementation details of the robot study, see App.~\ref{sec:appendix:user_study:conditions}-\ref{sec:appendix:user_study:procedure_details}.


\textbf{Experimental Procedure.}
In this study, participants are seated in a similar setup as that used for HHCD data collection in Sec.~\ref{sec:dataset:collection}. 
All participants were asked to bring their own food, and each group chose who would be fed by the robot. 
We recruited 30 participants over 10 sessions. There were 16 female and 14 male participants, and their ages ranged from 19-70 ($\mu=27$, $\sigma=9$) years.

A single trial consists of bite acquisition, followed by one of the three bite timing strategies, then bite transfer.
For bite acquisition, the robot alternates feeding the user cantaloupes and strawberries. We chose these fruits due to their high acquisition success rates \cite{feng2019robot}. We used the bite acquisition strategies and bite transfer strategies from~\cite{gallenberger2019transfer, belkhale2021balancing}.
All participants take a survey after each trial, which administers a forced-choice question on the participants' preferences between the previous and current conditions. 
Each pair of comparisons between any two conditions occurs three times, leading to ten trials.
The condition orderings are counterbalanced over ten trials. 
Additionally, we ask participants whether they felt the robot fed them too early, on-time, or too late. 
The experiment questionnaire after each trial further includes questions about bite timing appropriateness, distractions due to the robot, ability to have natural conversations, ability to feel comfortable around the robot, as well as system reliability and trust in the robot~\cite{jian2000foundations}.
For details on user study questionnaires see App.~\ref{sec:appendix:user_study:questionnaires}. 
%
%
%
To avoid interruptions in social conversations due to the presence of a robot in human groups, we provide the participants with a list of questions (see App.~\ref{sec:appendix:user_study:prompts}), which they could optionally use to help get the conversation started at each trial, similarly to previous work~\cite{herlant2018algorithms}. 

\textbf{Results and Discussion.}\label{sec:user_study:results}
As shown in Fig.~\ref{fig:exp} (top right), users and co-diners preferred the Learned strategy for bite timing as compared to Fixed-Interval or Mouth-Open Timing. 
This confirms that our insight to incorporate social signals in model structure (SoNNET) improves bite timing prediction. These results using Couplet-SoNNET also imply that it is possible to predict the behavior of a user using only their co-diner information, which indicates that there is social coordination in human groups even in the presence of a robot.
In Fig.~\ref{fig:exp} (bottom right) we further compared the level of distraction by the robot as perceived by participants. We performed Kruskal-Wallis H-tests and Tukey HSD post-hoc tests and found that Mouth-Open Timing distracts dining participants significantly more than Learned or Fixed-Interval Timing. We believe this is because the Mouth-Open strategy prompts the user using a voice interface, which can disrupt the rhythm of conversation. Even though the participants had a clear preference for the Learned strategy when given a forced-choice, when asked to individually rate the conditions using a 5-point Likert scale, interestingly we could not find any statistically significant differences between Mouth-Open Timing and Learned Timing. This is probably because the Mouth-Open Timing strategy provides full control of bite timing to the users themselves. Note, regardless of the conditions, the users found the system comfortable, reliable, and trustworthy. Detailed analysis is given in App.~\ref{sec:appendix:user_study:results}.





\textbf{Limitations.}
There is a risk that our results from human-robot user studies on adults with no mobility limitations may not generalize to those with people with mobility limitations. People with mobility limitations may have different preferences and cognitive workload associated with a robotic intervention. 
Although our target diner is not a person with such C3-C5 SCI, our model does not use their movements to infer when to feed. As transferability is a function of their behavior, our experiments demonstrate good transferability across scenarios. We expect it to similarly transfer to users with C3-C5 SCI, though it remains to be investigated in future work.
We also made multiple assumptions when transferring our results from human-human to human-robot commensality scenarios. During human-human commensality, the user was self-feeding whereas in human-robot commensality the user was being fed. We also assumed that the addition of a robot into a human-human commensality scenario does not change the social dynamics of the diners significantly. Given these assumptions, it would be interesting to see how our models perform when trained on similar human-robot commensality scenarios. Finally, it is an open question as to how these models would perform with groups of different cultures. 
Social science literature on commensality studied the interplay between factors such as culture~\cite{danesi2012commensality, fischler2011commensality}, age~\cite{biggs2021intergenerational}, and social context~\cite{morrison1996sharing} on how long eating takes and what people are eating. 
We are excited about the potential to study how the presence of a robot can alter the communal act of eating together across cultures.
This motivates further investigation into human-robot commensality, both from technical and societal perspectives.








\clearpage
\acknowledgments{The authors would like to thank Rajat Jenamani, Rishabh Madan, and Sidharth Vasudev for their help with setting up and running the robotics user study. This work was funded in part by the National Science Foundation IIS (\#2132846). This work was also in part sponsored by the Office of Naval Research (N00014-19-1-2299). Any opinions, findings, and conclusions or recommendations expressed in this material are those of the author(s) and do not necessarily reflect the views of the Office of Naval Research.}

\begin{raggedright}
\bibliography{references}  

\begin{thebibliography}{74}
\providecommand{\natexlab}[1]{#1}
\providecommand{\url}[1]{\texttt{#1}}
\expandafter\ifx\csname urlstyle\endcsname\relax
  \providecommand{\doi}[1]{doi: #1}\else
  \providecommand{\doi}{doi: \begingroup \urlstyle{rm}\Url}\fi

\bibitem[Bureau.(2014)]{united2014americans}
U.~S.~C. Bureau.
\newblock Americans with disabilities: 2014.
\newblock 2014.

\bibitem[Perry(2008)]{perry2008assisted}
L.~Perry.
\newblock Assisted feeding.
\newblock \emph{Journal of advanced nursing}, 62\penalty0 (5):\penalty0
  511--511, 2008.

\bibitem[Mlinac and Feng(2016)]{Mlinac2016}
M.~E. Mlinac and M.~C. Feng.
\newblock Assessment of activities of daily living, self-care, and
  independence.
\newblock \emph{Archives of Clinical Neuropsychology}, 31\penalty0
  (6):\penalty0 506--516, 2016.

\bibitem[agi()]{agisLiving}
Agis living.
\newblock
  \url{http://www.agis.com/Document/484/assisted-living-care-with-an-independent-flavor}.

\bibitem[obi(2018)]{obi}
Obi, 2018.
\newblock \url{https://meetobi.com/},[Online; Retrieved on 25th January, 2018].

\bibitem[mys(2018)]{myspoon}
My spoon, 2018.
\newblock \url{https://www.secom.co.jp/english/myspoon/food.html},[Online;
  Retrieved on 25th January, 2018].

\bibitem[mea(2018{\natexlab{a}})]{mealmate}
Meal-mate, 2018{\natexlab{a}}.
\newblock
  \url{https://www.made2aid.co.uk/productprofile?productId=8&company=RBF\%20Healthcare&product=Meal-Mate},[Online;
  Retrieved on 25th January, 2018].

\bibitem[mea(2018{\natexlab{b}})]{mealbuddy}
Meal buddy, 2018{\natexlab{b}}.
\newblock \url{https://www.performancehealth.com/meal-buddy-system},[Online;
  Retrieved on 25th January, 2018].

\bibitem[win(2018)]{winsfordfeeder}
Winsford feeder, 2018.
\newblock \url{https://www.youtube.com/watch?v=KZRFj1UZl-c},[Online; Retrieved
  on 15th February, 2018].

\bibitem[bes(2019)]{bestic}
Bestic, 2019.
\newblock \url{https://www.camanio.com/us/products/bestic/},[Online; Retrieved
  on 18th April, 2019].

\bibitem[mea(2018)]{mealtimepartner}
The mealtime partner dining system, 2018.
\newblock \url{http://mealtimepartners.com/},[Online; Retrieved on 15th
  February, 2018].

\bibitem[nea(2019)]{neater}
Neater eater robot, 2019.
\newblock \url{http://www.neater.co.uk/neater-eater-2-2/},[Online; Retrieved on
  18th April, 2019].

\bibitem[bee(2019)]{beeson}
Beeson automaddak feeder, 2019.
\newblock
  \url{https://abledata.acl.gov/product/beeson-automaddak-feeder-model-h74501},[Online;
  Retrieved on 18th April, 2019].

\bibitem[Bhattacharjee et~al.(2019)Bhattacharjee, Lee, Song, and
  Srinivasa]{bhattacharjee2018food}
T.~Bhattacharjee, G.~Lee, H.~Song, and S.~S. Srinivasa.
\newblock Towards robotic feeding: Role of haptics in fork-based food
  manipulation.
\newblock \emph{IEEE Robotics and Automation Letters}, 2019.

\bibitem[Candeias et~al.(2018)Candeias, Rhodes, Marques, Veloso,
  et~al.]{candeias2018vision}
A.~Candeias, T.~Rhodes, M.~Marques, M.~Veloso, et~al.
\newblock Vision augmented robot feeding.
\newblock In \emph{Proceedings of the European Conference on Computer Vision
  (ECCV) Workshops}, 2018.

\bibitem[Gordon et~al.(2021)Gordon, Roychowdhury, Bhattacharjee, Jamieson, and
  Srinivasa]{gordon2021leveraging}
E.~K. Gordon, S.~Roychowdhury, T.~Bhattacharjee, K.~Jamieson, and S.~S.
  Srinivasa.
\newblock Leveraging post hoc context for faster learning in bandit settings
  with applications in robot-assisted feeding.
\newblock In \emph{2021 IEEE International Conference on Robotics and
  Automation (ICRA)}, pages 10528--10535. IEEE, 2021.

\bibitem[Higa et~al.(2014)Higa, Kurisu, and Uehara]{higa2014vision}
H.~Higa, K.~Kurisu, and H.~Uehara.
\newblock A vision-based assistive robotic arm for people with severe
  disabilities.
\newblock \emph{Transactions on Machine Learning and Artificial Intelligence},
  2\penalty0 (4):\penalty0 12--23, 2014.

\bibitem[Bhattacharjee et~al.(2019)Bhattacharjee, Cabrera, Caspi, Cakmak, and
  Srinivasa]{bhattacharjee2019community}
T.~Bhattacharjee, M.~E. Cabrera, A.~Caspi, M.~Cakmak, and S.~S. Srinivasa.
\newblock A community-centered design framework for robot-assisted feeding
  systems.
\newblock In \emph{The 21st international ACM SIGACCESS conference on computers
  and accessibility}, pages 482--494, 2019.

\bibitem[Jard{\'o}n et~al.(2012)Jard{\'o}n, Monje, and
  Balaguer]{jardon2012functional}
A.~Jard{\'o}n, C.~A. Monje, and C.~Balaguer.
\newblock Functional evaluation of asibot: A new approach on portable robotic
  system for disabled people.
\newblock \emph{Applied Bionics and Biomechanics}, 9\penalty0 (1):\penalty0
  85--97, 2012.

\bibitem[Bhattacharjee et~al.(2019)Bhattacharjee, Lee, Song, and
  Srinivasa]{bhattacharjee2019towards}
T.~Bhattacharjee, G.~Lee, H.~Song, and S.~S. Srinivasa.
\newblock Towards robotic feeding: Role of haptics in fork-based food
  manipulation.
\newblock \emph{IEEE Robotics and Automation Letters}, 4\penalty0 (2):\penalty0
  1485--1492, 2019.

\bibitem[Alqasemi et~al.(2005)Alqasemi, McCaffrey, Edwards, and
  Dubey]{alqasemi2005wheelchair}
R.~M. Alqasemi, E.~J. McCaffrey, K.~D. Edwards, and R.~V. Dubey.
\newblock Wheelchair-mounted robotic arms: Analysis, evaluation and
  development.
\newblock In \emph{Proceedings, 2005 IEEE/ASME International Conference on
  Advanced Intelligent Mechatronics.}, pages 1164--1169. IEEE, 2005.

\bibitem[Bien et~al.(2004)Bien, Chung, Chang, Kwon, Kim, Han, Kim, Kim, Park,
  Kang, et~al.]{bien2004integration}
Z.~Bien, M.-J. Chung, P.-H. Chang, D.-S. Kwon, D.-J. Kim, J.-S. Han, J.-H. Kim,
  D.-H. Kim, H.-S. Park, S.-H. Kang, et~al.
\newblock Integration of a rehabilitation robotic system (kares ii) with
  human-friendly man-machine interaction units.
\newblock \emph{Autonomous robots}, 16\penalty0 (2):\penalty0 165--191, 2004.

\bibitem[Park et~al.(2016)Park, Kim, Erickson, and Kemp]{park2016towards}
D.~Park, Y.~K. Kim, Z.~M. Erickson, and C.~C. Kemp.
\newblock Towards assistive feeding with a general-purpose mobile manipulator.
\newblock \emph{arXiv preprint arXiv:1605.07996}, 2016.

\bibitem[Feng et~al.(2019)Feng, Kim, Lee, Gordon, Schmittle, Kumar,
  Bhattacharjee, and Srinivasa]{feng2019robot}
R.~Feng, Y.~Kim, G.~Lee, E.~K. Gordon, M.~Schmittle, S.~Kumar,
  T.~Bhattacharjee, and S.~S. Srinivasa.
\newblock Robot-assisted feeding: Generalizing skewering strategies across food
  items on a plate.
\newblock In \emph{The International Symposium of Robotics Research}, pages
  427--442. Springer, 2019.

\bibitem[Gallenberger et~al.(2019)Gallenberger, Bhattacharjee, Kim, and
  Srinivasa]{gallenberger2019transfer}
D.~Gallenberger, T.~Bhattacharjee, Y.~Kim, and S.~S. Srinivasa.
\newblock Transfer depends on acquisition: Analyzing manipulation strategies
  for robotic feeding.
\newblock In \emph{2019 14th ACM/IEEE International Conference on Human-Robot
  Interaction (HRI)}, pages 267--276. IEEE, 2019.

\bibitem[Belkhale et~al.(2021)Belkhale, Gordon, Chen, Srinivasa, Bhattacharjee,
  and Sadigh]{belkhale2021balancing}
S.~Belkhale, E.~K. Gordon, Y.~Chen, S.~Srinivasa, T.~Bhattacharjee, and
  D.~Sadigh.
\newblock Balancing efficiency and comfort in robot-assisted bite transfer.
\newblock \emph{arXiv preprint arXiv:2111.11401}, 2021.

\bibitem[Canal et~al.(2016)Canal, Aleny{\`a}, and
  Torras]{canal2016personalization}
G.~Canal, G.~Aleny{\`a}, and C.~Torras.
\newblock Personalization framework for adaptive robotic feeding assistance.
\newblock In \emph{International conference on social robotics}, pages 22--31.
  Springer, 2016.

\bibitem[Rhodes and Veloso(2018)]{rhodes2018robot}
T.~Rhodes and M.~Veloso.
\newblock Robot-driven trajectory improvement for feeding tasks.
\newblock In \emph{2018 IEEE/RSJ International Conference on Intelligent Robots
  and Systems (IROS)}, pages 2991--2996. IEEE, 2018.

\bibitem[Naotunna et~al.(2015)Naotunna, Perera, Sandaruwan, Gopura, and
  Lalitharatne]{naotunna2015meal}
I.~Naotunna, C.~J. Perera, C.~Sandaruwan, R.~Gopura, and T.~D. Lalitharatne.
\newblock Meal assistance robots: A review on current status, challenges and
  future directions.
\newblock In \emph{2015 IEEE/SICE International Symposium on System Integration
  (SII)}, pages 211--216. IEEE, 2015.

\bibitem[Gordon et~al.(2019)Gordon, Meng, Barnes, Bhattacharjee, and
  Srinivasa]{gordon2019learning}
E.~K. Gordon, X.~Meng, M.~Barnes, T.~Bhattacharjee, and S.~S. Srinivasa.
\newblock Learning from failures in robot-assisted feeding: Using online
  learning to develop manipulation strategies for bite acquisition.
\newblock 2019.

\bibitem[Park et~al.(2018)Park, Hoshi, and Kemp]{park2018multimodal}
D.~Park, Y.~Hoshi, and C.~C. Kemp.
\newblock A multimodal anomaly detector for robot-assisted feeding using an
  lstm-based variational autoencoder.
\newblock \emph{IEEE Robotics and Automation Letters}, 3\penalty0 (3):\penalty0
  1544--1551, 2018.

\bibitem[Park et~al.(2017)Park, Kim, Hoshi, Erickson, Kapusta, and
  Kemp]{park2017multimodal}
D.~Park, H.~Kim, Y.~Hoshi, Z.~Erickson, A.~Kapusta, and C.~C. Kemp.
\newblock A multimodal execution monitor with anomaly classification for
  robot-assisted feeding.
\newblock In \emph{2017 IEEE/RSJ International Conference on Intelligent Robots
  and Systems (IROS)}, pages 5406--5413. IEEE, 2017.

\bibitem[Sebo et~al.(2020)Sebo, Stoll, Scassellati, and Jung]{sebo2020robots}
S.~Sebo, B.~Stoll, B.~Scassellati, and M.~F. Jung.
\newblock Robots in groups and teams: a literature review.
\newblock \emph{Proceedings of the ACM on Human-Computer Interaction},
  4\penalty0 (CSCW2):\penalty0 1--36, 2020.

\bibitem[Jung et~al.(2020)Jung, DiFranzo, Shen, Stoll, Claure, and
  Lawrence]{jung2020robot}
M.~F. Jung, D.~DiFranzo, S.~Shen, B.~Stoll, H.~Claure, and A.~Lawrence.
\newblock Robot-assisted tower construction—a method to study the impact of a
  robot’s allocation behavior on interpersonal dynamics and collaboration in
  groups.
\newblock \emph{ACM Transactions on Human-Robot Interaction (THRI)},
  10\penalty0 (1):\penalty0 1--23, 2020.

\bibitem[J{\"o}nsson et~al.(2021)J{\"o}nsson, Michaud, and
  Neuman]{jonsson2021commensality}
H.~J{\"o}nsson, M.~Michaud, and N.~Neuman.
\newblock What is commensality? a critical discussion of an expanding research
  field.
\newblock \emph{International Journal of Environmental Research and Public
  Health}, 18\penalty0 (12):\penalty0 6235, 2021.

\bibitem[Spence et~al.(2019)Spence, Mancini, and Huisman]{spence2019digital}
C.~Spence, M.~Mancini, and G.~Huisman.
\newblock Digital commensality: Eating and drinking in the company of
  technology.
\newblock \emph{Frontiers in psychology}, 10:\penalty0 2252, 2019.

\bibitem[Athlin et~al.(1990)Athlin, Norberg, and Asplund]{athlin1990caregivers}
E.~Athlin, A.~Norberg, and K.~Asplund.
\newblock Caregivers' perceptions and interpretations of severely demented
  patients during feeding in a task assignment system.
\newblock \emph{Scandinavian Journal of Caring Sciences}, 4\penalty0
  (4):\penalty0 147--156, 1990.

\bibitem[Athlin and Norberg(1987)]{athlin1987interaction}
E.~Athlin and A.~Norberg.
\newblock Interaction between the severely demented patient and his caregiver
  during feeding.
\newblock \emph{Scandinavian Journal of Caring Sciences}, 1\penalty0
  (3-4):\penalty0 117--123, 1987.

\bibitem[Herlant(2018)]{herlant2018algorithms}
L.~V. Herlant.
\newblock Algorithms, implementation, and studies on eating with a shared
  control robot arm.
\newblock \emph{PhD Dissertation}, 2018.
\newblock URL
  \url{http://www.cs.cmu.edu/afs/cs/user/lcv/www/herlant-thesis.pdf}.

\bibitem[Bhattacharjee et~al.(2020)Bhattacharjee, Gordon, Scalise, Cabrera,
  Caspi, Cakmak, and Srinivasa]{bhattacharjee2020more}
T.~Bhattacharjee, E.~K. Gordon, R.~Scalise, M.~E. Cabrera, A.~Caspi, M.~Cakmak,
  and S.~S. Srinivasa.
\newblock Is more autonomy always better? exploring preferences of users with
  mobility impairments in robot-assisted feeding.
\newblock In \emph{2020 15th ACM/IEEE International Conference on Human-Robot
  Interaction (HRI)}, pages 181--190. IEEE, 2020.

\bibitem[McFarlane(2002)]{mcfarlane2002comparison}
D.~C. McFarlane.
\newblock Comparison of four primary methods for coordinating the interruption
  of people in human-computer interaction.
\newblock \emph{Human-Computer Interaction}, 17\penalty0 (1):\penalty0 63--139,
  2002.

\bibitem[Czerwinski et~al.(2000)Czerwinski, Cutrell, and
  Horvitz]{czerwinski2000instant}
M.~Czerwinski, E.~Cutrell, and E.~Horvitz.
\newblock Instant messaging: Effects of relevance and timing.
\newblock In \emph{People and computers XIV: Proceedings of HCI}, volume~2,
  pages 71--76, 2000.

\bibitem[Pielot et~al.(2017)Pielot, Cardoso, Katevas, Serr\`{a}, Matic, and
  Oliver]{pielot2017beyond}
M.~Pielot, B.~Cardoso, K.~Katevas, J.~Serr\`{a}, A.~Matic, and N.~Oliver.
\newblock Beyond interruptibility: Predicting opportune moments to engage
  mobile phone users.
\newblock \emph{Proc. ACM Interact. Mob. Wearable Ubiquitous Technol.},
  1\penalty0 (3), Sept. 2017.
\newblock \doi{10.1145/3130956}.
\newblock URL \url{https://doi.org/10.1145/3130956}.

\bibitem[Takemae et~al.(2007)Takemae, Ohno, Yoda, and
  Ozawa]{takemae2007estimating}
Y.~Takemae, T.~Ohno, I.~Yoda, and S.~Ozawa.
\newblock Estimating interruptibility in the home for remote communication
  based on audio-visual tracking.
\newblock \emph{IPSJ Digital Courier}, 3:\penalty0 125--133, 2007.

\bibitem[Banerjee et~al.(2018)Banerjee, Silva, Feigh, and
  Chernova]{banerjee2018effects}
S.~Banerjee, A.~Silva, K.~Feigh, and S.~Chernova.
\newblock Effects of interruptibility-aware robot behavior.
\newblock \emph{arXiv preprint arXiv:1804.06383}, 2018.

\bibitem[Wu et~al.(2021)Wu, Martelaro, Stent, Ortiz, and Ju]{wu2021learning}
T.~Wu, N.~Martelaro, S.~Stent, J.~Ortiz, and W.~Ju.
\newblock Learning when agents can talk to drivers using the inagt dataset and
  multisensor fusion.
\newblock \emph{Proceedings of the ACM on Interactive, Mobile, Wearable and
  Ubiquitous Technologies}, 5\penalty0 (3):\penalty0 1--28, 2021.

\bibitem[Wang et~al.(2021)Wang, Xu, Narasimhan, and Wang]{wang2021multi}
J.~Wang, H.~Xu, M.~Narasimhan, and X.~Wang.
\newblock Multi-person 3d motion prediction with multi-range transformers.
\newblock \emph{Advances in Neural Information Processing Systems}, 34, 2021.

\bibitem[M{\"u}ller et~al.(2018)M{\"u}ller, Huang, Zhang, and
  Bulling]{muller2018robust}
P.~M{\"u}ller, M.~X. Huang, X.~Zhang, and A.~Bulling.
\newblock Robust eye contact detection in natural multi-person interactions
  using gaze and speaking behaviour.
\newblock In \emph{Proceedings of the 2018 ACM Symposium on Eye Tracking
  Research \& Applications}, pages 1--10, 2018.

\bibitem[Park et~al.(2012)Park, Jain, and Sheikh]{park20123d}
H.~Park, E.~Jain, and Y.~Sheikh.
\newblock 3d social saliency from head-mounted cameras.
\newblock \emph{Advances in Neural Information Processing Systems}, 25, 2012.

\bibitem[Joo et~al.(2019)Joo, Simon, Cikara, and Sheikh]{joo2019towards}
H.~Joo, T.~Simon, M.~Cikara, and Y.~Sheikh.
\newblock Towards social artificial intelligence: Nonverbal social signal
  prediction in a triadic interaction.
\newblock In \emph{Proceedings of the IEEE/CVF Conference on Computer Vision
  and Pattern Recognition}, pages 10873--10883, 2019.

\bibitem[Cao et~al.(2018)Cao, Hidalgo, Simon, Wei, and Sheikh]{cao2018openpose}
Z.~Cao, G.~Hidalgo, T.~Simon, S.-E. Wei, and Y.~Sheikh.
\newblock Openpose: realtime multi-person 2d pose estimation using part
  affinity fields.
\newblock \emph{arXiv preprint arXiv:1812.08008}, 2018.

\bibitem[Zhao et~al.(2017)Zhao, Lu, Chen, Liu, and Wu]{zhao2017convolutional}
B.~Zhao, H.~Lu, S.~Chen, J.~Liu, and D.~Wu.
\newblock Convolutional neural networks for time series classification.
\newblock \emph{Journal of Systems Engineering and Electronics}, 28\penalty0
  (1):\penalty0 162--169, 2017.

\bibitem[Yang et~al.(2015)Yang, Nguyen, San, Li, and
  Krishnaswamy]{yang2015deep}
J.~Yang, M.~N. Nguyen, P.~P. San, X.~Li, and S.~Krishnaswamy.
\newblock Deep convolutional neural networks on multichannel time series for
  human activity recognition.
\newblock In \emph{Ijcai}, volume~15, pages 3995--4001. Buenos Aires,
  Argentina, 2015.

\bibitem[Liu et~al.(2018)Liu, Hsaio, and Tu]{liu2018time}
C.-L. Liu, W.-H. Hsaio, and Y.-C. Tu.
\newblock Time series classification with multivariate convolutional neural
  network.
\newblock \emph{IEEE Transactions on Industrial Electronics}, 66\penalty0
  (6):\penalty0 4788--4797, 2018.

\bibitem[RealSense()]{realsense}
RealSense.
\newblock {Introducing the Intel® realsense™ depth camera D455}, 2020.
\newblock URL \url{https://www.intelrealsense.com/depth-camera-d455/}.

\bibitem[Zuo(2018)]{micarray}
B.~Zuo.
\newblock {ReSpeaker Mic Array v2.0}, 2018.
\newblock URL \url{https://wiki.seeedstudio.com/ReSpeaker_Mic_Array_v2.0/}.

\bibitem[Wittenburg et~al.(2006)Wittenburg, Brugman, Russel, Klassmann, and
  Sloetjes]{elan}
P.~Wittenburg, H.~Brugman, A.~Russel, A.~Klassmann, and H.~Sloetjes.
\newblock Elan: A professional framework for multimodality research.
\newblock In \emph{5th International Conference on Language Resources and
  Evaluation (LREC 2006)}, pages 1556--1559, 2006.

\bibitem[Lea et~al.(2016)Lea, Vidal, Reiter, and Hager]{lea2016temporal}
C.~Lea, R.~Vidal, A.~Reiter, and G.~D. Hager.
\newblock Temporal convolutional networks: A unified approach to action
  segmentation.
\newblock In \emph{European conference on computer vision}, pages 47--54.
  Springer, 2016.

\bibitem[Lea et~al.(2017)Lea, Flynn, Vidal, Reiter, and Hager]{lea2017temporal}
C.~Lea, M.~D. Flynn, R.~Vidal, A.~Reiter, and G.~D. Hager.
\newblock Temporal convolutional networks for action segmentation and
  detection.
\newblock In \emph{proceedings of the IEEE Conference on Computer Vision and
  Pattern Recognition}, pages 156--165, 2017.

\bibitem[He and Zhao(2019)]{he2019temporal}
Y.~He and J.~Zhao.
\newblock Temporal convolutional networks for anomaly detection in time series.
\newblock In \emph{Journal of Physics: Conference Series}, volume 1213, page
  042050. IOP Publishing, 2019.

\bibitem[Chicco and Jurman(2020)]{chicco2020advantages}
D.~Chicco and G.~Jurman.
\newblock The advantages of the matthews correlation coefficient (mcc) over f1
  score and accuracy in binary classification evaluation.
\newblock \emph{BMC genomics}, 21\penalty0 (1):\penalty0 1--13, 2020.

\bibitem[gen(2022)]{gen3}
Kinova gen3.
\newblock \url{https://www.kinovarobotics.com/product/gen3-robots}, 2022.
\newblock Accessed: 2022-06-14.

\bibitem[rov(2022)]{rovi}
Rovi wheelchair.
\newblock \url{https://www.rovimobility.com/}, 2022.
\newblock Accessed: 2022-06-14.

\bibitem[Jian et~al.(2000)Jian, Bisantz, and Drury]{jian2000foundations}
J.-Y. Jian, A.~M. Bisantz, and C.~G. Drury.
\newblock Foundations for an empirically determined scale of trust in automated
  systems.
\newblock \emph{International journal of cognitive ergonomics}, 4\penalty0
  (1):\penalty0 53--71, 2000.

\bibitem[Danesi(2012)]{danesi2012commensality}
G.~Danesi.
\newblock Commensality in french and german young adults: An ethnographic
  study.
\newblock \emph{Hospitality \& Society}, 1\penalty0 (2):\penalty0 153--172,
  2012.

\bibitem[Fischler(2011)]{fischler2011commensality}
C.~Fischler.
\newblock Commensality, society and culture.
\newblock \emph{Social science information}, 50\penalty0 (3-4):\penalty0
  528--548, 2011.

\bibitem[Biggs and Haapala(2021)]{biggs2021intergenerational}
S.~Biggs and I.~Haapala.
\newblock Intergenerational commensality: A critical discussion on non-familial
  age groups eating together.
\newblock \emph{International Journal of Environmental Research and Public
  Health}, 18\penalty0 (15):\penalty0 7905, 2021.

\bibitem[Morrison(1996)]{morrison1996sharing}
M.~Morrison.
\newblock Sharing food at home and school: perspectives on commensality.
\newblock \emph{The Sociological Review}, 44\penalty0 (4):\penalty0 648--674,
  1996.

\bibitem[Fischer et~al.(2018)Fischer, Chang, and Demiris]{fischer2018rt}
T.~Fischer, H.~J. Chang, and Y.~Demiris.
\newblock Rt-gene: Real-time eye gaze estimation in natural environments.
\newblock In \emph{Proceedings of the European Conference on Computer Vision
  (ECCV)}, pages 334--352, 2018.

\bibitem[Fischer and Demiris(2016)]{fischer2016markerless}
T.~Fischer and Y.~Demiris.
\newblock Markerless perspective taking for humanoid robots in unconstrained
  environments.
\newblock In \emph{2016 IEEE International Conference on Robotics and
  Automation (ICRA)}, pages 3309--3316. IEEE, 2016.

\bibitem[web(2022)]{webrtcvad}
{ython interface to the WebRTC Voice Activity Detector}.
\newblock \url{https://github.com/wiseman/py-webrtcvad}, 2022.
\newblock Accessed: 2022-06-21.

\bibitem[Remy(2020)]{KerasTCN}
P.~Remy.
\newblock Temporal convolutional networks for keras.
\newblock \url{https://github.com/philipperemy/keras-tcn}, 2020.

\bibitem[Sankaran et~al.(2021)Sankaran, Derechin, and
  Christakis]{sankaran2021curmelo}
S.~Sankaran, J.~Derechin, and N.~A. Christakis.
\newblock Curmelo: The theory and practice of a forced-choice approach to
  producing preference rankings.
\newblock \emph{PloS one}, 16\penalty0 (5):\penalty0 e0252145, 2021.

\bibitem[Mehrani and Peterson(2015)]{mehrani2015recency}
M.~B. Mehrani and C.~Peterson.
\newblock Recency tendency: Responses to forced-choice questions.
\newblock \emph{Applied Cognitive Psychology}, 29\penalty0 (3):\penalty0
  418--424, 2015.

\end{thebibliography}
\end{raggedright}
\newpage
\section{Appendix}\label{sec:appendix}

For a video of our work, see \url{https://www.youtube.com/watch?v=5KLGrvrPjMc}

\subsection{Human-Human Commensality Dataset (HHCD) Details}\label{sec:appendix:dataset}

\subsubsection{Summary of Available Data}\label{sec:appendix:dataset:summary}
Overall, the Human-Human Commensality Dataset (HHCD) contains 30 sessions, totalling over 18 hours of multistream, multimodal recordings of 90 people, and provides the following data.
\begin{itemize}
    \item ROS bags with topics: 4x mic audio, mixed audio, sound direction, per-participant RGBD, and scene RGBD
    
    \item Raw data (extracted from ROS bags): scene audio, sound direction, per-participant videos, and scene videos
    
    \item Processed data (extracted from raw data): per-participant speaking status, per-participant face and body keypoints from OpenPose~\cite{cao2018openpose}, per-participant gaze and head pose from RT-GENE~\cite{fischer2018rt}, per-participant bite count, and per-participant times since last bite lifted and since last bite delivered to mouth
    
    \item Annotations: per-participant interactions with food, drink, and napkins (all entered, lifted, delivered to mouth, and mouth open events), per-participant food type labels and observations of interesting behaviors
\end{itemize}
The HHCD dataset is available at \url{https://emprise.cs.cornell.edu/hrcom/}

\subsubsection{Data Collection Setup Measures}\label{sec:appendix:dataset:measures}
\begin{figure}[ht]
    \centering
     \includegraphics[width=0.495\textwidth]{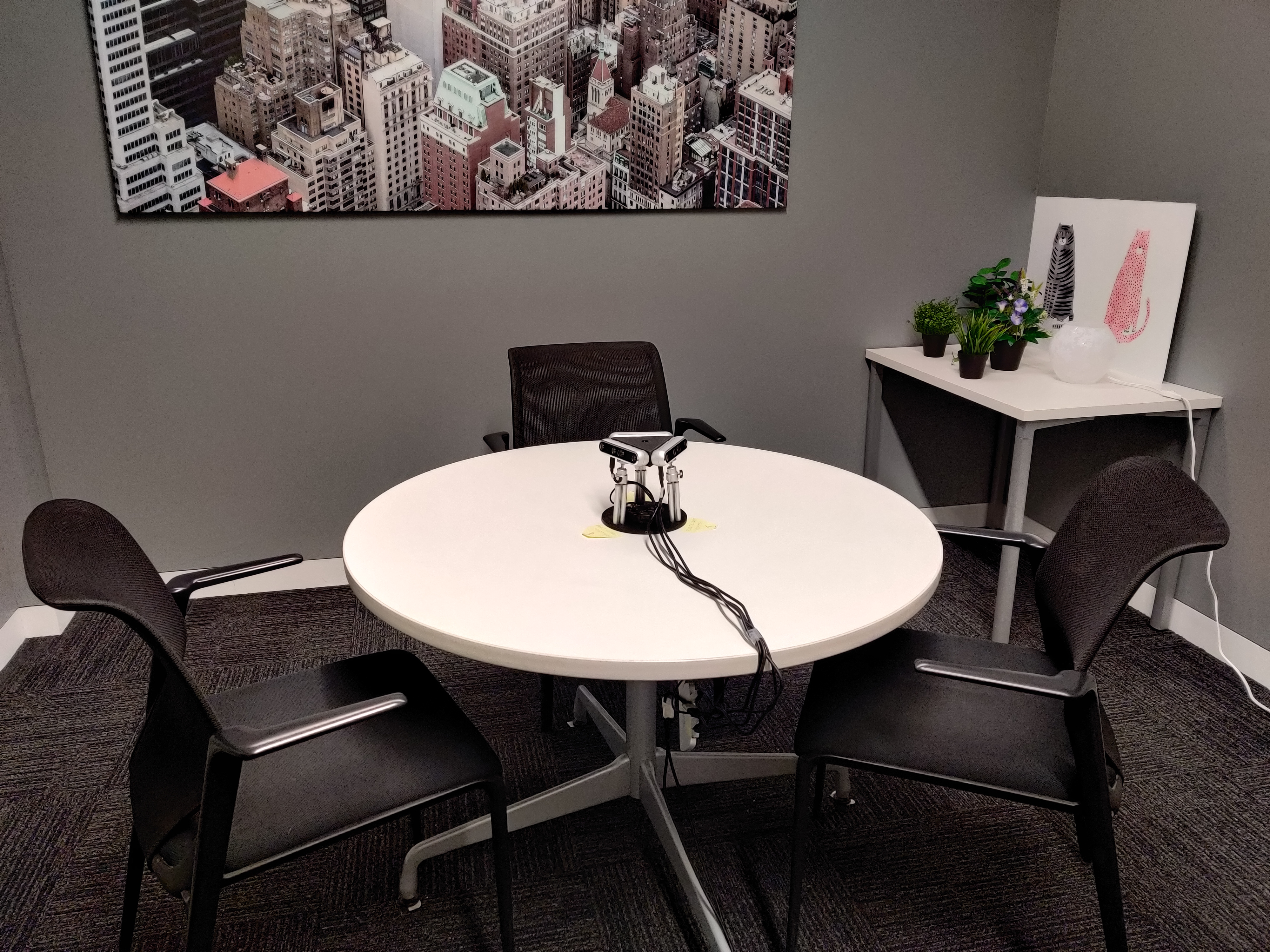}
     \includegraphics[width=0.495\textwidth]{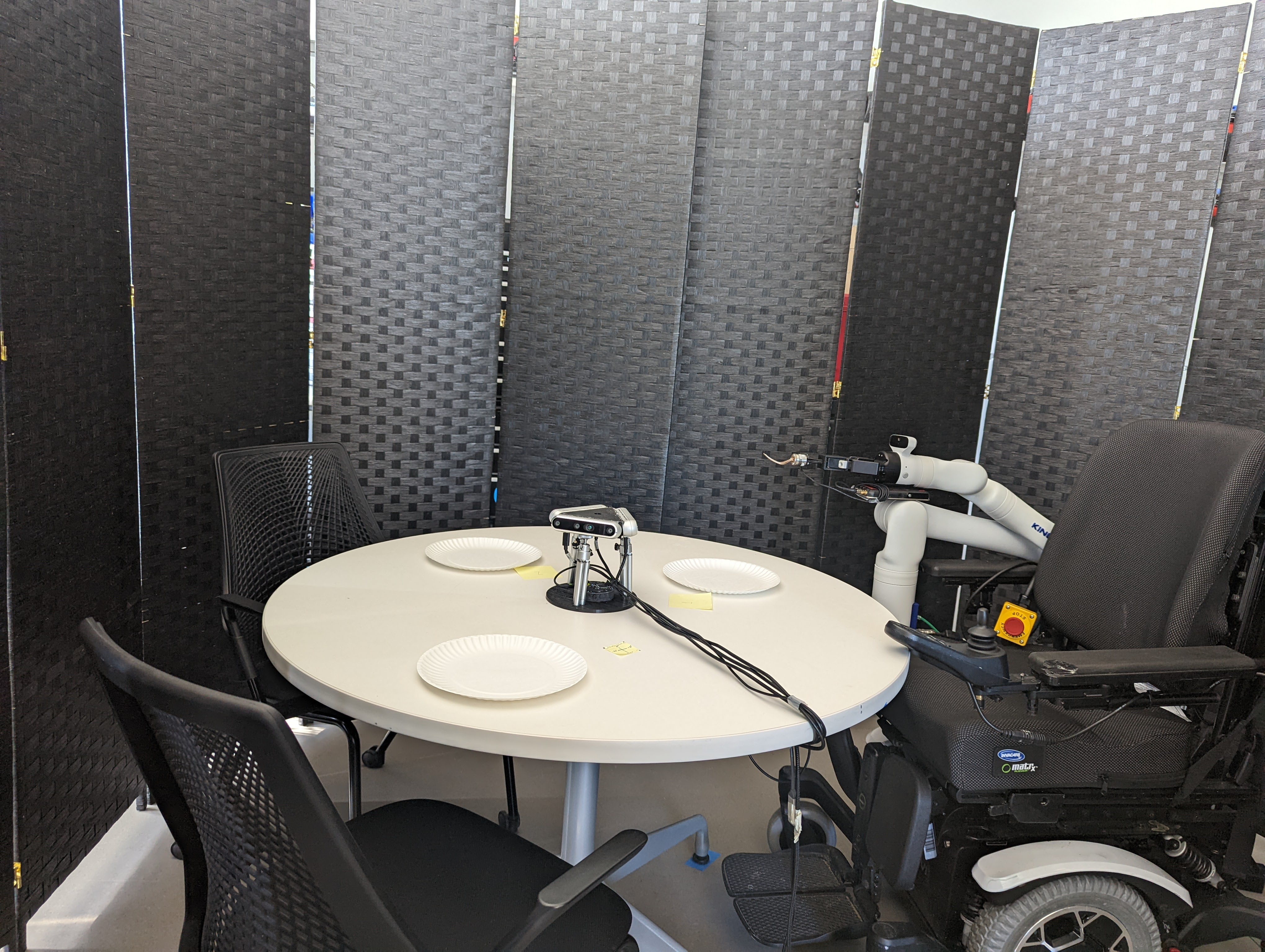}
    \caption{
        \textbf{Left:} Human-human commensality experimental setup, described in App.~\ref{sec:appendix:dataset:measures}.
        \textbf{Right:} Human-robot commensality experimental setup, described in App.~\ref{sec:appendix:user_study}.
    }
    \label{fig:appendix:exp_setups}
    \vspace{-2mm}
\end{figure}

We set up the data collection study with the following measures, depicted in Fig.~\ref{fig:appendix:exp_setups}:
\begin{itemize}
    \item Table diameter: 105 cm
    \item Distance between the ground and the top of the table: 72.5 cm
    \item Distance between the table center and a participant camera center: 6 cm horizontally
    \item Camera triangle side: 11.6 cm horizontally
    \item Distance between the top of the table and the center of a participant camera lens: 14 cm vertically
    \item Participant camera tilt: 12$^\circ$ above the horizontal plane
    \item Distance between the table center and the scene camera: 170 cm horizontally
    \item Distance between the ground and the scene camera: 119.5 cm
    \item Angle between the camera of the participant at position 1 and the scene camera: 41$^\circ$ in the clockwise direction (toward the participant at position 3) in horizontal plane around the table center
    \item Microphone array square side: 4.5 cm
    \item Angle between the zero degree sound direction of the microphone array and the camera of the participant at position 3: 49$^\circ$ in the counter-clockwise direction (toward the participant at position 1) in horizontal plane around the table center
\end{itemize}

\subsubsection{Questionnaires}\label{sec:appendix:dataset:collection}
The questions we asked the participants in the pre-study and post-study questionnaires are shown in Fig.~\ref{fig:appendix:dataset:collection:prestudy_questionnaire} and Fig.~\ref{fig:appendix:dataset:collection:poststudy_questionnaire} respectively.

\subsubsection{Data Annotation Details}\label{sec:appendix:dataset:annotation}
Using the ELAN annotation tool~\cite{elan}, we annotated each participant's video (excluding the scene videos) based on participant's interactions with food, drink, and napkins. 
We defined the following annotation types and associated sets of annotation values. The annotation value was assigned based on the type of utensil involved.
\begin{itemize}
    \item \textit{\textbf{mouth\_open}} $\in$ \{$\emptyset$\}: 
    From the time the mouth opened due to an immediately following food-to-mouth handover until it closed. The frames where the mouth was open for other reasons were ignored. 
    If the mouth was open even when not eating, the mouth\_open annotation began when the mouth started opening more due to an incoming food item and similarly, the mouth\_open annotation ended when the mouth closed the most the first time after eating the bite.

    \item \textit{\textbf{food\_to\_mouth}} $\in$ \{fork, knife, spoon, chopsticks, hand, $\emptyset$\}: 
    From the time the food item entered the mouth (i.e., got above teeth) until the given utensil/hand first lost contact with the mouth (or started moving away from mouth in case the utensil/hand did not touch the mouth). 
    Subsequent actions (if any) to correct/fix an unsuccessful feeding attempt were ignored unless they involved a proper food item pick up. 
    There was exactly one food\_to\_mouth annotation for each mouth\_open annotation such that the mouth\_open annotation always started before the food\_to\_mouth annotation but they could have ended in any order. 
    If the food was consumed without the use of utensil/hand and the person just moved head towards the table to eat a bite, an empty annotation value was assigned.
    
    \item \textit{\textbf{food\_entered}} $\in$ \{fork, knife, spoon, chopsticks, hand\}: 
    First 400 ms after the person touched/entered the food with a utensil/hand. 
    If there were multiple such events before the next food\_to\_mouth annotation (e.g., the person first entered the food, then rested, and later entered the food again), only the first such event was annotated. The reason was to record the first intention to eat. 
    Events when the utensil touched/entered the food just because it was put on top of the food to free up hands were ignored, and the food\_entered annotation started once they touched/entered the food again. 
    So there was exactly one food\_entered annotation prior to each food\_to\_mouth annotation. However, when the person used two/more kinds of utensils/hands at the same time, the food\_entered annotation was made for each utensil/hand independently and not each of them was followed by the food\_to\_mouth annotation of the same utensil/hand type (e.g., food entered by fork, food entered by knife, food lifted by fork, food delivered to mouth using fork but without knife). Also, when the food was grabbed by hand, there might not have been a food\_entered annotation prior to each food\_to\_mouth annotation (e.g., when the person kept holding their food, such as a sandwich, in their hand between bites). 
    If two/more food\_entered annotations with different values overlapped, some annotations were shortened below 400 ms, as ELAN does not allow overlapping annotations within one tier. 
    
    \item \textit{\textbf{food\_lifted}} $\in$ \{fork, knife, spoon, chopsticks, hand, $\emptyset$\}: 
    First 400 ms after the utensil performing the food-to-mouth handover lost contact with the rest of the food or with another utensil/hand involved in food manipulation, whichever occurred later. In case the food was grabbed by hand, the first 400 ms after the food started moving towards the mouth. 
    If there were multiple such events before the food\_to\_mouth annotation (e.g., the person first lifted the food item a bit, then returned it back to the rest of the food to dip it in a sauce, and later lifted it again), only the last lift off event was annotated. The reason was to record only such food lift off events that immediately led to feeding. 
    So there was exactly one food\_lifted annotation prior to each food\_to\_mouth annotation. However, when the person used two/more kinds of utensils/hands at the same time, the food\_lifted annotation was made only for the last lift off before the food\_to\_mouth annotation of the same utensil/hand type (e.g., if the food was entered by fork, lifted by fork, handed over to spoon, lifted by spoon, and finally, delivered to mouth using spoon, then the fork lift off was not annotated). 
    If the food was consumed without the use of utensil/hand and the person just moved head towards the table to eat a bite, the annotation was made when the head started moving towards the food item and the empty annotation value was used. 
    When candies/chocolates were consumed, the food\_lifted annotation was made only after the candy/chocolate was unwrapped. 

    \item \textit{\textbf{drink\_to\_mouth}} $\in$ \{cup, bottle\}: 
    From the time the cup/bottle/straw touched the mouth until it left the mouth. 
    
    \item \textit{\textbf{drink\_entered}} $\in$ \{cup, bottle\}: 
    First 400 ms after the person grabbed the drink with their hand. 
    If there were multiple such events before the drink\_to\_mouth annotation (e.g., the person first grabbed the drink, then dropped it, and later grabbed the drink again), only the first such event was annotated. The reason was to record the first intention to drink. 
    So there was exactly one drink\_entered annotation prior to each drink\_to\_mouth annotation unless they kept holding the drink between two drink\_to\_mouth annotations. Also, if the person used a bottle to pour drink into a cup, the drink\_entered annotation was made for both: when they grabbed the bottle and when they grabbed the cup. 

    \item \textit{\textbf{drink\_lifted}} $\in$ \{cup, bottle\}: 
    First 400 ms after the drink lost contact with the table and started moving towards the mouth (or just started moving towards the mouth in case they kept the drink in hand after the last drink\_to\_mouth annotation). 
    If there were multiple such events before the drink\_to\_mouth annotation (e.g., the person first moved the drink towards the mouth, then stopped a bit, and later completed the move) only the last move towards the mouth was annotated. The reason was to record only such drink lift off events that immediately led to drinking. 
    So there was exactly one drink\_lifted annotation prior to each drink\_to\_mouth annotation. 

    \item \textit{\textbf{napkin\_to\_mouth}} $\in$ \{$\emptyset$\}: 
    From the time the napkin touched the mouth until it left the mouth. 
    
    \item \textit{\textbf{napkin\_entered}} $\in$ \{$\emptyset$\}: 
    First 400 ms after the person grabbed the napkin with their hand. 
    If there were multiple such events before the napkin\_to\_mouth annotation (e.g., the person first grabbed the napkin, then dropped it, and later grabbed the napkin again), only the first such event was annotated. The reason was to record the first intention to use the napkin. 
    So there was exactly one napkin\_entered annotation prior to each napkin\_to\_mouth annotation unless they kept holding the napkin between two napkin\_to\_mouth annotations. 
    
    \item \textit{\textbf{napkin\_lifted}} $\in$ \{$\emptyset$\}: 
    First 400 ms after the napkin lost contact with the table and started moving towards the mouth (or just started moving towards the mouth in case they kept the napkin in hand after the last napkin\_to\_mouth annotation).
    If there were multiple such events before the napkin\_to\_mouth annotation (e.g., the person first moved the napkin towards the mouth, then stopped a bit, and later completed the move) only the last move towards the mouth was annotated. The reason was to record only such napkin lift off events that immediately led to its use. 
    So there was exactly one napkin\_lifted annotation prior to each napkin\_to\_mouth annotation. 

    \item \textit{\textbf{disruption}} $\in$ \{light\_off, participant\_left\}: 
    From the time the recording became disrupted due to the light turning off or due to a participant leaving the room until the normal conditions were restored. 
\end{itemize}

We further defined the following additional annotation rules:
\begin{itemize}
    
    \item When people were just unpacking their food/drink or loading their plates from shared bowls/containers
    \begin{itemize}
        \item No food/drink\_entered and food/drink\_lifted associated annotations
        \item Reason: we are not researching the preparation phase prior to eating
    \end{itemize}
    
    \item When people tore their food (e.g., a piece of bread)
    \begin{itemize}
        \item No additional food\_entered annotations when the other hand touches the food
        \item Reason: we consider tearing the food as a part of the food manipulation that follows the most recent food\_entered annotation and precedes the food\_lifted annotation
    \end{itemize}
    
    \item When people licked their empty utensil/fingers/hands or foils (e.g., yogurt lid)
    \begin{itemize}
        \item No food/drink\_entered, food/drink\_lifted, food/drink\_to\_mouth, and mouth\_open associated annotations
        \item Reason: there is no food/drink consumed
    \end{itemize}
    
    \item When people smelled their food/drink
    \begin{itemize}
        \item No food/drink\_entered, food/drink\_lifted, food/drink\_to\_mouth, and mouth\_open associated annotations
        \item Reason: there is no food/drink consumed
    \end{itemize}
    
    \item When people used a napkin for anything else than cleaning their mouth (e.g., blowing/swiping their nose, cleaning their hands/eyes/utensil/table)
    \begin{itemize}
        \item No napkin\_entered, napkin\_lifted, napkin\_to\_mouth associated annotations, but if the person cleaned their hands and then suddenly decided to clean their mouth, then the napkin\_lifted annotation was made when the napkin started to move towards mouth and also the napkin\_to\_mouth annotation was made. If the initial intention to pick up the napkin seemed to be to eventually clean the mouth, then also the napkin\_entered annotation was made. 
        \item Reason: blowing/swiping nose and cleaning hands/etc.~is not directly related to eating/drinking
    \end{itemize}
    
    \item When people picked up a napkin from their lap
    \begin{itemize}
        \item No napkin\_entered associated annotation
        \item Reason: the napkin was most likely entered earlier and just put on their lap
    \end{itemize}
    
    \item When people grabbed the bottle only to close it or read its label
    \begin{itemize}
        \item No drink\_entered, nor drink\_lifted associated annotations
        \item Reason: there is no drink consumed

    \end{itemize}
    
    \item When the food/drink/napkin\_to\_mouth or mouth\_open event was already in progress at the beginning of the video
    \begin{itemize}
        \item No food/drink/napkin\_to\_mouth, and mouth\_open associated annotation
        \item Reason: we are not able to determine the beginning of such an event
    \end{itemize}
    
    \item When there was a disruption (light went off or participant left)
    \begin{itemize}
        \item No other annotations (besides the disruption annotation) during the disruption interval. New *\_entered and *\_lifted annotations had to be made after the disruption (i.e., any *\_entered and *\_lifted annotations from before the disruption occurred were forgotten).
        \item Reason: the data from the disrupted interval are not used and the disruption is considered as a reset
    \end{itemize}
    
    
    \item When people used coffee stirrer sticks to put spread/jam on a piece of bread
    \begin{itemize}
        \item All the associated events were annotated with the "knife" annotation value
    \end{itemize}
    
    \item When people drank soup (e.g., from a cup)
    \begin{itemize}
        \item The food\_entered/lifted/to\_mouth and mouth\_open annotations were used with the "hand" annotation value.
    \end{itemize}
    
    \item When people drank from the bottle cap
    \begin{itemize}
        \item All the associated events were annotated with the "cup" annotation value
    \end{itemize}
    
    \item When people grabbed or lifted the food/drink/napkin outside of the camera view 
    \begin{itemize}
        \item The start of the associated annotation was estimated but the annotation was not skipped
    \end{itemize}
    
    \item When people picked up and ate small food items such as crumbs
    \begin{itemize}
        \item The food\_entered/lifted/to\_mouth and mouth\_open annotations were not skipped
    \end{itemize}
    
    \item When people ate a sandwich/wrap and decided to pick a small piece with fingers from the rest of the sandwich
    \begin{itemize}
        \item The food\_entered/lifted/to\_mouth and mouth\_open annotations were not skipped
    \end{itemize}
    
    \item When the food entered the mouth but the person did not take a bite
    \begin{itemize}
        \item The food\_entered/lifted/to\_mouth and mouth\_open annotations were not skipped
    \end{itemize}
    
    \item When there was an incomplete (*\_entered, *\_lifted, *\_to\_mouth) sequence at the beginning or end of the video
    \begin{itemize}
        \item For example, the first annotation could be food\_lifted, mouth open or food\_to\_mouth without prior food\_entered. Similarly, the last annotation could be food\_entered or food\_lifted.
    \end{itemize}
    
    \item When the feeding failed at the mouth (e.g., even if the whole food item falls down during the food-to-mouth handover)
    \begin{itemize}
        \item The food\_entered/lifted/to\_mouth and mouth\_open annotations were not skipped
    \end{itemize}

\end{itemize}

\subsubsection{Additional Data Statistics}\label{sec:appendix:dataset:stats}

\begin{figure}[t]
    \centering
     \includegraphics[width=\textwidth]{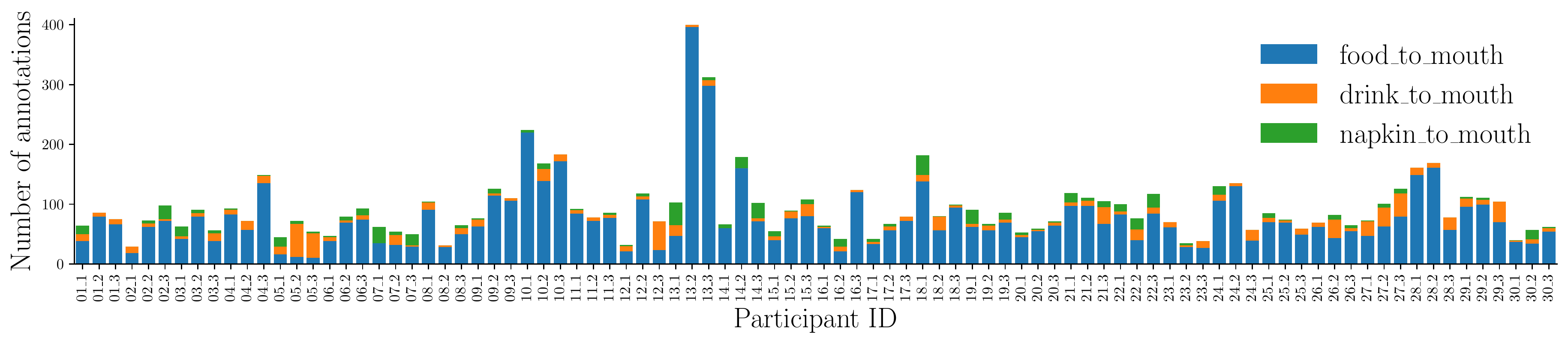}
     \includegraphics[width=\textwidth]{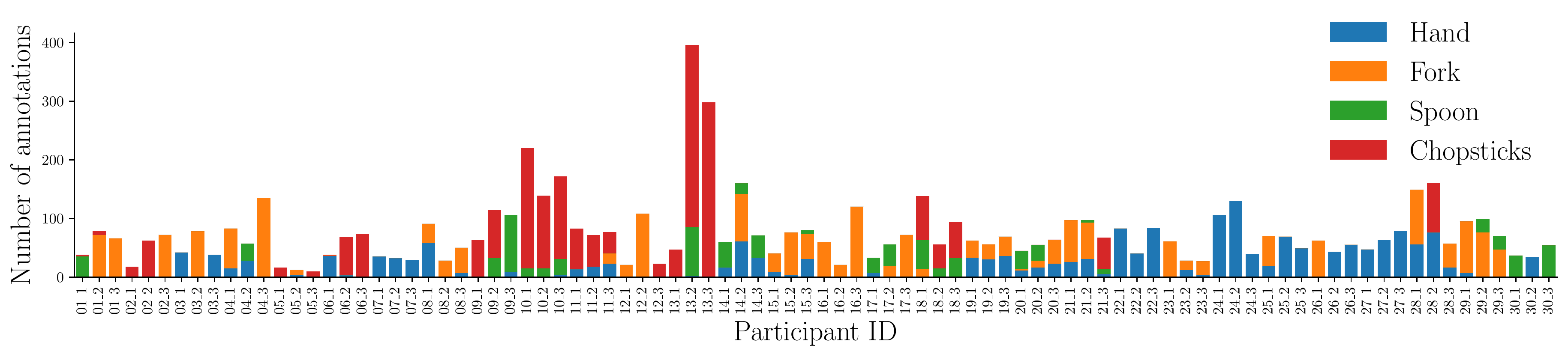}
     \includegraphics[width=\textwidth]{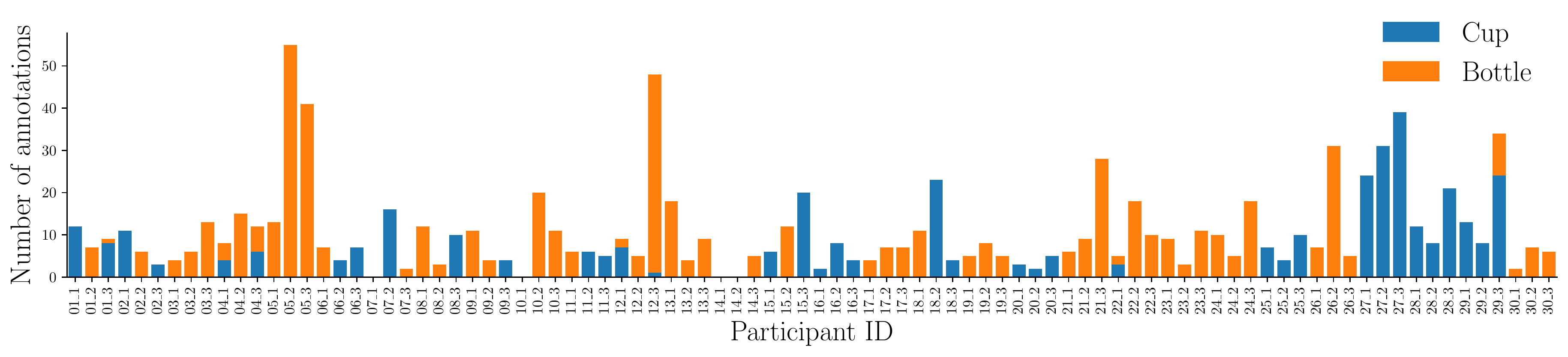}
    \caption{HHCD: Distribution of annotations across participants/videos. 
    \textbf{Top:} Distribution of annotations by annotation type.
    \textbf{Middle} Distribution of \textit{food\_to\_mouth} annotations by annotation value.
    \textbf{Bottom:} Distribution of \textit{drink\_to\_mouth} annotations by annotation value.
    Participant ID is encoded as \{session-number\}\_\{participant-position\}.
    }
    \label{fig:dataset:stats:annotation_counts:per_participant}
\end{figure}

\textbf{Annotation counts.}
The summary of all annotation counts by annotation type is provided in Tab.~\ref{tab:dataset:annotation_counts:by_type} and the distribution of annotations by annotation value is shown in Fig.~\ref{fig:dataset:annotation_counts:by_value}.
Figure~\ref{fig:dataset:stats:annotation_counts:per_participant} further shows the distribution of annotations by types and values across participants/videos.

\begin{table}[t]
\centering
    \begin{minipage}{.31\textwidth}
        \centering
        \caption{HHCD: Annotation counts by annotation type.}
        \begin{tabular}{lc}
            \toprule
            Annotation type     & Count\\ \midrule
            mouth\_open         & 6,834\\
            food\_entered       & 6,000\\
            food\_lifted        & 6,830\\
            food\_to\_mouth     & 6,834\\ \midrule
            drink\_entered      & 755\\
            drink\_lifted       & 981\\
            drink\_to\_mouth    & 978\\ \midrule
            napkin\_entered     & 380\\
            napkin\_lifted      & 600\\
            napkin\_to\_mouth   & 598\\ \midrule
            disruption          & 16\\ \midrule
            Total               & 30,806\\ \bottomrule
        \end{tabular}
        \label{tab:dataset:annotation_counts:by_type}
    \end{minipage}
    \begin{minipage}{.68\textwidth}
        \centering
        \centering
        \includegraphics[height=0.205\textheight]{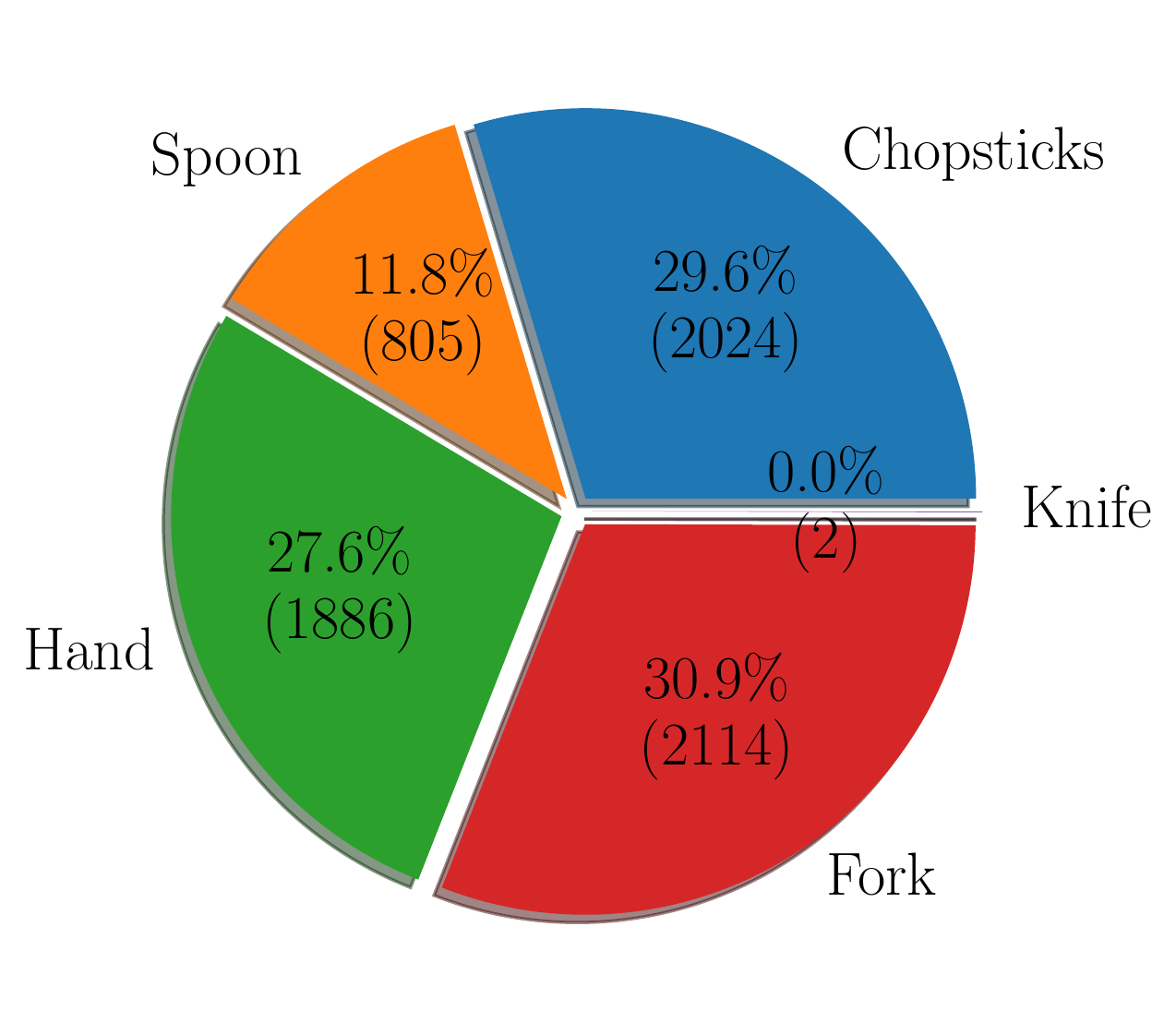}
        \includegraphics[height=0.205\textheight,trim=1.1cm 0 1cm 0,clip]{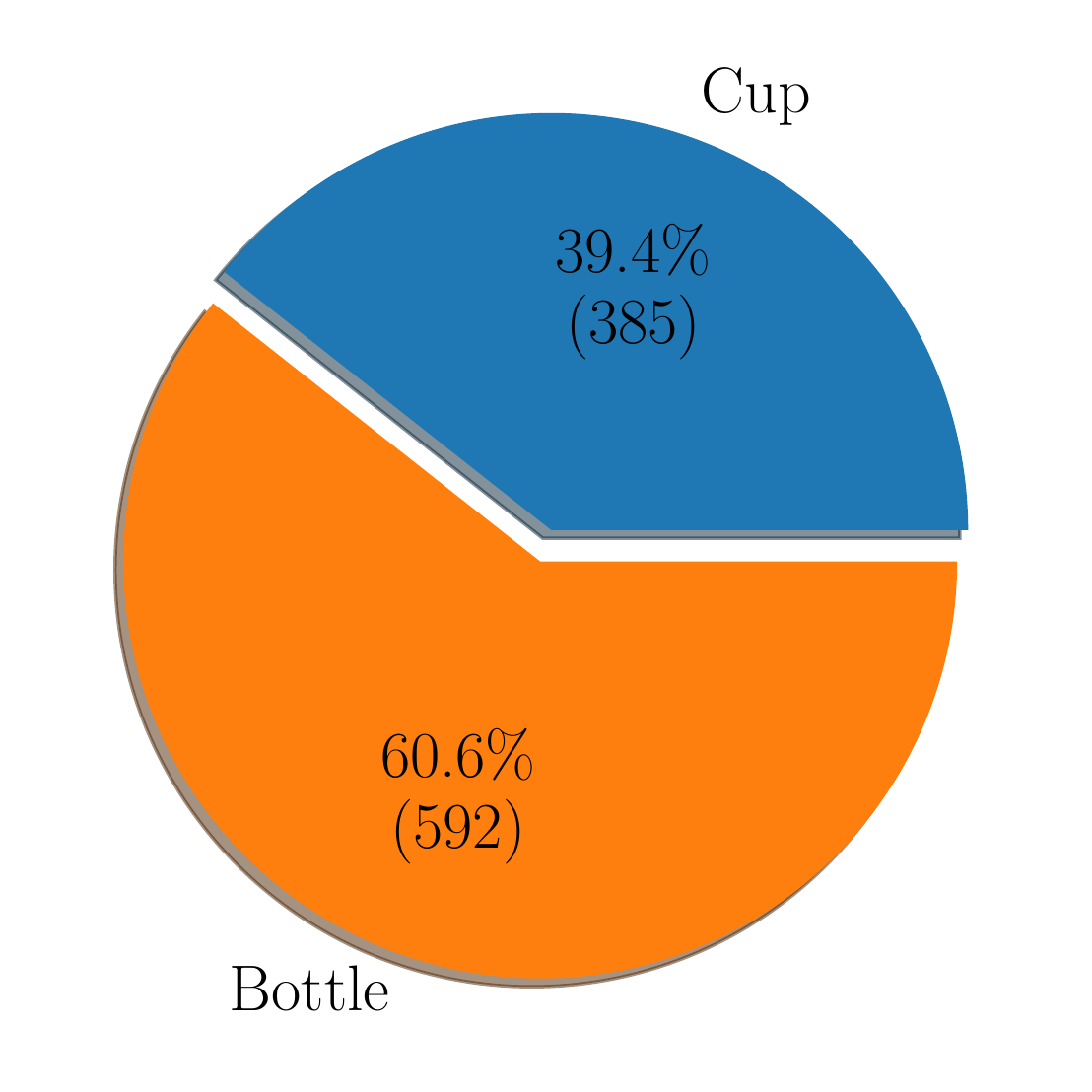}
        \captionof{figure}{HHCD: Distribution of annotations by annotation value. 
            \textbf{Left:} Distribution of \textit{food\_to\_mouth} annotations. 
            \textbf{Right:} Distribution of \textit{drink\_to\_mouth} annotations. 
            All annotations of interactions with napkin have an empty annotation value.
            The \textit{disruption} annotations include one annotation of participant leaving the room for a while and 15 annotations of light turning off for a bit.
        }
        \label{fig:dataset:annotation_counts:by_value}
    \end{minipage}
\end{table}

\textbf{Annotation durations.}
Means and standard deviations of annotation durations by annotation type and annotation value are shown in Tab.~\ref{tab:dataset:annotation_durations}.

\renewcommand*{\minval}{0.7}
\renewcommand*{\maxval}{3.0}
\begin{table}[b] 
\centering
    \caption{HHCD: Annotation durations (mean $\pm$ std) by annotation type and annotation value.
    We report only variable-length annotation types and exclude \textit{disruption}. 
    Cell colors (yellow--red) correspond to mean annotation duration on a scale 0.7--3.0 seconds.
    }
    \begin{minipage}{.35\textwidth}
        \centering
        \begin{tabular}{lc}
            \toprule
            Annotation type & Duration (s)\\ \midrule
            mouth\_open         & \gradient{1.2}$\pm$ 0.9 \\
            food\_to\_mouth     & \gradient{0.9}$\pm$ 0.8 \\
            drink\_to\_mouth    & \gradient{2.9}$\pm$ 1.8 \\
            napkin\_to\_mouth   & \gradient{1.6}$\pm$ 1.8 \\ \bottomrule
        \end{tabular}
    \end{minipage}
    \begin{minipage}{.2\textwidth}
        \centering
        \begin{tikzpicture}
        \draw[white] (-1.5,  0) -- (1.1, 0);
        \draw (-1.5,  -2.05) -- (1.1, -0.6);
        \draw (-1.5,  -2.05) -- (1.1, -2);
        \draw (-1.5,  -2.45) -- (1.1, -2.3);
        \draw (-1.5,  -2.45) -- (1.1, -3.7);
        \draw (-1.5, -2.85) -- (1.1, -3.95);
        \draw (-1.5, -2.85) -- (1.1, -4.7);
        \end{tikzpicture}
    \end{minipage}
    \begin{minipage}{.35\textwidth}
        \centering
        \begin{tabular}{lc}
            \toprule
            Annotation value & Duration (s)\\ \midrule
            Chopsticks & \gradient{0.8}$\pm$ 0.6 \\
            Spoon      & \gradient{1.0}$\pm$ 0.4 \\
            Hand       & \gradient{1.9}$\pm$ 1.2 \\
            Fork       & \gradient{0.9}$\pm$ 0.6 \\ \midrule
            Chopsticks & \gradient{0.7}$\pm$ 0.4 \\
            Spoon      & \gradient{0.9}$\pm$ 0.5 \\
            Hand       & \gradient{1.5}$\pm$ 1.3 \\
            Fork       & \gradient{0.7}$\pm$ 0.4 \\ \midrule
            Bottle     & \gradient{3.0}$\pm$ 1.6 \\
            Cup        & \gradient{2.7}$\pm$ 2.0 \\ \bottomrule
        \end{tabular}
    \end{minipage}
    \label{tab:dataset:annotation_durations}
\end{table}

\textbf{Time gaps between annotations.}
In Tab.~\ref{tab:dataset:time_between_annotations}, we report mean and standard deviation of duration (time gap) between two consequent annotations of both the same annotation type (e.g., from \textit{food\_lifted} to \textit{food\_lifted}) as well as different annotation type (e.g., from \textit{food\_lifted} to \textit{food\_to\_mouth}).
We aggregate the times by annotation type and annotation value.

\begin{table}[!htb] 
\centering
    \caption{HHCD: Time gaps (mean $\pm$ std) between two consequent annotations of the same annotation type \textbf{(left)} and of a different annotation type \textbf{(right)}.
    Aggregated by annotation type and annotation value. 
    The \textit{disruption} annotation type is excluded.
    Cell colors (yellow--red) correspond to mean time gap between annotations on a scale 18.9--196.3 seconds (left) and 0.3--11.3 seconds~(right).
    }
    \begin{minipage}{.485\textwidth}
    \renewcommand*{\minval}{18.9}
    \renewcommand*{\maxval}{196.3}
    \centering
    \scalebox{0.88}{
    \begin{tabular}{clc}
        \toprule
        Annotation type & Ann.~value & Time gap (s)\\ \midrule
        \multirow{5}*{mouth\_open} & \textit{All} & \gradient{23.5}$\pm$ 39.8 \\
                            & Chopsticks        & \gradient{18.9}$\pm$ 34.6 \\
                            & Spoon             & \gradient{27.9}$\pm$ 48.6 \\
                            & Hand              & \gradient{26.6}$\pm$ 41.5 \\
                            & Fork              & \gradient{23.6}$\pm$ 38.7 \\
        \midrule
        \multirow{5}*{food\_entered}       & \textit{All}      & \gradient{26.5}$\pm$ 47.2 \\
                            & Chopsticks        & \gradient{19.2}$\pm$ 34.0 \\
                            & Spoon             & \gradient{27.4}$\pm$ 51.8 \\
                            & Hand              & \gradient{47.1}$\pm$ 71.8 \\
                            & Fork              & \gradient{24.7}$\pm$ 40.2 \\
        \midrule
        \multirow{5}*{food\_lifted}        & \textit{All}      & \gradient{23.6}$\pm$ 39.8 \\
                            & Chopsticks        & \gradient{18.9}$\pm$ 34.8 \\
                            & Spoon             & \gradient{28.0}$\pm$ 48.6 \\
                            & Hand              & \gradient{26.6}$\pm$ 41.4 \\
                            & Fork              & \gradient{23.6}$\pm$ 38.7 \\
        \midrule
        \multirow{5}*{food\_to\_mouth}     & \textit{All} & \gradient{23.5}$\pm$ 39.8 \\        
                            & Chopsticks        & \gradient{18.9}$\pm$ 34.6 \\
                            & Spoon             & \gradient{27.9}$\pm$ 48.6 \\
                            & Hand              & \gradient{26.6}$\pm$ 41.5 \\
                            & Fork              & \gradient{23.6}$\pm$ 38.7 \\
        \midrule
        \multirow{3}*{drink\_entered}      & \textit{All}      & \gradient{192.4}$\pm$ 222.1 \\
                            & Bottle            & \gradient{196.3}$\pm$ 206.2 \\
                            & Cup               & \gradient{187.2}$\pm$ 241.5 \\
        \midrule
        \multirow{3}*{drink\_lifted}       & \textit{All}      & \gradient{144.3}$\pm$ 204.3 \\
                            & Bottle            & \gradient{138.0}$\pm$ 188.9 \\
                            & Cup               & \gradient{154.1}$\pm$ 225.8 \\
        \midrule
        \multirow{3}*{drink\_to\_mouth}    & \textit{All}      & \gradient{143.8}$\pm$ 204.8 \\
                            & Bottle            & \gradient{137.1}$\pm$ 189.4 \\
                            & Cup               & \gradient{154.2}$\pm$ 226.3 \\
        \midrule
        napkin\_entered     & $\emptyset$       & \gradient{184.0}$\pm$ 253.5 \\
        napkin\_lifted      & $\emptyset$       & \gradient{134.1}$\pm$ 209.5 \\
        napkin\_to\_mouth   & $\emptyset$       & \gradient{132.6}$\pm$ 206.2 \\
        \bottomrule
    \end{tabular}
    }
    \end{minipage}\hspace{-1mm}
    \begin{minipage}{.51\textwidth}\vspace{-18.5mm}
    \renewcommand*{\minval}{0.3}
    \renewcommand*{\maxval}{11.3}
    \centering
    \scalebox{0.88}{
    \begin{tabular}{clc}
        \toprule
        Annotation sequence & Ann.~value & Time gap (s)\\ \midrule
        \multirow{5}{2.1cm}{\centering food\_entered \\ $\downarrow$ \\ food\_lifted}         
                            & \textit{All}      & \gradient{9.9}$\pm$ 27.3 \\
                            & Chopsticks        & \gradient{8.9}$\pm$ 24.0 \\
                            & Spoon             & \gradient{10.0}$\pm$ 19.5 \\
                            & Hand              & \gradient{9.9}$\pm$ 28.5 \\
                            & Fork              & \gradient{10.8}$\pm$ 31.9 \\
        \midrule
        \multirow{5}{2.1cm}{\centering food\_lifted \\ $\downarrow$ \\ food\_to\_mouth}       
                            & \textit{All}      & \gradient{1.8}$\pm$ 4.0 \\
                            & Chopsticks        & \gradient{1.3}$\pm$ 2.1 \\
                            & Spoon             & \gradient{1.5}$\pm$ 2.6 \\
                            & Hand              & \gradient{1.9}$\pm$ 4.3 \\
                            & Fork              & \gradient{2.3}$\pm$ 5.2 \\
        \midrule
        \multirow{5}{2.1cm}{\centering mouth\_open \\ $\downarrow$ \\ food\_to\_mouth}        
                            & \textit{All}      & \gradient{0.3}$\pm$ 0.2 \\
                            & Chopsticks        & \gradient{0.3}$\pm$ 0.1 \\
                            & Spoon             & \gradient{0.3}$\pm$ 0.1 \\
                            & Hand              & \gradient{0.3}$\pm$ 0.2 \\
                            & Fork              & \gradient{0.3}$\pm$ 0.2 \\
        \midrule
        \multirow{3}{2.1cm}{\centering drink\_entered \\ $\downarrow$ \\ drink\_lifted}         
                            & \textit{All}      & \gradient{9.1}$\pm$ 37.1 \\
                            & Bottle            & \gradient{7.3}$\pm$ 32.7 \\
                            & Cup               & \gradient{11.3}$\pm$ 41.8 \\
        \midrule
        \multirow{3}{2.1cm}{\centering drink\_lifted \\ $\downarrow$ \\ drink\_to\_mouth}       
                            & \textit{All}      & \gradient{4.2}$\pm$ 8.9 \\
                            & Bottle            & \gradient{5.1}$\pm$ 10.5 \\
                            & Cup               & \gradient{2.9}$\pm$ 5.2 \\
        \midrule
        \multirow{3}{2.1cm}{\centering napkin\_entered \\ $\downarrow$ \\ napkin\_lifted}     
                            & & \\
                            & $\emptyset$       & \gradient{3.0}$\pm$ 24.0 \\
                            & & \\
        \midrule
        \multirow{3}{2.1cm}{\centering napkin\_lifted \\ $\downarrow$ \\ napkin\_to\_mouth}   
                            & & \\
                            & $\emptyset$       & \gradient{1.5}$\pm$ 2.0 \\
                            & & \\
        \bottomrule
    \end{tabular}
    }
    \end{minipage}
    \label{tab:dataset:time_between_annotations}
    \vspace{-2mm}
\end{table}

\begin{figure}[t]
    \centering
     \includegraphics[width=0.495\textwidth]{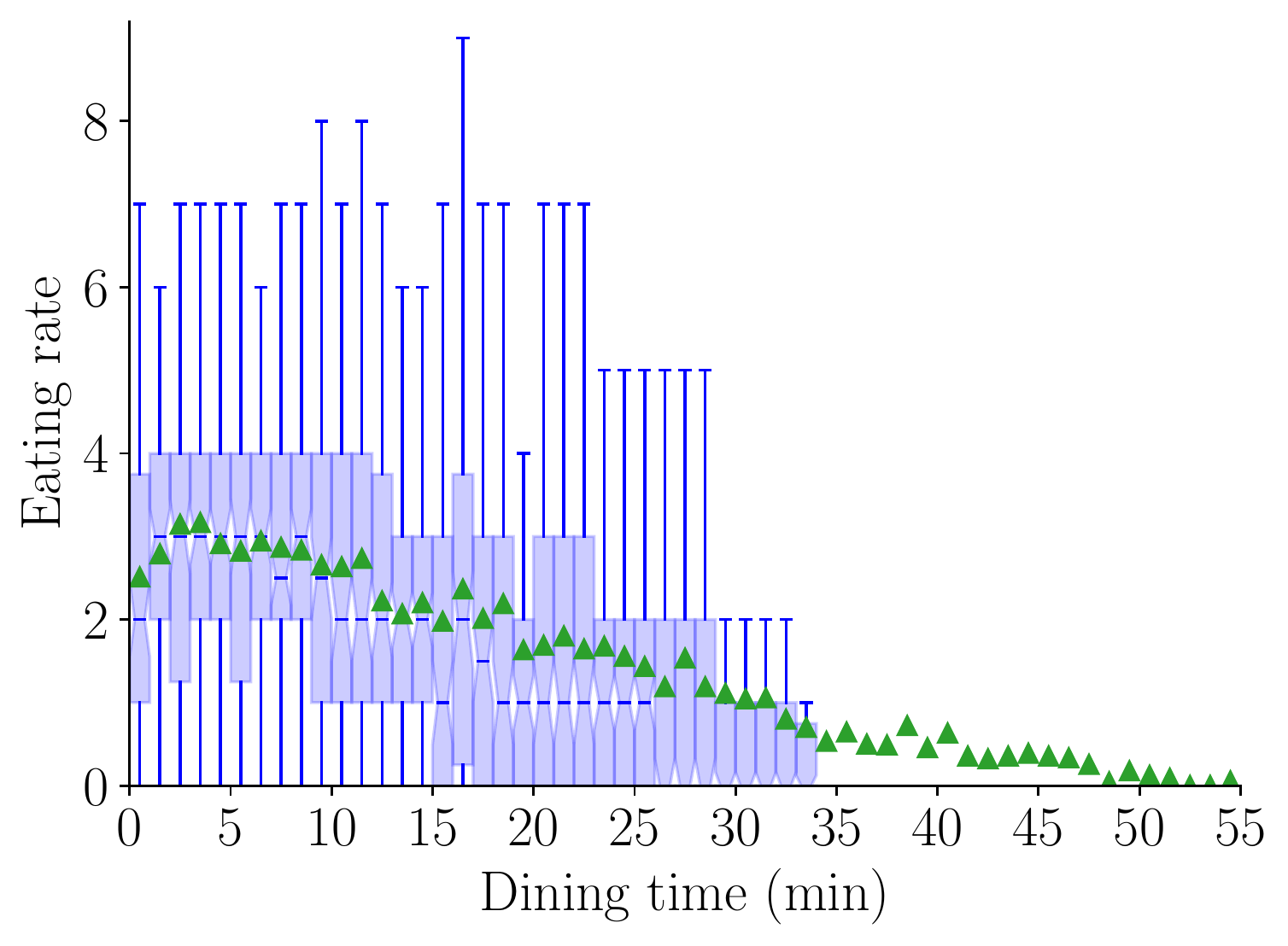}
     \includegraphics[width=0.495\textwidth]{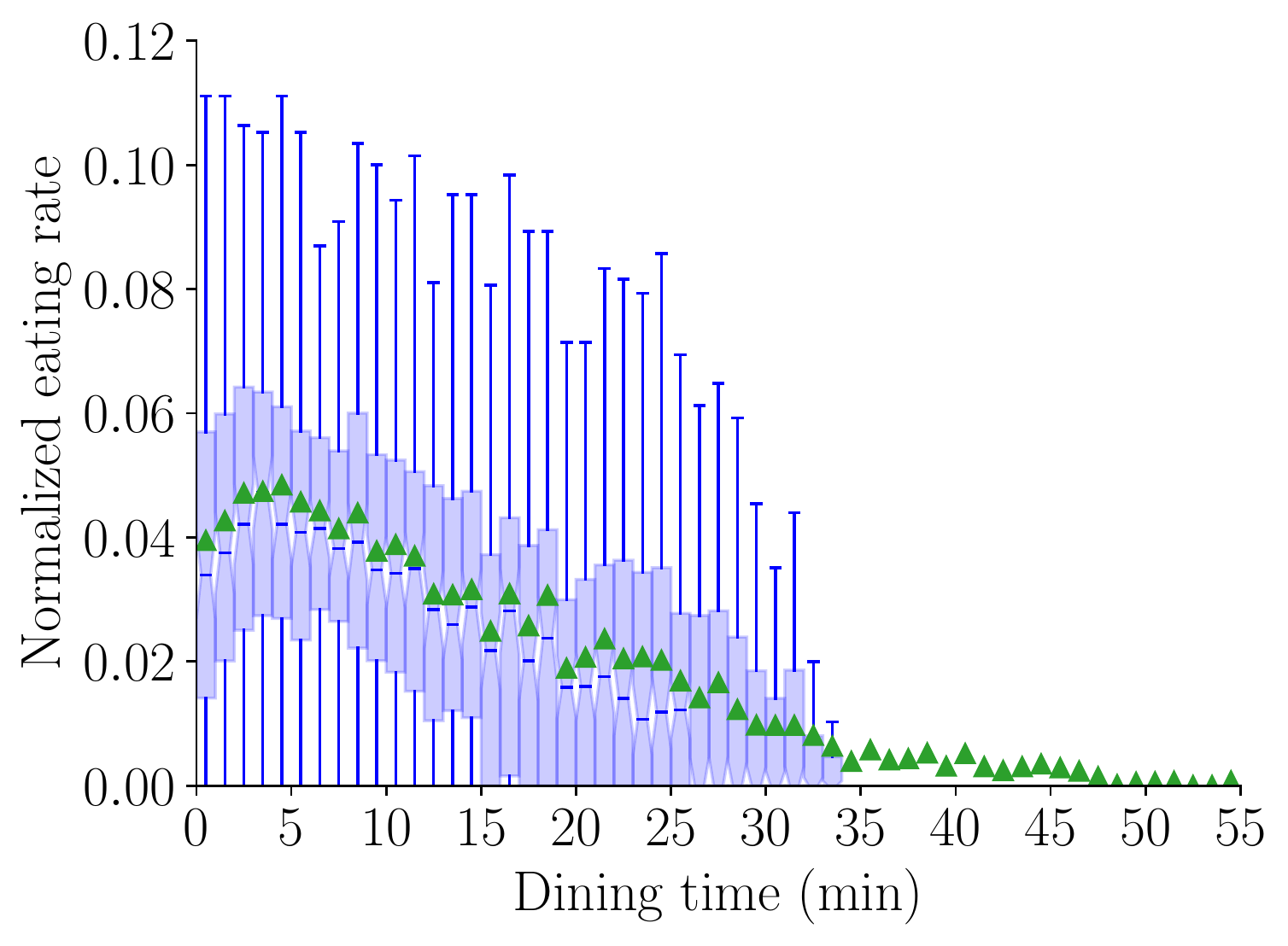}
    \caption{HHCD: Eating rate during dining.
        \textbf{Left:} Eating rate: number of eating actions per minute.
        \textbf{Right:} Normalized eating rate: number of eating actions per minute normalized by the total number of eating actions the diner made.
        One eating action corresponds to one \textit{food\_to\_mouth} annotation.
    }
    \label{fig:dataset:eating_rate}
\end{figure}

\textbf{Eating rate during dining.}
Figure~\ref{fig:dataset:eating_rate} (left) shows the eating rate (number of eating actions per minute) where one eating action corresponds to one \textit{food\_to\_mouth} annotation.
Since the number of eating actions might vary based on the total amount of food the diner had (and hence total number of eating actions they made), in Fig.~\ref{fig:dataset:eating_rate}~(right) we also normalize the eating rate by the total number of eating actions the diner made.
As we can see in both cases the eating rate increases from the start till around the 5th minute of dining time and decreases thereafter. 
This confirms the eating is a non-stationary activity and needs to be accounted for when designing models of commensality.

\begin{figure}[t]
    \centering
     \includegraphics[height=0.213\textheight]{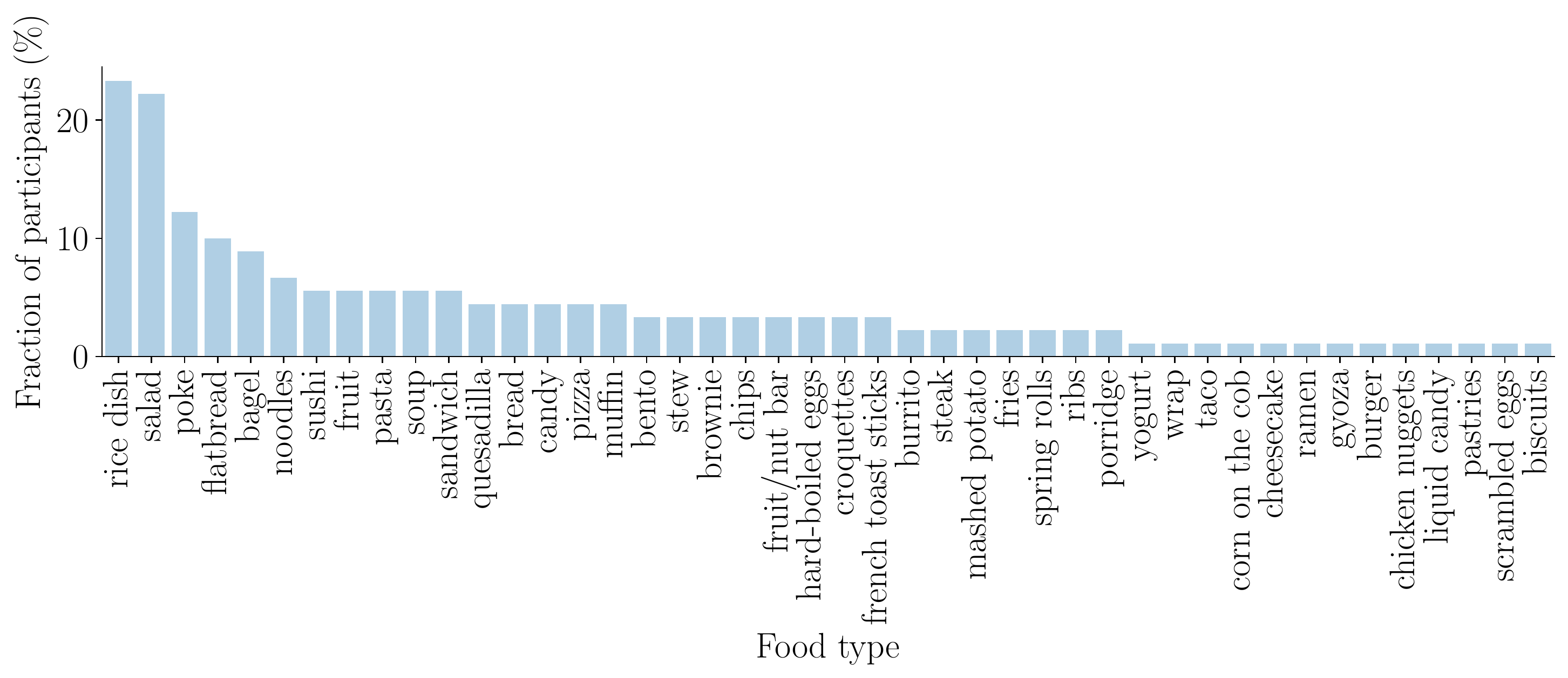}
     \includegraphics[height=0.213\textheight]{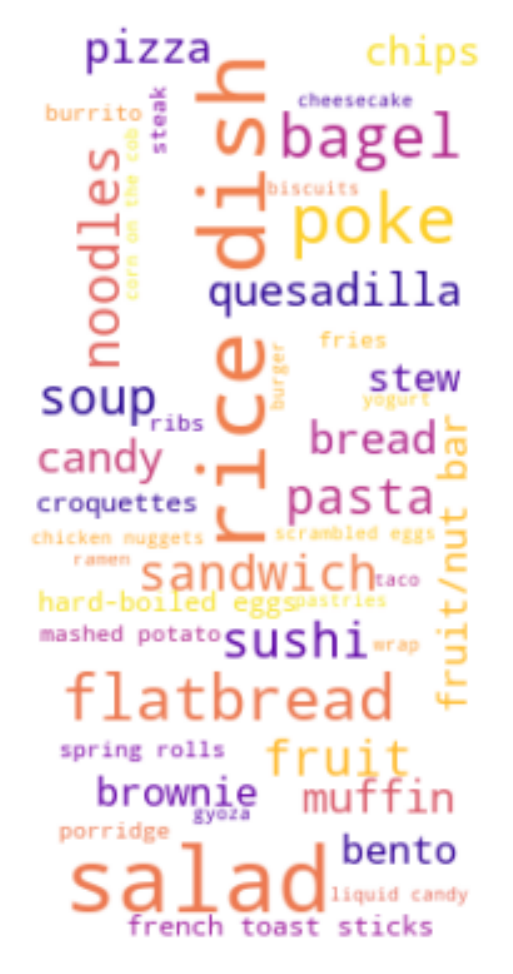}
    \caption{HHCD: Distribution of types of food the participants ate.
    Some participants ate multiple types of food.
    }
    \label{fig:dataset:stats:food_types}
    \vspace{-3mm}
\end{figure}

\textbf{Food types.}
The distribution of types of food the participants ate can be found in Fig.~\ref{fig:dataset:stats:food_types}.

\begin{figure}[t]
    \centering
    \includegraphics[width=0.495\textwidth]{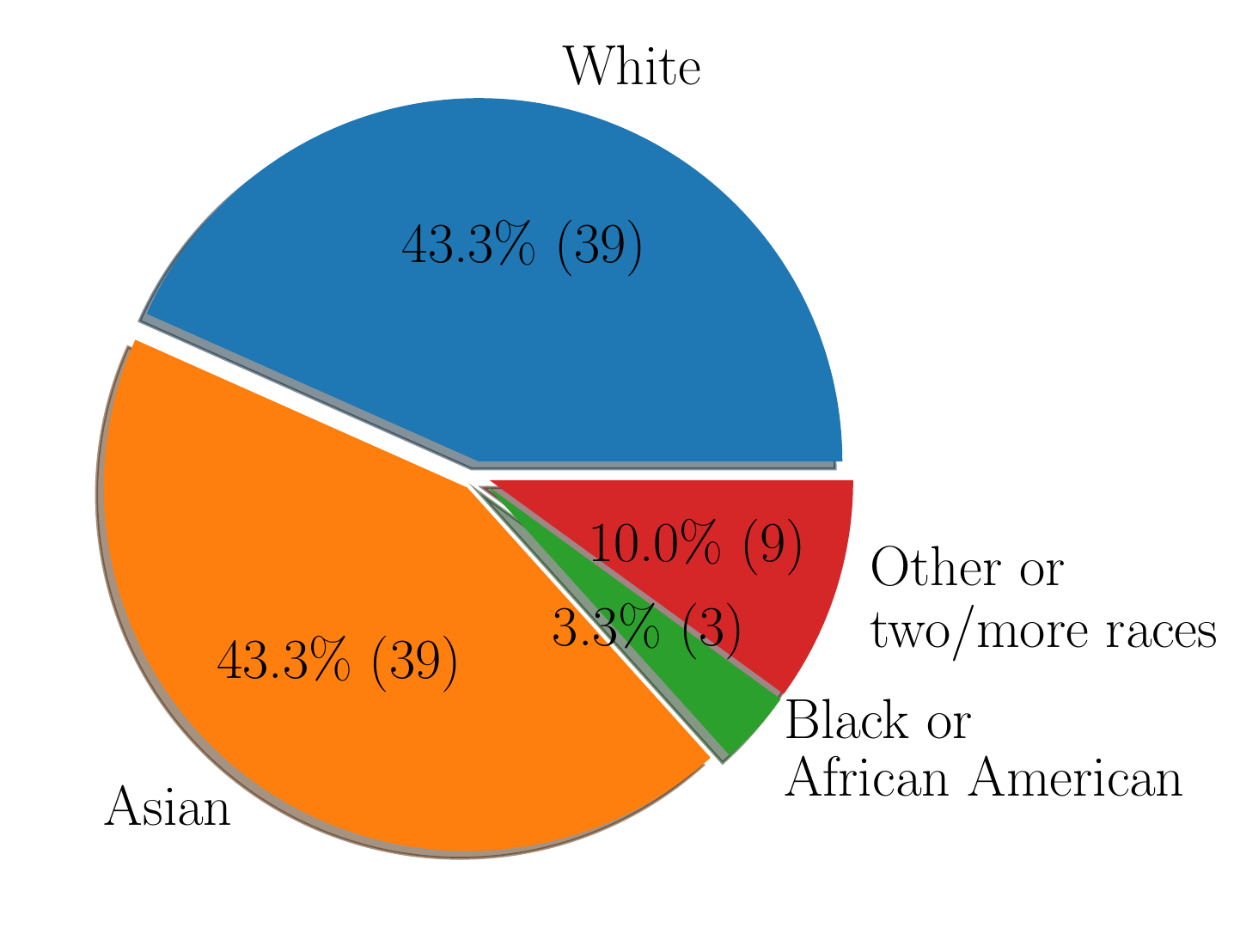}
    \includegraphics[width=0.495\textwidth]{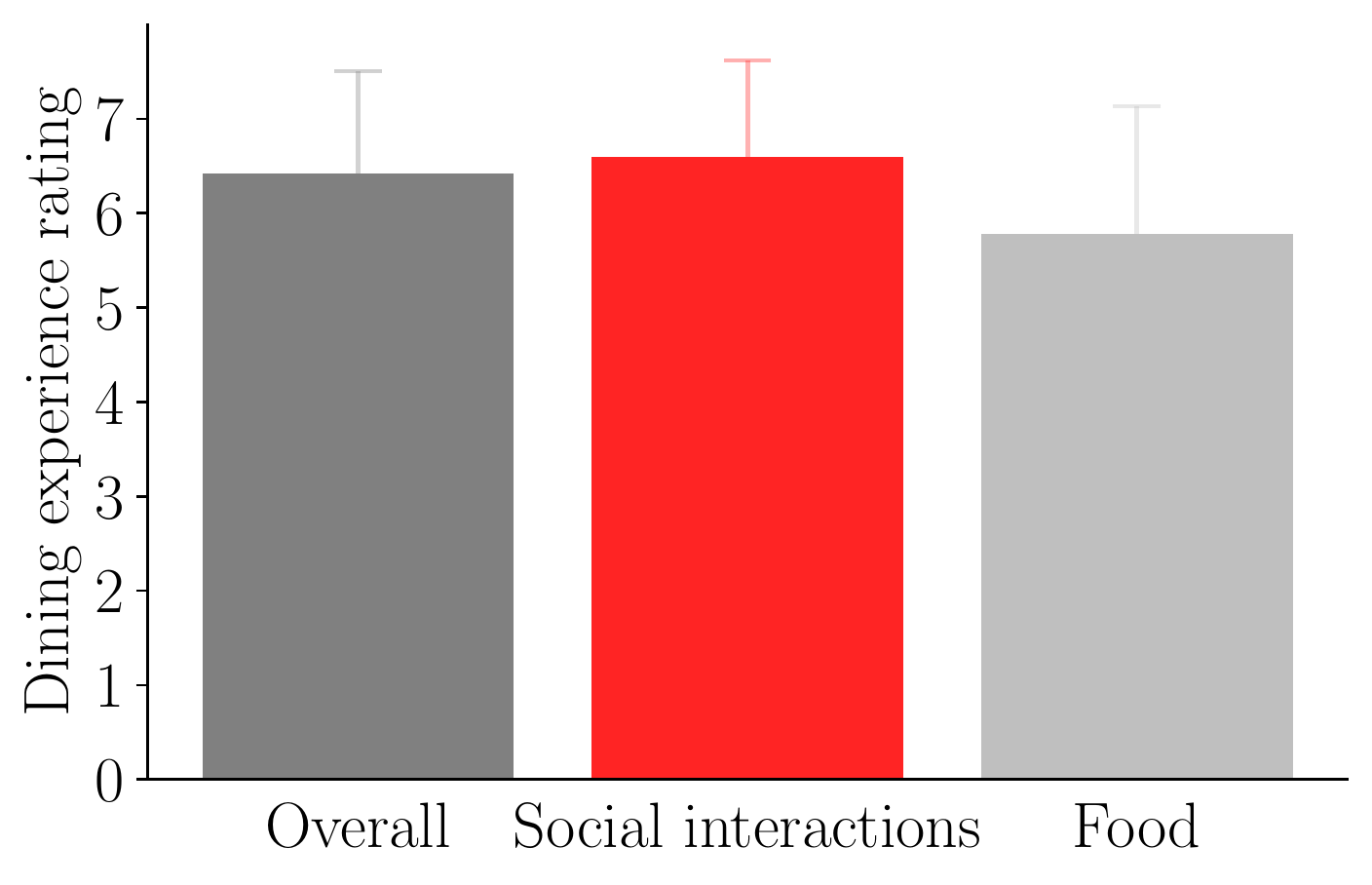}
    \caption{HHCD:
        \textbf{Left:} Distribution of participants' race. \textit{"Other or two/more races"} includes
        three White-Asians,
        three White-Hispanics/Latinos,
        two Latinos,
        and one Asian-Hispanic.
        \textbf{Right:} Participants' ratings of their overall meal experience, social interactions with other participants, and food on a Likert scale 1-7 (Strongly disagree - Strongly agree with positive experience).
    }
    \label{fig:dataset:race_and_dining_experience}
    \vspace{-3mm}
\end{figure}

\textbf{Demographic background.}
82 participants were right-handed and 8 left-handed. 
The distribution of participants' race is shown in Fig.~\ref{fig:dataset:race_and_dining_experience} (left). 

\begin{figure}[t]
    \centering
     \includegraphics[width=0.495\textwidth]{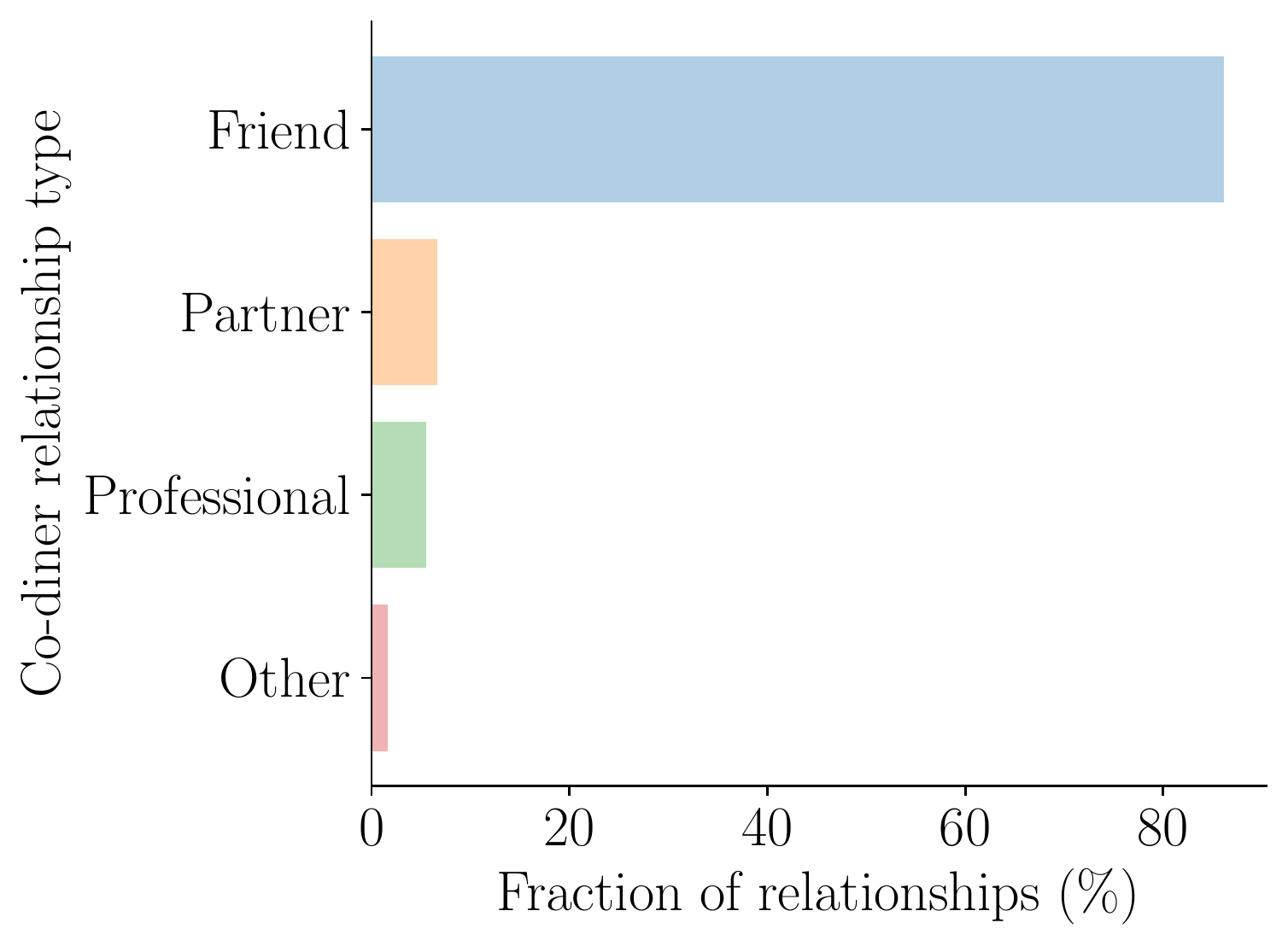}
     \includegraphics[width=0.495\textwidth]{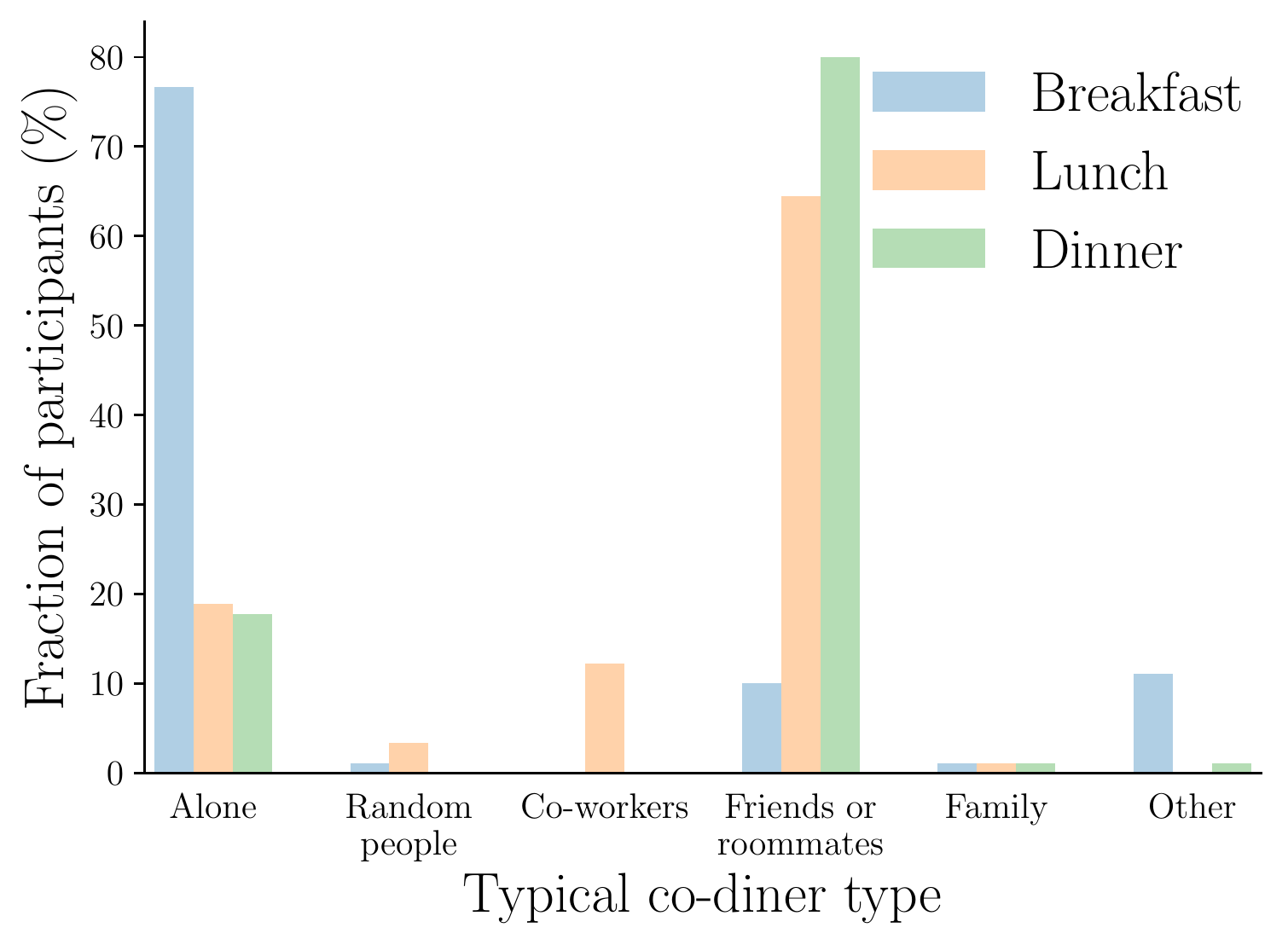}
     \includegraphics[width=0.495\textwidth]{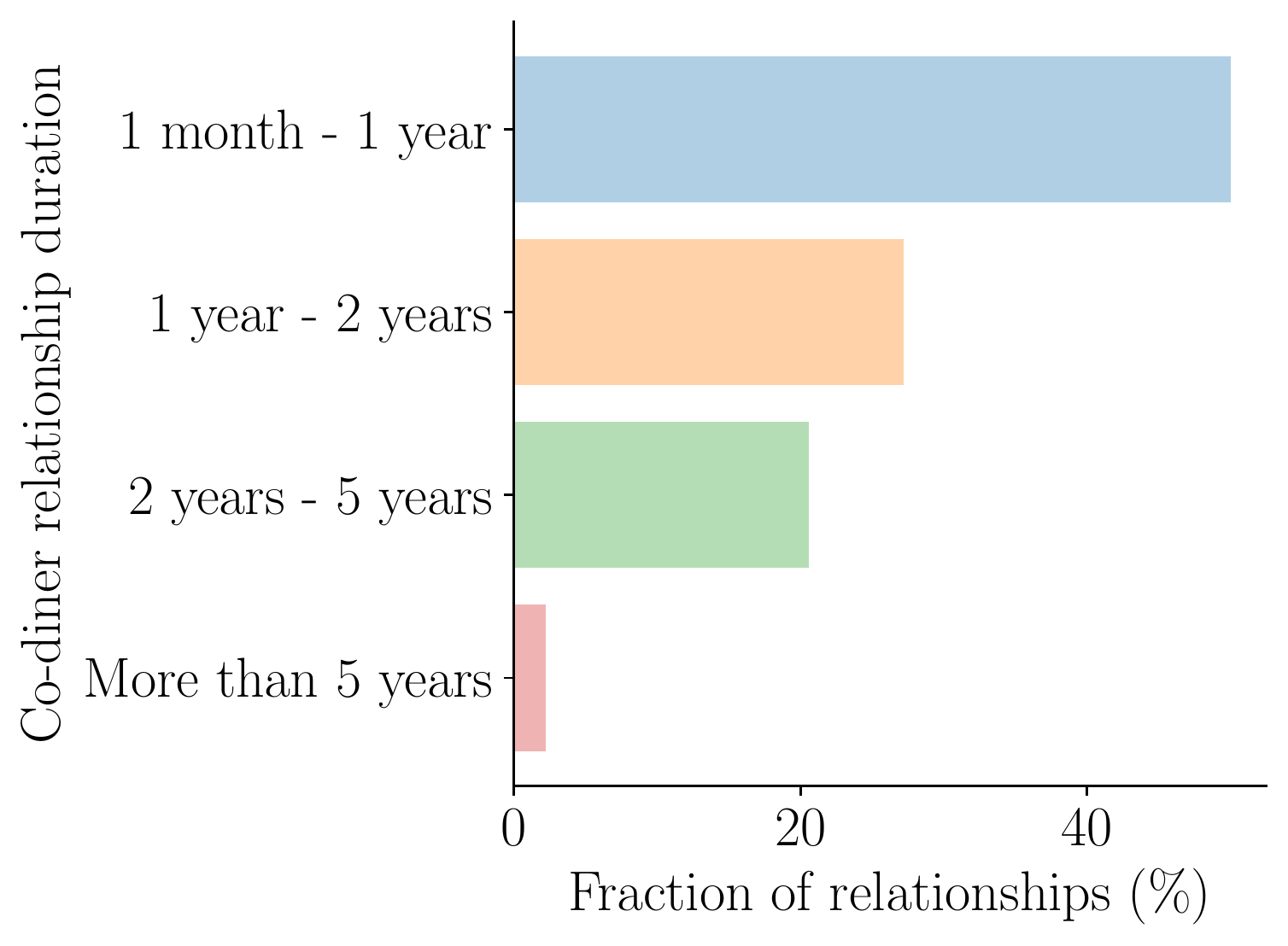}
     \includegraphics[width=0.495\textwidth]{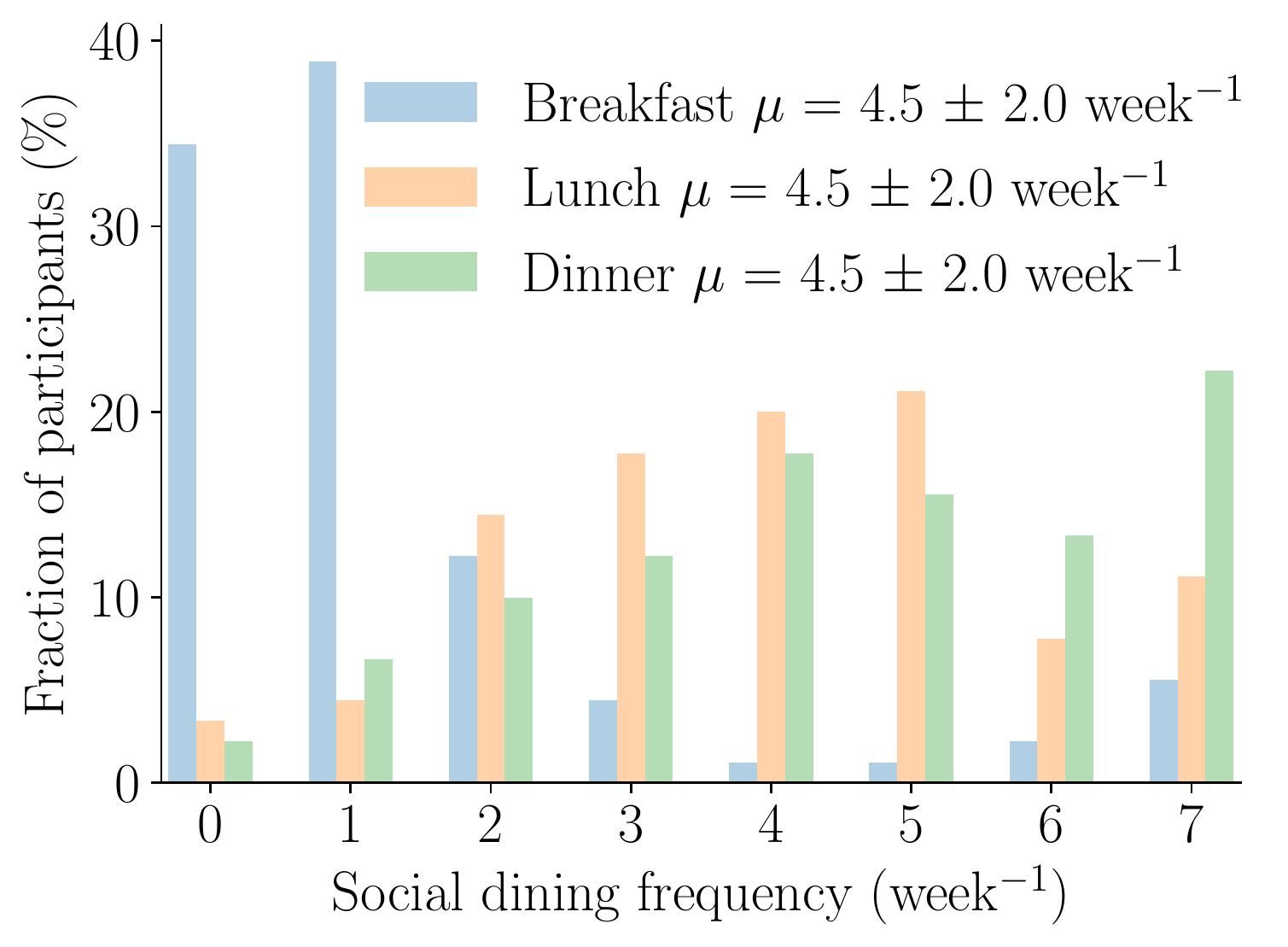}
     \includegraphics[width=0.495\textwidth]{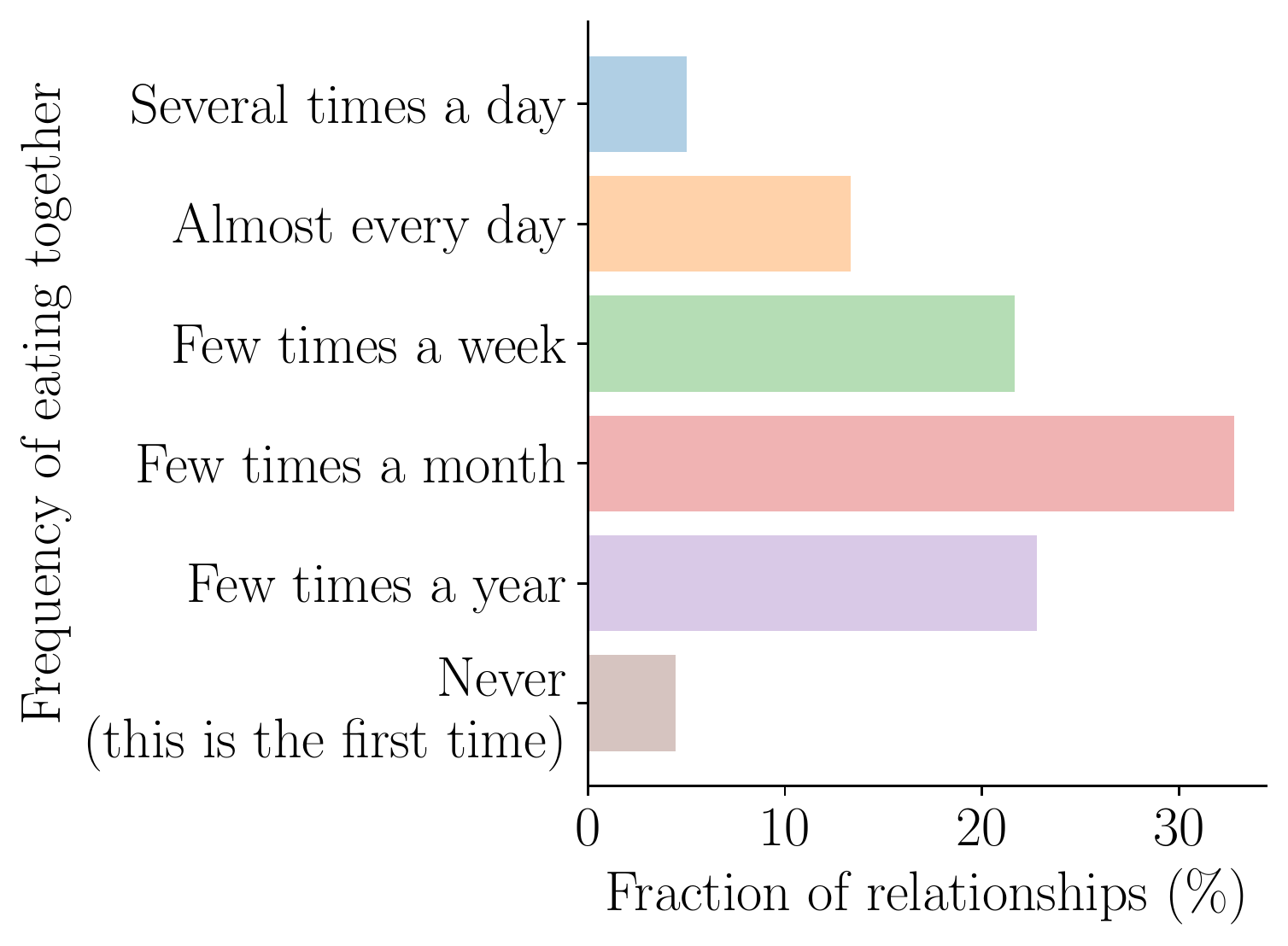}
     \includegraphics[width=0.495\textwidth]{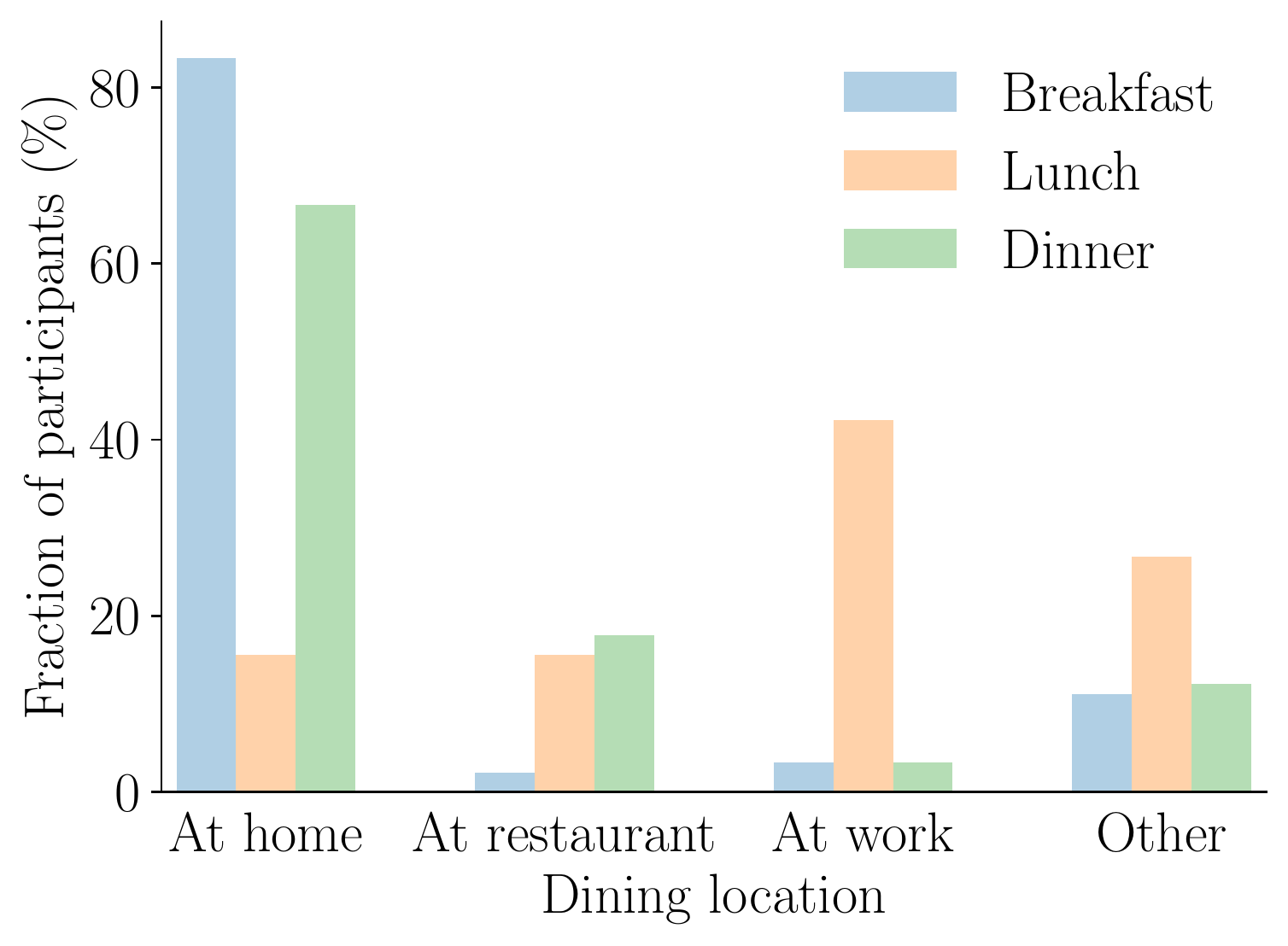}
    \caption{HHCD: 
        \textbf{Left top-bottom:} Distributions of co-diner relationship types, durations, and frequency of eating together. 
        \textit{"Other"} co-diner relationship type includes two acquaintances and one boyfriend.
        \textbf{Right top-bottom:} Distributions of participants' typical co-diner type, social dining frequency, and dining location.
        \textit{"Other"} typical co-diner type includes partner or n/a for skipped breakfasts.
        \textit{"Other"} typical dining locations include dining hall, campus, n/a for skipped breakfast, or a friends' place for dinner.
    }
    \label{fig:dataset:stats:relationships_and_habits}
    \vspace{-6mm}
\end{figure}

\textbf{Relationship between diners.}
The distributions of co-diner relationship types, durations, and frequency of eating together are provided in Fig.~\ref{fig:dataset:stats:relationships_and_habits} (left top-bottom) respectively.

\textbf{Social dining habits.}
The distributions of participants' typical co-diner type, social dining frequency, and dining location are shown in Fig.~\ref{fig:dataset:stats:relationships_and_habits} (right top-bottom) respectively.

\textbf{Dining experience.}
Participants' ratings of their overall meal experience, social interactions with other participants, and food are presented in Fig.~\ref{fig:dataset:race_and_dining_experience} (right).

\textbf{Replies to open-ended post-study questions.}
We also analyze the study participants' answers to open-ended questions in the post-study questionnaire (Fig.~\ref{fig:appendix:dataset:collection:poststudy_questionnaire}~(right)).
We observe the following patterns.

\textcolor{purple}{\textasteriskcentered{} \textit{When participants think it is appropriate to take a bite of food when they are eating with others}}
\begin{itemize}
    \item \textbf{Talking-related rules:} \textit{"When I am not speaking",
    "When listening to others",
    "When others are talking or if there is a pause in the conversation",
    "After sharing a long piece of speech and expecting a lot of response",
    "When someone else is talking and I don't think they're going to ask me anything",
    "It is appropriate when someone is not talking about a very serious topic you need to give your full attention to."
    }
    
    \item \textbf{Eye gaze-related rules:} \textit{"When the person talking is not making eye contact",
    "\dots when i'm not making direct eye contact with someone, \dots"
    }
    
    \item \textbf{Diner physical state-related rules:} \textit{"when you are hungry, it should be ok to take a bite of food."
    }
    
    \item \textbf{Social interaction-related rules:} \textit{"\dots when other people are taking a bite too",
    "It is appropiate when my bite it is at the same time when the others are putting food in their mouth. \dots",
    "\dots when two other people are having a subconversation that I am less engaged in"
    }
    
    \item \textbf{Time-related rules:} \textit{"every 10 seconds or so, \dots",
    "\dots when a lot of time has passed between your previous bite"
    }
    
    \item Several participants also replied with \textit{"whenever i want"} or similarly.
\end{itemize}

Note, the replies to the bite timing questions align with choices of modalities and features we use for bite timing prediction.

\textcolor{purple}{\textasteriskcentered{} \textit{What participants liked about the meal experience}}
\begin{itemize}
    \item Most participants liked \textbf{food, conversation, and time spent with friends}. 
    For example, \textit{"It was super interactive and I got to know my friends better"}
    
    \item \textbf{Research contribution:} \textit{"Time with friends, spicy foods, contributing to research", 
    "fun experience to help the robots take over the world"
    }
    
    \item \textbf{The study environment:} \textit{"It felt comfortable and natural and the food was yummy.",
    "Felt like a natural interaction",
    "I enjoyed the food, being able to soley talk to my friends without distraction",
    "food was good, after getting used to cameras conversation felt pretty natural"
    }
\end{itemize}

\textcolor{purple}{\textasteriskcentered{} \textit{What participants did not like about the meal experience}}
\begin{itemize}
    \item 18 participants (20\%) replied \textit{"nothing"} or similarly.
    
    \item \textbf{Complaints about food they brought:} \textit{"We didn't buy enough food.",
    "i ate too much and my stomach hurts",
    "one friend talked too much, it was a bit long, I ordered too much food and did not eat all of it.",
    "The pizza was slightly cold and i ordered the wrong pizza from domino pizza company."
    }
    
    \item \textbf{The study room and the recording setup:} \textit{"I think I would rather be in a more comfortable chair and have lower lighting",
    "The room was too quiet for my comfort",
    "It was in an enclosed room. The physical setting didn't feel natural.",
    "A little conscious of the camera",
    "It was a little odd to be monitored the whole time",
    "I was nervous speaking about somethings because it was recorded",
    "The camera directly in our faces",
    "Not much! Cameras in the middle of the table made it slightly more awkward to pass food, I guess.",
    "Maybe the fact that we were being watched, recorded; felt a like bit performative",
    "I did not like that I felt that I had to lean backwards to fit in the frame",
    "We definitely knew and acted like we were being recorded at times",
    "I think just because we were participating in the study but I didn't feel uncomfortable with the cameras or anything. So, I feel like our dinner was still authentic."
    }
    \item \textbf{Conversation topics:} \textit{"Participant 2 was talking too much about politics that were boring."}
    
    \item \textbf{Dining duration:} \textit{"The amount of time I spent could have been longer to have more of a conversation"}
    
    \item \textbf{Eating with others:} \textit{"I did not choose the food we ordered and I didn't enjoy the food very much, and I get embarrassed eating around others"}
    
    \item \textbf{Use of mobile devices while eating:} \textit{"some things that i dont like about the meal is sometimes people tend to still like using their devices, which makes it feel like they dont want to be there eating a meal with you"}

\end{itemize}

\subsection{Human-Human Commensality Model Experiments}\label{sec:appendix:model_exps}


\subsubsection{Feature Extraction}
We utilize several feature extraction techniques to obtain various high-level features that might indicate semantic visual and audio cues. We combine these features from each person and align target user with two co-diners for each sample event.
\begin{enumerate}
    \item \textbf{Visual features:} Video clips from cameras facing diners explicitly capture dining behaviours and social interactions. We estimate people's body, hand and face skeletons using OpenPose~\cite{cao2018openpose} across consecutive frames. Each frame at time $t$ contains body gesture and face representation as a 168-dimensional vector $o \in \mathbb{R}^{168}$. 
    Gaze plays a crucial role during communications and interactions. 
    It is a predictor of participants' interests in human-robot interactions~\cite{fischer2016markerless}. 
    We extract participants' gaze and head pose directions using Real-Time Eye Gaze and Blink Estimation in Natural Environments (RT-GENE)~\cite{fischer2018rt}. 
    Gaze and head pose direction data points are represented by Euler angles $\theta$ and $\phi$, and together form the feature $d \in \mathbb{R}^{4}$. 
    
    \item \textbf{Audio features:} Using ReSpeaker Mic Array v2.0~\cite{micarray} we extract raw audio (a mixture of three diners' voices) and a sound direction channel from ROS messages.
    We use ROS messages as they can be naturally transferred to a robot.
    We align these ROS messages to video frames using nearest neighbor based on the video frame and audio message timestamps.
    There can be repeated audio frames aligned to video frames due to audio messages varying in speed.
    For each aligned audio message, we apply voice activity detection using WebRTC VAD~\cite{webrtcvad} and use k-means clustering on the sound direction information to localize speakers in the scene. 
    We combine the directional clusters and detected voice activity to create a binary vector that indicates whether a diner is speaking or not at each video frame.
    We also refer to this binary feature as speaking status.

    \item \textbf{Temporal features:} Upon analyzing eating rate in HHCD (App.~\ref{sec:appendix:dataset:stats}), we notice that the participant's eating rate increases a bit at the beginning and then decreases as the dining comes to the end. 
    Therefore, we believe that explicitly providing the model with time and bite count information can better capture the non-stationary nature of commensality.
    We thus generate two bite features $b \in \mathbb{R}^{2}$, which indicate the time since the last bite of food was taken and the number of bites a person has consumed during the eating session.

\end{enumerate}
\vspace{1mm}

\subsubsection{Design Choices for Couplet-SoNNET} \label{couplet_sonnet_design}
 
We chose to restrict the features of the user in Couplet-SoNNET due to a distribution shift between human-human commensality and human-robot commensality. We chose to remove most social signals from the user because it would be more generalizable across our target population. Whether a user is talking could be relevant to predicting bite timing. During preliminary testing however, we found that modeling the user's speaking status led to the model never feeding at all if they kept talking. This makes sense as talking is highly correlated with not-feeding in HHCD. Since the user is not self-feeding, they are not incentivized to stop talking. Therefore, we believe some level of coercion is required to ensure the user is fed, which we realize is a common subtle practice when we spoke with the caregivers who feed care recipients. By removing the user's speaking status, we can ensure that feeding does occur.

\subsubsection{Implementation Details of the SoNNET}
\label{sec:appendix:sonnet}
Both Triplet-SoNNET and Couplet-SoNNEt are trained using an Adam optimizer with a learning rate of 0.0001 and a batch size of 128.
To prevent overfitting, we early stop if the validation loss does not increase after 10 epochs. The number of filters at each convolutional layer can be seen in Fig.~\ref{fig:methods:sonnet}. We use batch normalization layers after each convolutional layer.
All experiments are performed on a 64-core cluster with five NVIDIA RTX 3090s.

\subsubsection{Implementation Details of the Baseline Models}
\label{sec:appendix:baselines}
We use the Keras TCN implementation~\cite{KerasTCN} and train the Triplet-TCN and Couplet-TCN using the same hyperparameters as the SoNNET models.
We set the filter size to 50, which ensures a similar number of learnable parameters as the SoNNET models. 
In the case of the Triplet-TCN, we simply concatenate all the features of all three participants, while for the Couplet-TCN, we use features of the co-diners and only the bite features of the User.



\subsection{Human-Robot Commensality (HRCom) Experiments}
\label{sec:appendix:user_study}

\subsubsection{Study Design Rationale}\label{sec:appendix:user_study:rationale}
We considered various experimental designs for our user study. Our experiment design is a within-subjects repeated-measures design where the conditions are counterbalanced such that A→B and B→A occur a total of 3 times and there is only 1 bite per condition at one time. This helps mitigate the recency tendency and guarantees that within each session, there is a tie-breaker. Within one session, there are 9 forced-choice comparisons from 10 trials. Across 10 sessions we ensure that each ordered pair occurs an equal number of times. This gives us a total of 30 comparisons, 15 A→B and 15 B→A (and similarly for BC, AC). With these forced-choice questions, this study design is generally better for eliciting preference data \cite{sankaran2021curmelo} and has less recency bias \cite{mehrani2015recency} than alternative study designs where conditions are presented repeatedly at a time. For example, an alternative study design could have $X$ bites of condition A, followed by $X$ bites of condition B, and $X$ bites of condition C with a preference question at the end. This alternative study design would exacerbate the recency effect, as the participants would have to remember what they felt several bites ago. We also considered other similar alternative study designs but settled with the one we presented as we believe this  design would represent people’s preferences in a sample-efficient manner with less recency biases.

\subsubsection{Bite Timing Strategy Details}\label{sec:appendix:user_study:conditions}
We designed our three bite timing strategies (Learned Timing, Fixed-Interval Timing, and Mouth-Open Timing) based on discussions with care recipients, occupational therapists, and caregivers. 

While consulting people with mobility limitations, caregivers, and occupational therapists on what features we should look at based on what movements are consistent across people with mobility limitations. Our target users (with C3-C5 SCI) cannot move their arms to feed themselves. Also, there is a huge spectrum of severity of mobility limitations depending on the users’ conditions, and their movements are not consistent across these users. Therefore, the Learned Timing strategy uses Couplet-SoNNET which does not use arm gesture features but uses only the features of the co-diners to make it more generalizable across this target population.

For the Mouth-Open Timing, caregivers said that they estimate the appropriate time to feed when care recipients open their mouth and provide an explicit cue. This directly inspired the design of our Mouth-Open Timing. 
As described in App.~\ref{sec:appendix:user_study:rationale}, our study design mitigates the recency effect and ensures useful comparisons between conditions.
Since a new condition is presented after each bite, we use a speaker to prompt the user to open their mouth when they are ready.

To decide on the wait time for the Fixed-Interval Timing, we used data from HHCD to find a user-inspired wait-time. We found that a human on average takes 1.8s from lifting a food item off a plate / bowl (food\_lifted) to bringing it to the mouth (food\_to\_mouth). The robot’s equivalent approach duration is on average around 9 seconds (taking into account the variable motion planning time). Though the robot could mechanically move at a faster speed, we chose the speed that would feel safe and comfortable to a user when they are being approached (fed) by a robot with a fork. We determined this velocity of our robot to be perceived as safe and comfortable based on \cite{bhattacharjee2020more} which explicitly did a study on what approach speeds are preferred by people with mobility limitations. This scaling factor (9s / 1.8s = 5) between robot speeds and human speeds is thus user-inspired. Once we determined this scaling factor, we use this same scaling factor to scale up the bite timing from HHCD (9.9s) to human-robot commensality (9.9s*5 - 5s [for robot bite acquisition to bite-timing waiting position] = 44.5s) to make sure that the proportion of time for different phases of feeding (bite acquisition - bite timing - bite transfer) are all proportional and balanced. We would also like to note that although the average time for “food\_entered → food\_lifted” was 9.9s in HHCD, the standard deviation was 27.3s. So a wait-time of 44.5s is roughly 1 standard deviation away from the equivalent annotation in the HHCD data.

\subsubsection{Experimental Setup Details}\label{sec:appendix:user_study:setup_details}
The experiment was set up similarly the human-human commensality dataset collection described  App.~\ref{sec:appendix:dataset:measures} and is depicted in Fig.~\ref{fig:appendix:exp_setups}.
For the Learned timing, RT-GENE~\cite{fischer2018rt} and OpenPose~\cite{cao2018openpose} need to process video streams from all three cameras in real-time, in addition to the robot's planning and perception stack. 
We thus distribute compute over two machines: a 24-core PC with an NVIDIA RTX 3060 and a 32-core PC with an NVIDIA RTX 3090. 
We downsample the 30 FPS video streams to 15 FPS to ensure real-time performance. 

As noted in the formulation of the Fixed-Interval timing strategy (Sec.~\ref{sec:user_study}), the robot is $5x$ slower during feeding as compared to a human.
This means there is a distribution shift in the time since the last bite was taken on the robot compared to the training data for Couplet-SoNNET.
To mitigate this distribution shift, we scale down the computed time since the last bite during the user study by a factor of 5.

The robot used joint space velocity control. 
The robot's motion was generated from different calls to a library of planners available to our platform:
\begin{itemize}
    \item \texttt{planToConfiguration(goal\_config)}: plans from current configuration to a joint space goal configuration (6 degrees-of-freedom)
    \item \texttt{planToTaskSpaceRegion(ee\_goal\_pose, constraints)}: plans from current configuration to a task space end-effector (EE) goal pose with some given constraints [1]
    \item \texttt{planToEEOffset(ee\_offset)}: plans such that the end-effector moves in the direction of a certain vector.
\end{itemize}

It is paramount for the our robot platform to ensure safety to the user, so we familiarized participants to the four levels of safety we designed:
\begin{enumerate}
    \item We placed a conservative collision model around the user’s head. The users in our study were familiar with the general workspace of the robot (we moved the robot while they were seated on the chair as a part of the pre-study familiarization procedure).
    \item The fork has a Force/Torque sensor attached to it, where if a certain threshold of force is reached (beyond acceptable safety / comfortable force thresholds), the arm stops immediately.
    \item We had an observer watch the experiment while the emergency stop was ready to press in the case of unexpected behaviors. Additionally, an experimenter watched the system and was ready to stop it for safety.
    \item The compliant robot arm is also set up so that the user can stop it if absolutely necessary. We also designed the speed of the robot to be at comfortable levels.
\end{enumerate}

\subsubsection{Experimental Procedure Details}\label{sec:appendix:user_study:procedure_details}
Each participant was compensated for each hour of their time and for their food expenses. 
All participants were instructed to bring their own food. 
The user who was fed by the robot only ate fruits.
Each of the other two participants could choose if they also want to eat fruits during the study or the food they brought. 

Before starting the study, we familiarized the participants with robot-assisted feeding by showing them a trial of the Mouth-Open condition and shortened Fixed-Interval condition, along with explaining safety measures. 
The procedure than continues as described in Sec.~\ref{sec:user_study}.

\subsubsection{Conversation Starters}\label{sec:appendix:user_study:prompts}
List of questions that the user study participants could optionally use to help get the conversation started at each trial, similar to the past work~\cite{herlant2018algorithms}.

\begin{itemize}
    \item What are you studying?
    \item Who is your favorite singer and why?
    \item What is your favorite food and why?
    \item What is your favorite color and why?
    \item Do you give back or volunteer with any organizations?
    \item What are your favorite writers and books?
    \item Do you have any pets and if so, what are they?
    \item What sports do you play or watch and why?
    \item What is your favorite movie and why?
    \item Who is your favorite actor and why?
    \item Which languages do you speak and which ones do you want to learn?
    \item What was your favorite vacation?
    \item What are your hobbies?
\end{itemize}

\subsubsection{Questionnaires}\label{sec:appendix:user_study:questionnaires}
The questions we asked the participants in the pre-study questionnaire included all the questions asked during data collection (Fig.~\ref{fig:appendix:dataset:collection:prestudy_questionnaire}) and an additional question about the participant's level of hunger (Fig.~\ref{fig:appendix:user_study:user_study_questionnaire}~(a)).
The questions in the experiment questionnaire we asked after each trial are shown in Fig.~\ref{fig:appendix:user_study:user_study_questionnaire}~(b) and the final post-study questionnaire at the very end of the study is shown in Fig.~\ref{fig:appendix:user_study:user_study_questionnaire}~(c).

\vspace{2mm}
\subsubsection{Additional Results}\label{sec:appendix:user_study:results}

\begin{figure}[h]
    \centering
    \includegraphics[width=0.495\textwidth]{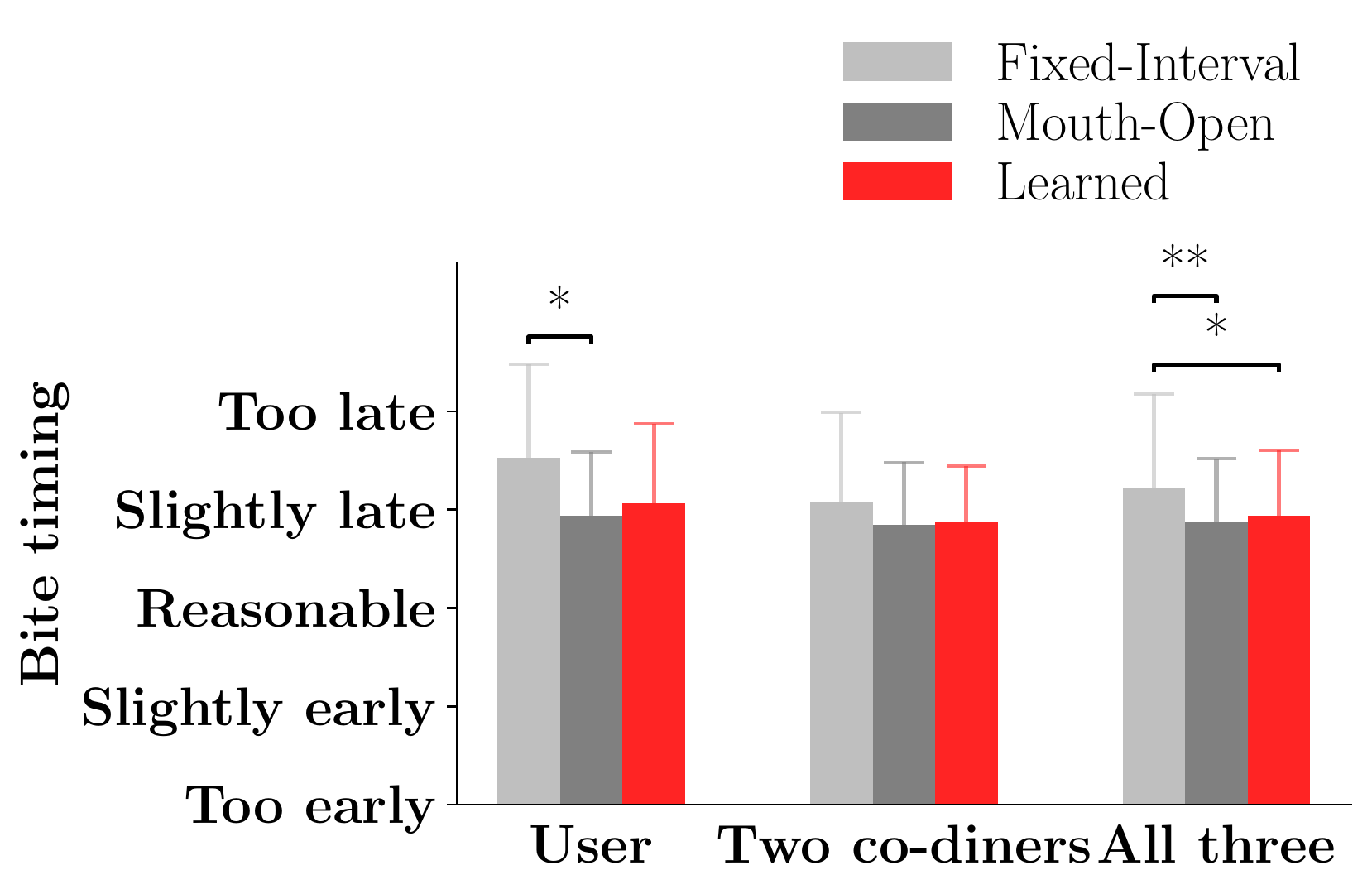}
    \includegraphics[width=0.495\textwidth]{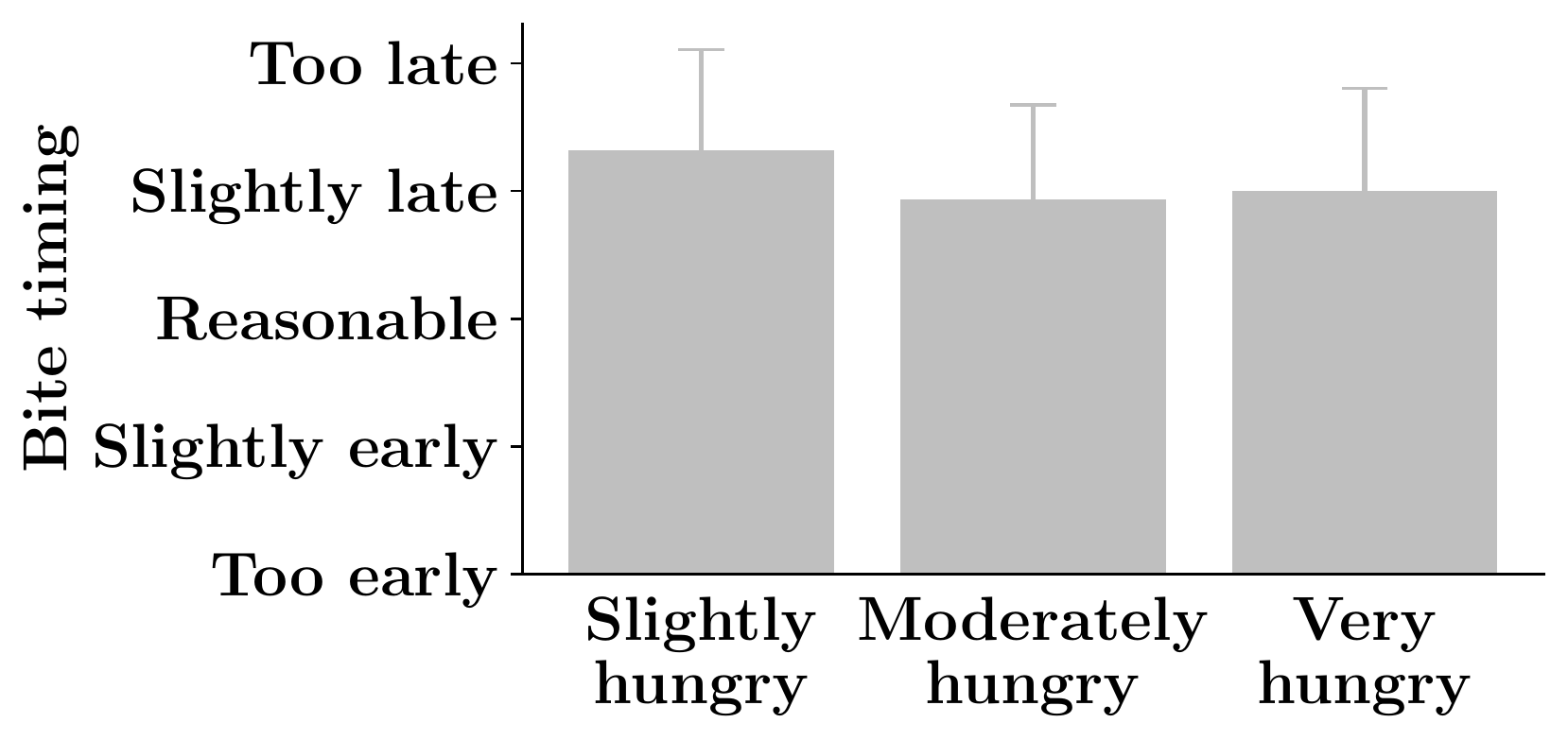}
    \caption{
        HRCom: 
        \textbf{Left:} Bite timing perceived by users, two co-diners, and all three diners on a Likert scale 1-5 (Too early - Too late), for each bite timing strategy (Fixed-Interval, Mouth-Open, and Learned timing).
        $*, **$ denote statistically significant differences with $p_{0.05}, p_{0.005}$ respectively.
        \textbf{Right:} Effect of robot users' hunger level on their bite timing ratings. 
        Our study did not find any statistically significant differences with $p_{0.05}$.
    }
    \label{fig:user_study:exp:bitetiming}
    \vspace{2mm}
\end{figure}

\textbf{Bite timing.} Besides the forced-choice assessment of bite timing strategies in terms of bite timing in Sec.~\ref{sec:user_study:results}, we also evaluate absolute ratings of "how timely" each trial was. 
In fact, one robot user reported that \textit{"Slight timing changes seemed more noticeable than I expected."}
As we can see from Fig.~\ref{fig:user_study:exp:bitetiming}~(left), the only statistically significant differences are with respect to Fixed-Interval timing suggesting that the user as well as all three diners found this strategy feeds rather late compared to other strategy/ies.
It might be interesting to further evaluate whether diners would prefer Fixed-Interval timing with a higher feeding frequency. 

We also investigate whether the robot users' pre-study hunger level affected their bite timing ratings. 
As shown in Fig.~\ref{fig:user_study:exp:bitetiming}~(right), we do not find any statistically significant differences with $p_{0.05}$ between the three hunger levels users reported.
This could suggest that their bite timing ratings were not biased by their hunger level.
However, we cannot draw any strong conclusions as the hunger level self-assessment is a very subjective metric.

\begin{figure}[t]
    \centering
     \includegraphics[width=0.495\textwidth]{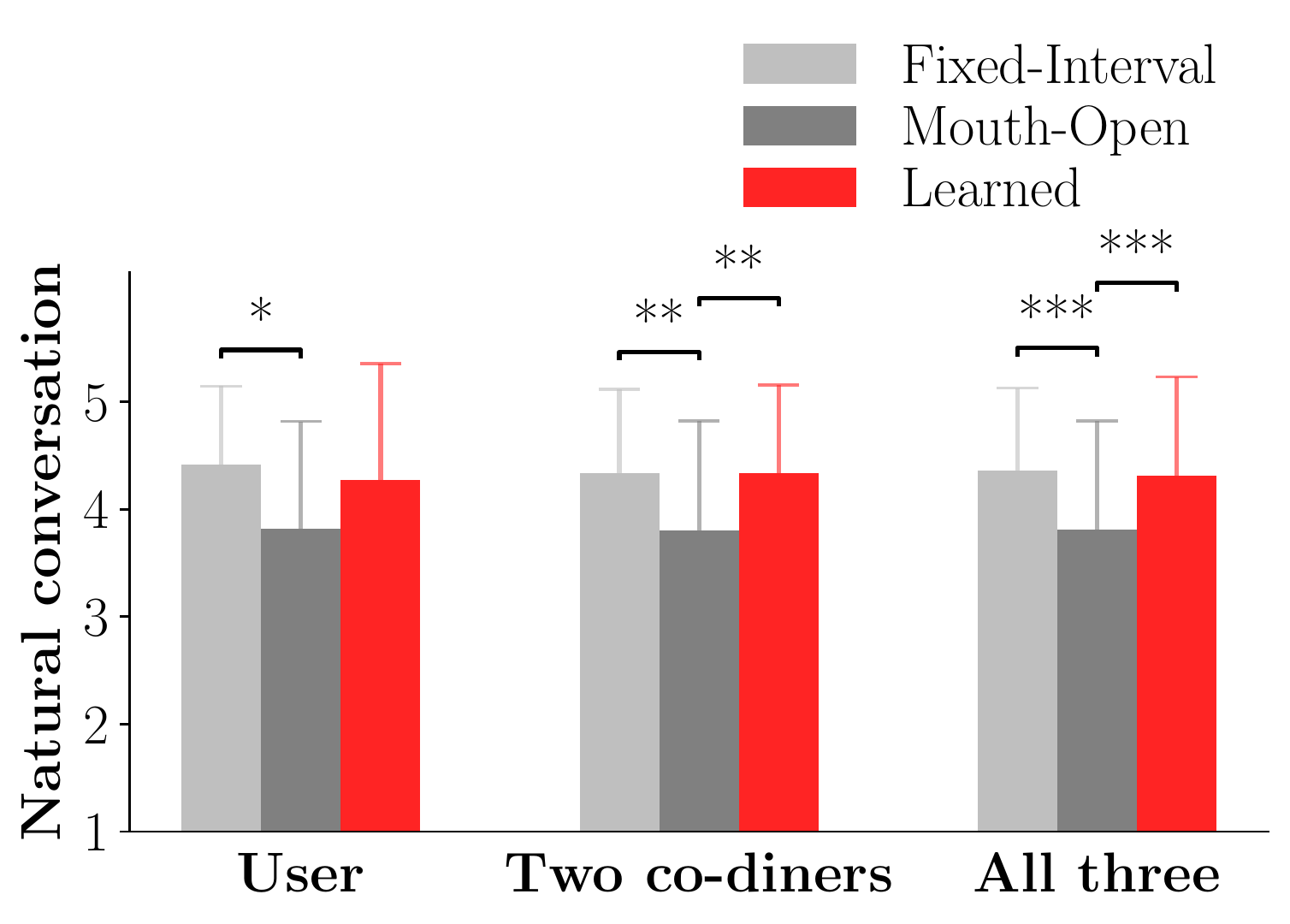}
     \includegraphics[width=0.495\textwidth]{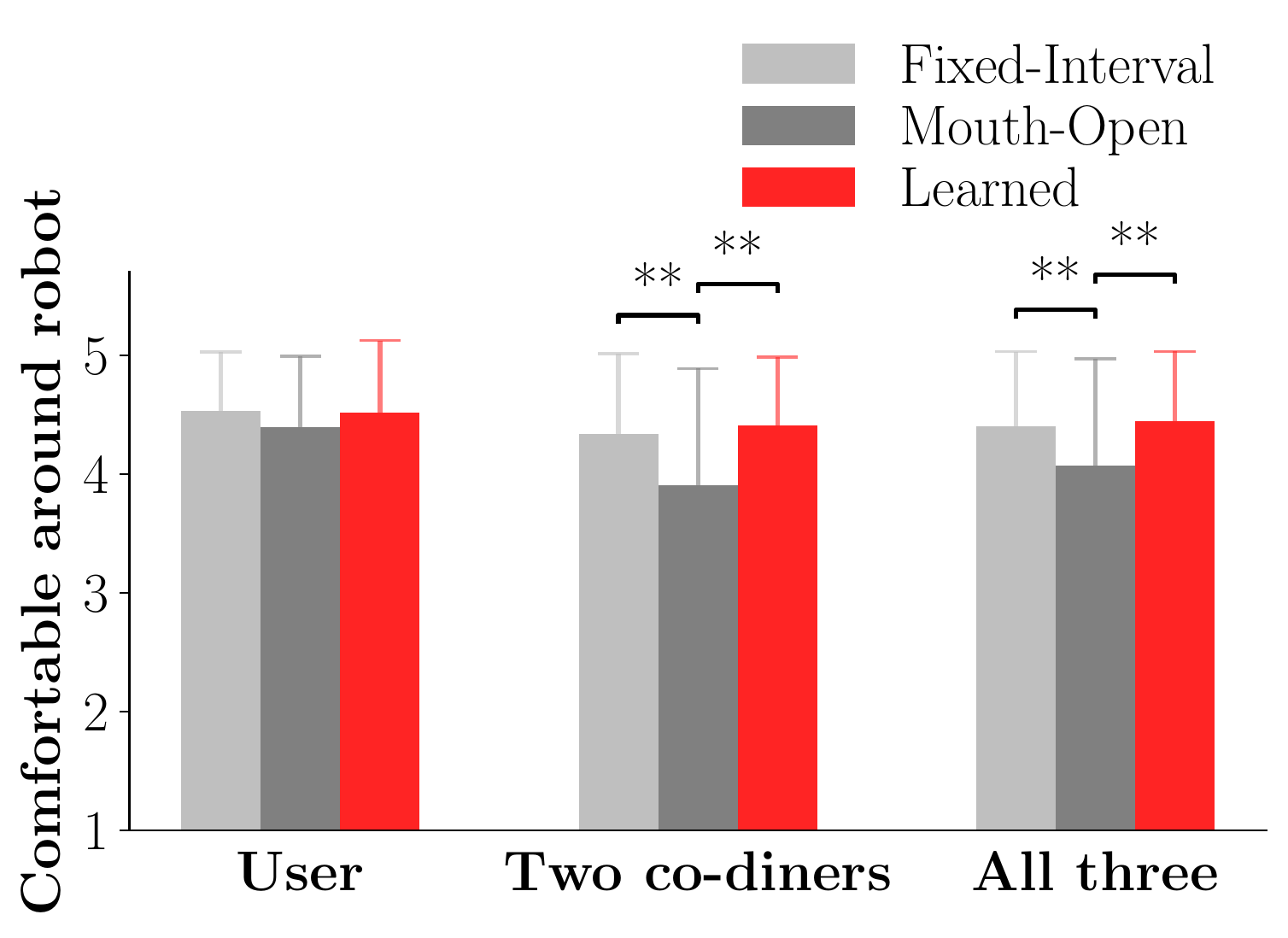}
     \includegraphics[width=0.495\textwidth]{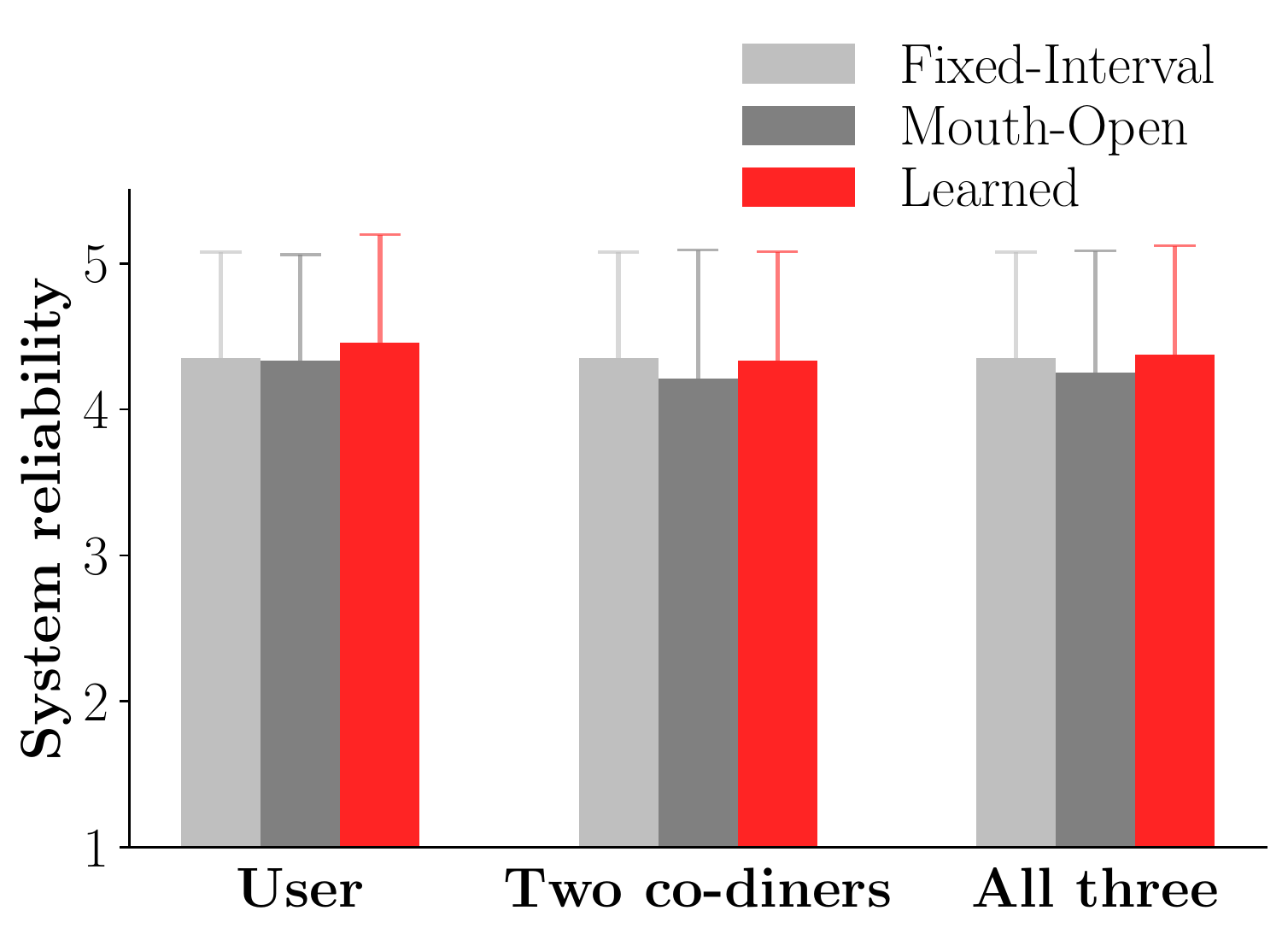}
     \includegraphics[width=0.495\textwidth]{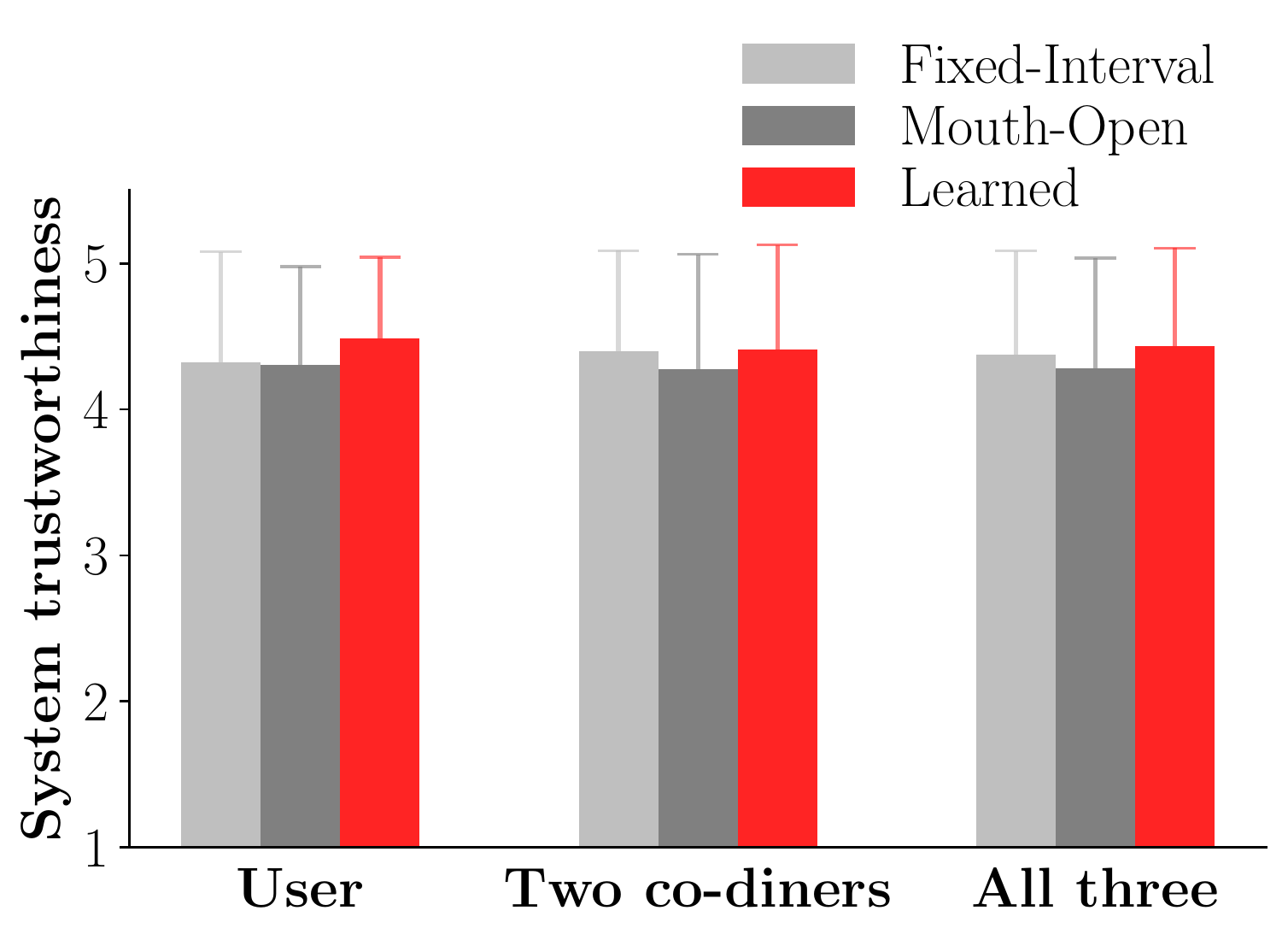}
    \caption{HRCom: 
        \textbf{Top left:} Ability to have natural conversation. 
        \textbf{Top right:} Ability to feel comfortable around the robot. 
        \textbf{Bottom left:} System reliability. 
        \textbf{Bottom right:} System trustworthiness. 
        Perceived by users, two co-diners, and all three diners on a Likert scale 1-5 (Strongly disagree - Strongly agree), for each bite timing strategy (Fixed-Interval, Mouth-Open, and Learned timing).
        $*, **, ***$ denote statistically significant differences with $p_{0.05}, p_{0.005}, p_{0.0005}$ respectively.
    }
    \label{fig:user_study:exp:likert5}
\end{figure}

\begin{figure}[!htb]
    \centering
     \includegraphics[width=0.495\textwidth]{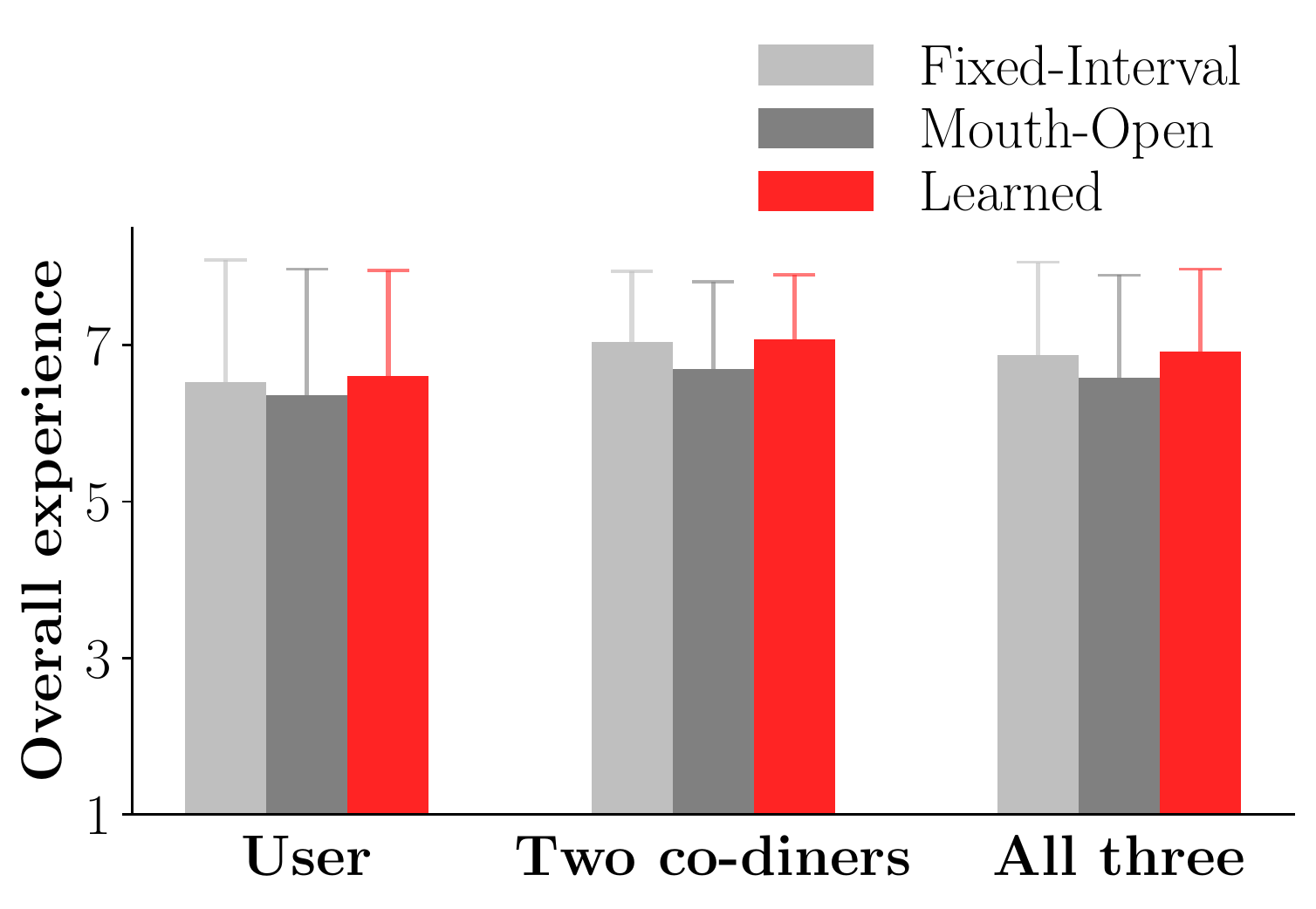}
     \includegraphics[width=0.495\textwidth]{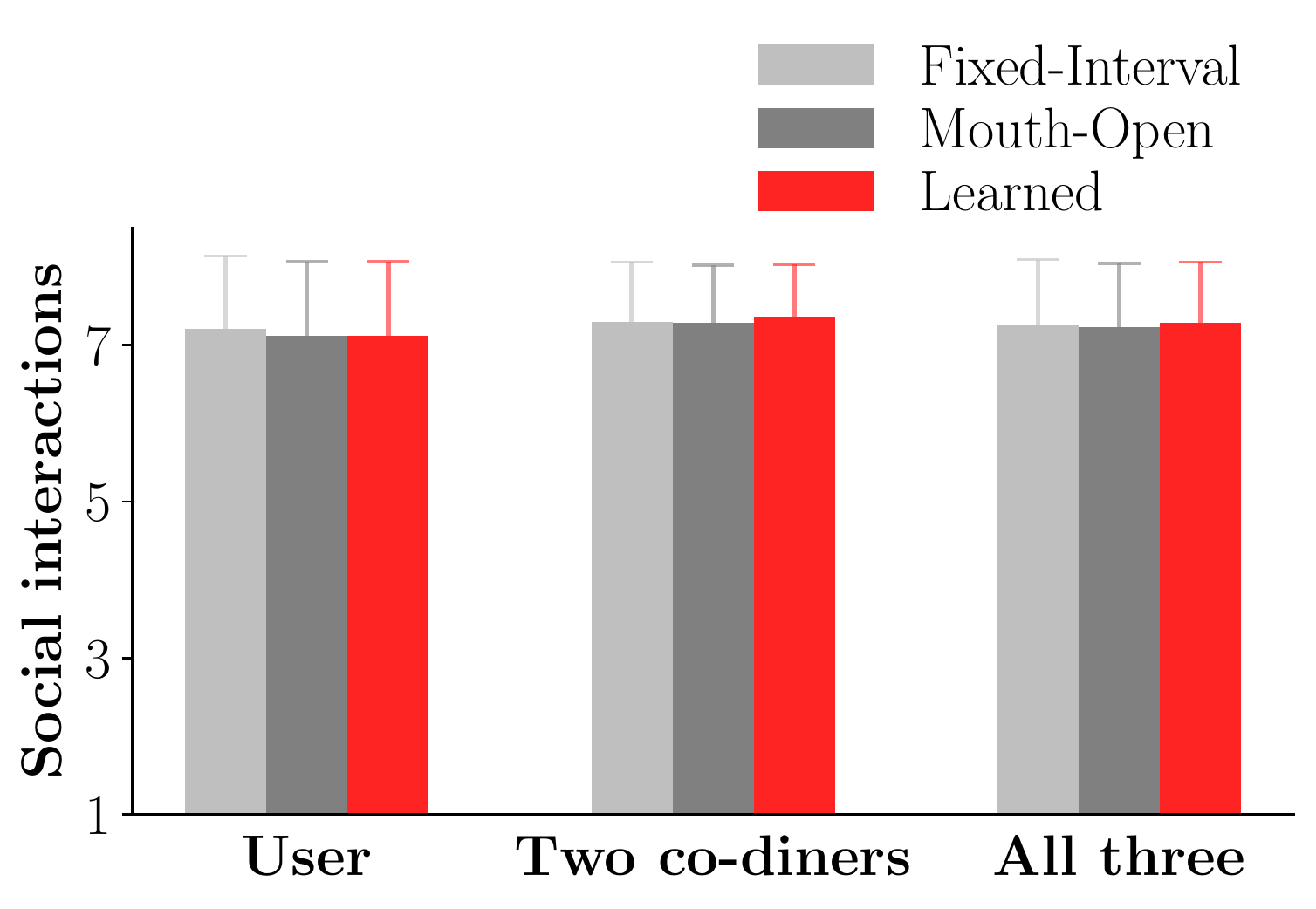}
    \caption{HRCom: 
        \textbf{Left:} Overall meal experience 
        \textbf{Right:} Social interactions
        perceived by users, two co-diners, and all three diners on a Likert scale 1-7 (Strongly disagree - Strongly agree), for each bite timing strategy (Fixed-Interval, Mouth-Open, and Learned timing).
        Our study did not find any statistically significant differences with $p_{0.05}$.
    }
    \label{fig:user_study:exp:likert7}
    \vspace{-2mm}
\end{figure}

\textbf{Other factors.} Besides bite timing itself, we evaluate differences between bite timing strategies for other factors: 
distraction by the robot (already discussed in Sec.~\ref{sec:user_study:results}),
ability to have natural conversations (Fig.~\ref{fig:user_study:exp:likert5}~(top left)),
ability to feel comfortable around the robot (Fig.~\ref{fig:user_study:exp:likert5}~(top right)),
system reliability (Fig.~\ref{fig:user_study:exp:likert5}~(bottom left)), 
system trustworthiness (Fig.~\ref{fig:user_study:exp:likert5}~(bottom right)), 
overall experience of the meal (Fig.~\ref{fig:user_study:exp:likert7}~(left)), 
social interactions with other participants (Fig.~\ref{fig:user_study:exp:likert7}~(right)).
We can see that the ability to have natural conversation and feel comfortable around robot is significantly lower for Mouth-Open timing than for Learned or Fixed-Interval timing.
This aligns with the finding in Sec.~\ref{sec:user_study:results} that
Mouth-Open timing distracts dining participants significantly more than Learned or Fixed-Interval timing.
It is however interesting to note that co-diners, not users of the robot, felt less comfortable around the robot during Mouth-Open Timing.
We speculate this is because co-diners perceive the prompt from a voice interface as an external disruption factor not related to their own eating whereas for robot users it is what makes them feed so it does not set robot users into discomfort as much.
For users, co-diners as well as all three diners, we do not find any statistically significant differences between bite timing strategies in system reliability, trustworthiness, overall experience nor social interactions they had.

\vspace{2mm}
\textbf{Replies to open-ended post-study questions.}
We also analyze the study participants' answers to open-ended questions from Fig.~\ref{fig:appendix:user_study:user_study_questionnaire}~(c).
We observe the following patterns. 

\vspace{3mm}
\textcolor{purple}{\textasteriskcentered{} \textit{Whether participants felt safe around the robot}}
\begin{itemize}
    \item Robot users:
    \begin{itemize}
        \item 7 users (70\%) replied with \textit{"Yes, \dots"}
        \item Their main concern was \textbf{safety when robot was moving around with the fork}: \textit{"Yes for the most part. Sometimes it felt surprising how close it got to my face when it went to pick up food.",
        "Yes. At first I thought it was going to stab me in the face but it moved slow and never hurt me.",
        "Sort of. My primary concern was the robot’s resting state. When the neutral position has the fork poised at eye level it is very concerning. Simply aiming the fork down and away from the table would make a huge difference."
        }
        \item Many noted that the \textbf{initial familiarization with the robot helped}: \textit{"Not very safe at beginning , but with trials go on, I feel more safe. If I have e stop myself I’ll feel more safe.",
        "I was a bit nervous, but after the first few trials I felt more comfortable around the robot",
        "Yes! Very quickly got used to it’s actions, which were very regular so easy to get used to safety-wise"
        }
    \end{itemize}
    
    \item Other co-diners:
    \begin{itemize}
        \item \textbf{No major concerns:} \textit{"Yes. It seemed to be under control nicely.",
        "Yes: it helped that the robot moved pretty slowly and along familiar "tracks" through the air. The e-stop was nice to have too!",
        "Yes. It avoided my friends and I well. It didnt seem overpowering.",
        "Moderately. The robot was cutting it close to the person's face while going to grab the food"
        }
        
        \item Similarly to robot users, the \textbf{initial familiarization with the robot helped}: \textit{"Yes, after a few trials I felt safe around the robot. But might be because it's far away from me as well",
        "I was a little uncomfortable initially but I started feeling safe after a few trials.",
        "I was little uneasy at first but then I quickly forgot about it and was comfortable",
        "Yes. I was a little worried at first, but wound up feeling very comfortable around the robot."
        }
    \end{itemize}

\end{itemize}

These replies show that familiarizing users as well as other participants with the robot helps them to feel safer around the robot.

\textcolor{purple}{\textasteriskcentered{} \textit{When participants think it is appropriate to take a bite of food when they are eating with others}}
\begin{itemize}

    \item \textbf{Talking-related rules:} \textit{"Usually when someone else is speaking and you are not expecting a question to be asked to you",
    "When I am not talking, or being directly talked to"
    }
    
    \item \textbf{Eye gaze-related rules:} \textit{"After a few second pause in speech combined with a stationary eye position.",
    "\dots if it is a very serious topic, or they are making eye contact, I would probaly wait."
    }
    
    \item \textbf{Diner physical state-related rules:} \textit{"When you are not speaking and have the desire to take a bit."
    }
    
    \item \textbf{Social interaction-related rules:} \textit{"\dots when the present speaker is not saying anything very emotional, energetic, or charged. For example I would not like to take a bite when consoling a crying friend.",
    "\dots if it is a very serious topic, or they are making eye contact, I would probaly wait."
    }
    
    \item \textbf{Time-related rules:} \textit{"When there is a stop(all people stop talking) longer than 1.5s, I feel it’s right time \dots"
    }
    
    \item \textbf{Robot-related:} \textit{"The robot shouldn't wait too long after the food is on the fork.",
    "Almost always. I would say worst case the biter can wait to make the move towards the robot, but it seems very appropriate for the robot to “always” be feeding and take a bite almost immediately when it’s ready."
    }
\end{itemize}

These replies 
match the same kinds of rules we find in replies to the same question asked during human-human commensality (App.~\ref{sec:appendix:dataset:stats}).

\textcolor{purple}{\textasteriskcentered{} \textit{What participants liked about the meal experience}}
\begin{itemize}
    \item Robot users:
    \begin{itemize}
        \item \textbf{Food and conversation:} \textit{"Conversation with people, fruit",
        "It was still easy to have a natural conversation, \dots"
        }
        
        \item \textbf{Robot and its behavior:} \textit{"I liked that the robot did the same thing over and over, making it easy to ignore",
        "\dots the robot was relatively quiet. It’s kind of nice to be fed and I like fruit.",
        "The food item is placed in a proper position, not too far or close, I have the choice to eat or not.",
        "The robot was generally out of the way. Once we went through a few trials, the robot was less distracting",
        "It was very nice not having to think about bite acquisition and delivery",
        "\dots enjoyed the novelty of the robot"
        }
    \end{itemize}
    
    \item Other co-diners:
    \begin{itemize}
        \item \textbf{Food and conversation:} \textit{"My food was great. People were too.",
        "I was able to have a natural conversation",
        "\dots The robot was not too intrusive and was almost a cool fourth diner."
        }
        
        \item \textbf{Robot and its behavior:} \textit{"It did not take too long to become accustomed to the robot.",
        "\dots the robot was not as much of a disruption as I imagined.",
        "I feel the pace was nice, and I felt more normal than I expected. The conversation flow was good and not interrupted by the robot.",
        "\dots that the robot waited till was a natural pause in conversation from participant 1 before "speaking" or coming forward with the food so the experience was pretty smooth.",
        "interesting to watch the robot moves, and I felt the robot wasn't that distracting when it didn't make any sound",
        "There were many times when the robot was very much in the background of the conversation and the conversational flow was uninterrupted"
        }
        
        \item \textbf{Dining setting:} \textit{"Circular table made for nice discussion atmosphere. 3 people is also nice so we can talk while the other is being fed."
        }
    \end{itemize}
\end{itemize}

Both users and co-diners liked food and conversation which matches what participants during human-human commensality experienced (in replies to open-ended questions in App.~\ref{sec:appendix:dataset:stats}). This suggests that the addition of the assistive feeding robot does not remove these particular factors of commensality that people like.
Also, participants seemed to like the robot behavior and its presence as a new element in commensality.

\textcolor{purple}{\textasteriskcentered{} \textit{What participants did not like about the meal experience}}
\begin{itemize}
    \item Robot users:
    \begin{itemize}
        \item 7 robot users (70\%) found the \textbf{voice prompts during Mouth-Open Timing distracting}: \textit{"Robot was very distracting especially when it spoke commands",
        "When the robot talks, it breaks the flow of the conversation.",
        "The voice that told me to look at the robot was sometimes distracting.",
        "I didn’t like the trials when the robot prompted to eat",
        "I don’t like voice interruption by robot \dots",
        "When the robot spoke it would cut off the conversation."
        }
        
        \item \textbf{Robot position, speed, and noise:} \textit{"I didn’t like how it became harder to make eye contact why talking because sometimes the robot would block out eye level \dots",
        "\dots robot was a bit slow so I didn’t get to eat much \dots",
        "\dots the noise make by robot in operation makes others voice hard to heard clearly."
        }
    
        \item \textbf{Questionnaires after each trial:} \textit{"\dots taking survey in between bites also was challenging as it interrupted the flow"
        }
        
        \item \textbf{Bite timing:} \textit{"Sometimes the robot was distracting when I was in the middle of a story",
        "\dots it was weird because I felt like I couldn’t signal when I wanted the food and had to wait.",
        "The robot took too long to feed me. It would take several hours to eat enough food with its speed. There is a tradeoff between timely feeding and fast enough feeding to finish a meal in a proper amount of time."
        }
    \end{itemize}
    
    \item Other co-diners:
    \begin{itemize}
    
        \item 9 co-diners (45\%) found the \textbf{voice prompts during Mouth-Open Timing distracting}: \textit{"I didn't love the trials when the robot spoke \dots",
        "Robot was sometimes speaking in the middle of the conversation"
        }
        
        \item \textbf{Robot position:} \textit{"robot blocked my sight when I talked to the person on the left side"}
        
        \item \textbf{Time to get used to the robot:} \textit{"Not much. It took a while to get used to the robot.",
        "When the machine spoke over us it was hard to keep the conversatiom going, although this became easier over time."
        }
    
        \item \textbf{Questionnaires after each trial:} \textit{"Interruptions for the survey broke up the conversation"
        }
        
        \item \textbf{Bite timing:} \textit{"A couple of trials, the robot came in slightly early or waited for a while."
        }
        
        \item \textbf{Conversation content:} \textit{"We all consciously or unconsciously had to structure our conversation around what the robot was doing at a particular point of time."
        }
    \end{itemize}
\end{itemize}

These replies clearly show that both robot users as well as co-diners find the Mouth-Open bite timing strategy disrupts the flow of the conversation.
As several participants reported that the robot movements interrupted their mutual eye contacts, it would be interesting to explore robot bite transfer trajectories that minimize eye gaze blockage.


\def \figsizehalf {.48\textwidth}
\begin{figure}[t]
	\centering
	\begin{tabular}{c c}
	    \centering
	    \parbox{\figsizehalf}{
	        \centering
            \includegraphics[width=\figsizehalf]{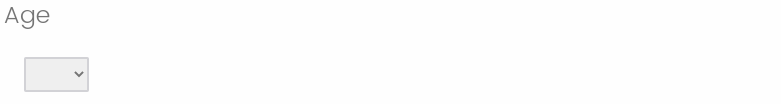}
            \includegraphics[width=\figsizehalf]{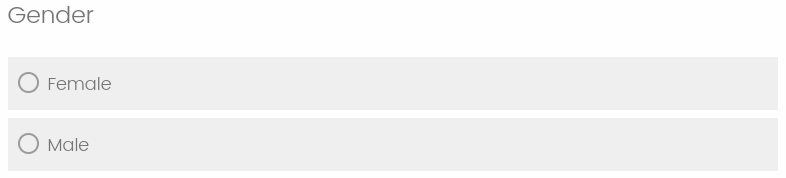}
            \includegraphics[width=\figsizehalf]{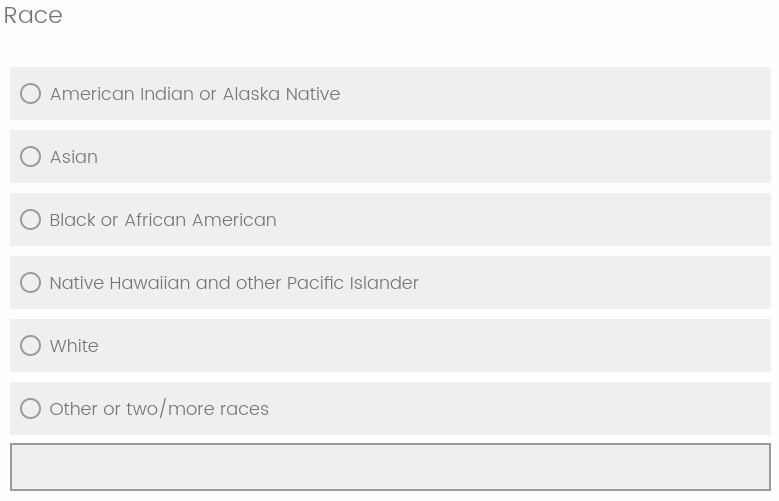}
            \includegraphics[width=\figsizehalf]{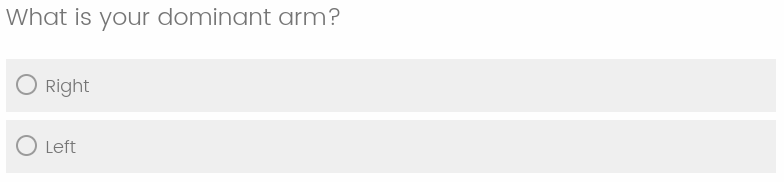}
    		\includegraphics[width=\figsizehalf]{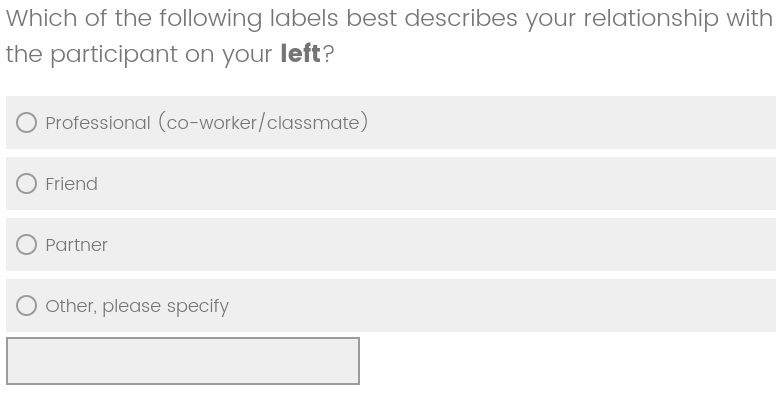}
    		\includegraphics[width=\figsizehalf]{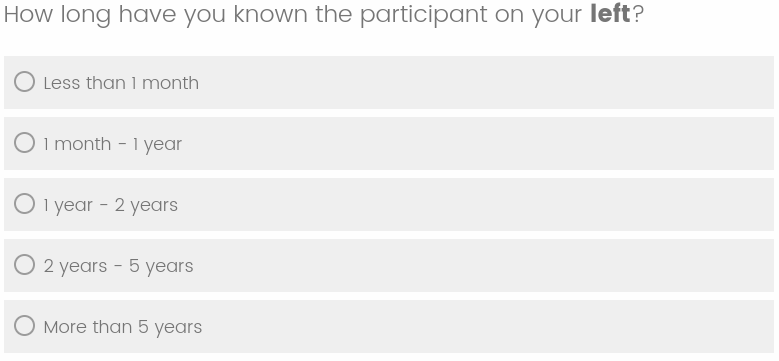}
    		\includegraphics[width=\figsizehalf]{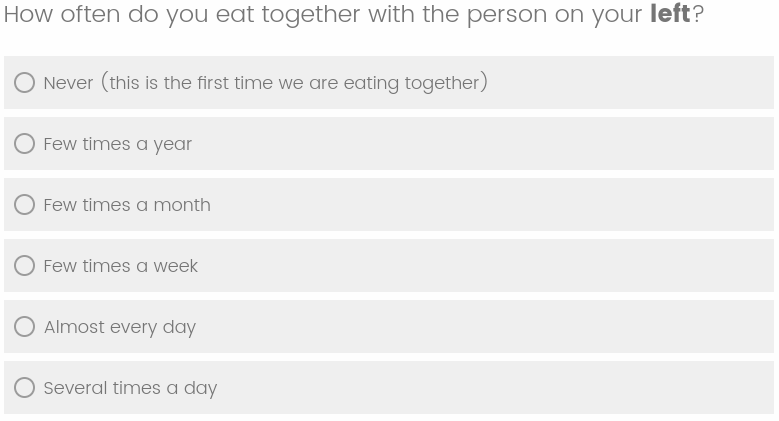}
    	}
    	\parbox{\figsizehalf}{
    	    \centering
            \includegraphics[width=\figsizehalf]{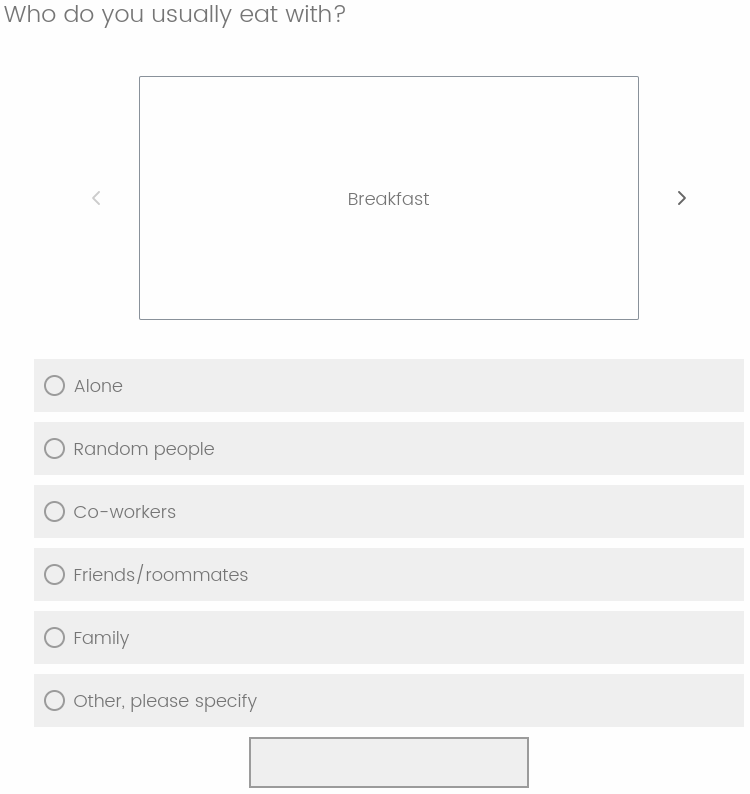}\vspace{5mm}
            \includegraphics[width=\figsizehalf]{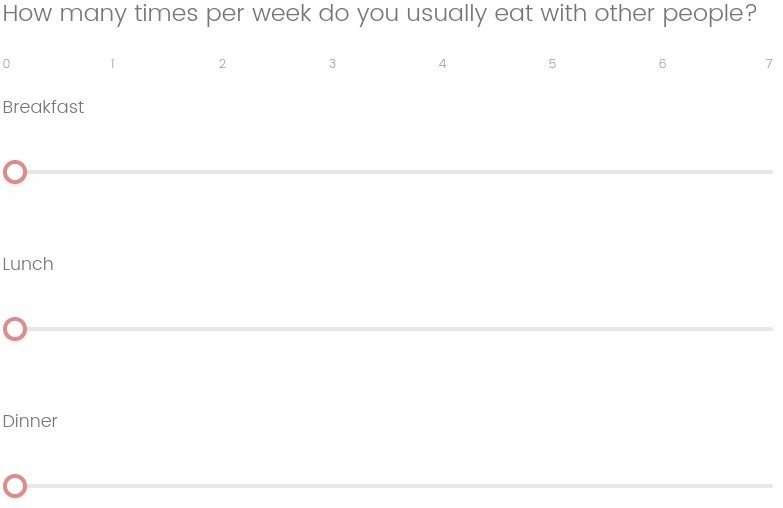}\vspace{6mm}
            \includegraphics[width=\figsizehalf]{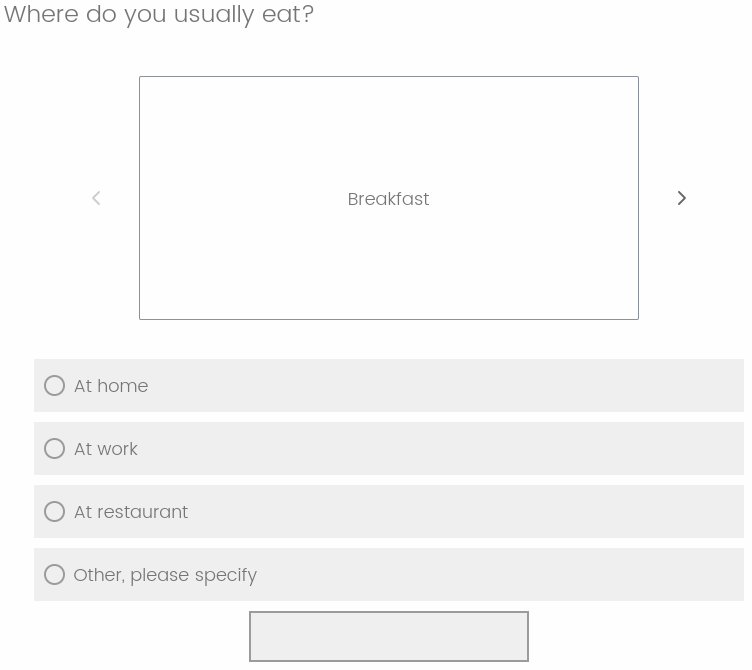}
        }
    \end{tabular}
	\caption{HHCD and HRCom: Pre-study questionnaire: questions about demographic background, relationship to other participants (the same questions were asked in relation to the participant on the right), and social dining habits.
	}
	\label{fig:appendix:dataset:collection:prestudy_questionnaire}
\end{figure}

\def \figsizehalf {.48\textwidth}
\begin{figure}[t]
    \centering
    \begin{tabular}{c c}
        \centering
    	\parbox{\figsizehalf}{
    	    \centering
            \includegraphics[width=\figsizehalf]{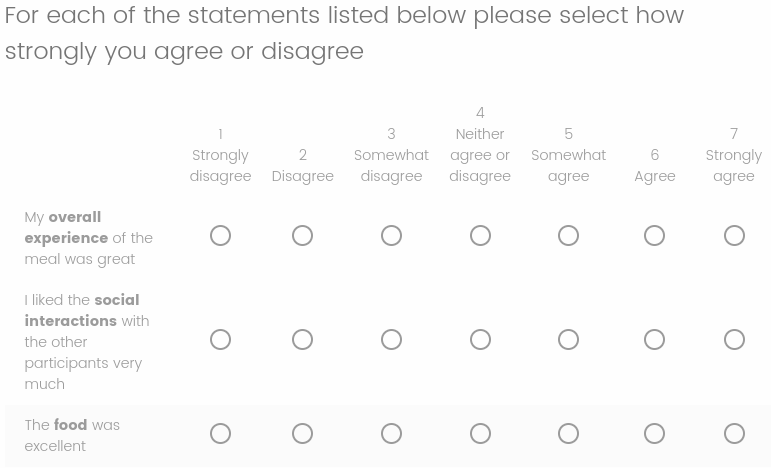}
    	}
    	\parbox{\figsizehalf}{
    	    \centering
            \includegraphics[width=\figsizehalf]{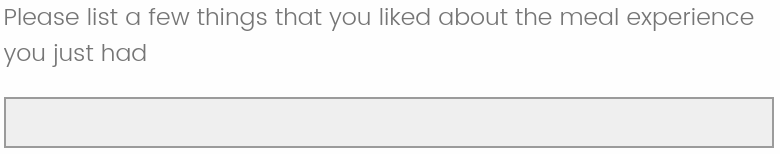}\vspace{1mm}\\
            \includegraphics[width=\figsizehalf]{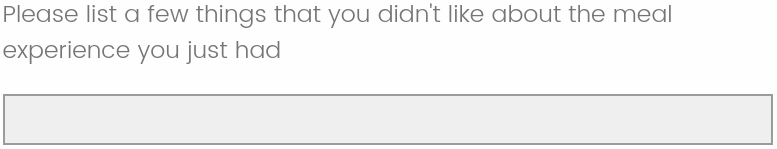}\vspace{1mm}\\
            \includegraphics[width=\figsizehalf]{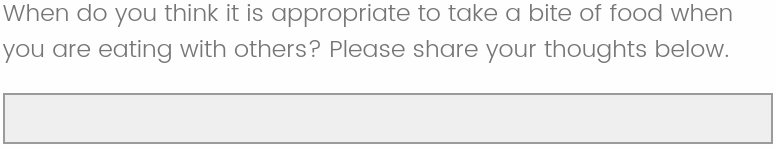}
    	}
    \end{tabular}
	\caption{HHCD and HRCom: Post-study questionnaire: questions about dining experience.
	}
	\label{fig:appendix:dataset:collection:poststudy_questionnaire}
\end{figure}

\def \figsizehalf {.48\textwidth}
\begin{figure}[t]
    \centering
    \begin{tabular}{c c}
        \centering
    	\parbox{\figsizehalf}{
    	    \centering
            \includegraphics[width=\figsizehalf]{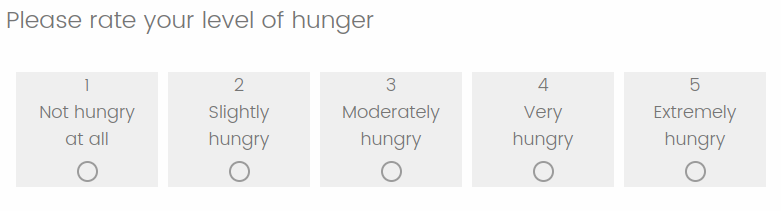}
            (a)\vspace{4mm}\\
    	    \includegraphics[width=\figsizehalf]{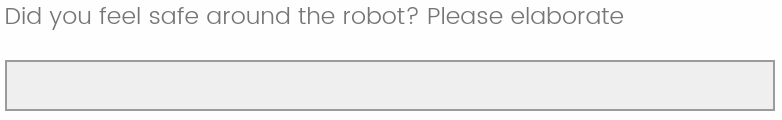}\vspace{1mm}\\
            \includegraphics[width=\figsizehalf]{figs/dataset/poststudy_questionnaire2.png}\vspace{1mm}\\
            \includegraphics[width=\figsizehalf]{figs/dataset/poststudy_questionnaire3.png}\vspace{1mm}\\
            \includegraphics[width=\figsizehalf]{figs/dataset/poststudy_questionnaire4.png}\vspace{1mm}\\
            (c)
    	}
    	\parbox{\figsizehalf}{
    	    \centering
            \includegraphics[width=\figsizehalf]{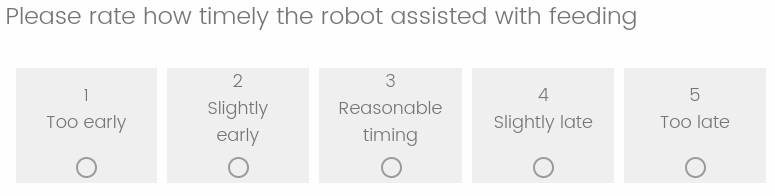}\vspace{1mm}\\
            \includegraphics[width=\figsizehalf]{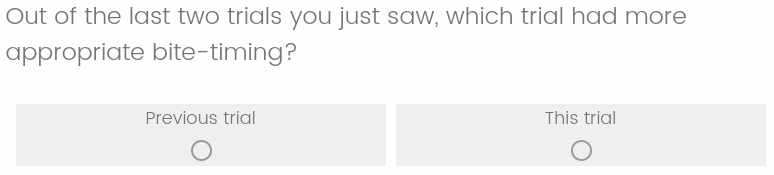}\vspace{1mm}\\
            \includegraphics[width=\figsizehalf]{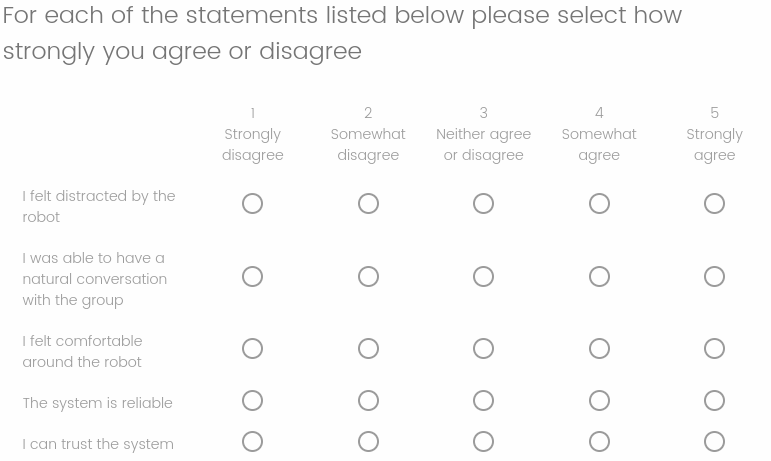}\vspace{1mm}\\
            \includegraphics[width=\figsizehalf]{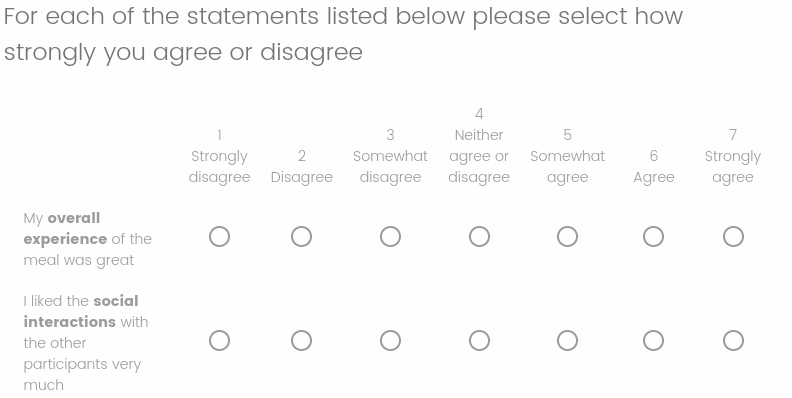}
            (b)
    	}
    \end{tabular}
	\caption{HRCom: Questionnaires: 
	(a) Additional question asked in pre-study questionnaire in addition to questions in Fig.~\ref{fig:appendix:dataset:collection:prestudy_questionnaire}.
	(b) Experiment questionnaire asked after each trial. Note, we did not ask the second forced-choice question after the first trial. 
	(c) Post-study questionnaire.
	}
	\label{fig:appendix:user_study:user_study_questionnaire}
\end{figure}


\end{document}